\def\year{2021}\relax
\documentclass[letterpaper]{article} % DO NOT CHANGE THIS
\usepackage{aaai21}  % DO NOT CHANGE THIS
\usepackage{times}  % DO NOT CHANGE THIS
\usepackage{helvet} % DO NOT CHANGE THIS
\usepackage{courier}  % DO NOT CHANGE THIS
\usepackage[hyphens]{url}  % DO NOT CHANGE THIS
\usepackage{graphicx} % DO NOT CHANGE THIS
\usepackage{natbib}  % DO NOT CHANGE THIS AND DO NOT ADD ANY OPTIONS TO IT
\usepackage{caption} % DO NOT CHANGE THIS AND DO NOT ADD ANY OPTIONS TO IT
\usepackage{hyperref}
\hypersetup{colorlinks,urlcolor=blue,citecolor=blue}
\usepackage{amsfonts}
\usepackage{mathtools}
\usepackage[english]{babel}
\usepackage{amsmath,amssymb,amsthm}
\usepackage{tabularx}
\usepackage{algorithm}
\usepackage{algpseudocode}
\usepackage{multirow}
\usepackage{array}
\usepackage{listings}
\usepackage{diagbox}
\usepackage{tikz}
\usepackage{scalerel}
\usepackage{comment}

\usepackage[utf8]{inputenc}

\usepackage{pgfplots}
\pgfplotsset{compat=newest}
\usepgfplotslibrary{statistics}
\usepgfplotslibrary{groupplots}
\usepgfplotslibrary{dateplot}
\usepackage[caption=false,labelformat=empty]{subfig}
\pgfplotsset{ every non boxed x axis/.append style={x axis line style=-},
     every non boxed y axis/.append style={y axis line style=-}}
     
\pgfplotsset{
    every first x axis line/.style={},
    every first y axis line/.style={},
    every first z axis line/.style={},
    every second x axis line/.style={},
    every second y axis line/.style={},
    every second z axis line/.style={},
    first x axis line style/.style={/pgfplots/every first x axis line/.append style={#1}},
    first y axis line style/.style={/pgfplots/every first y axis line/.append style={#1}},
    first z axis line style/.style={/pgfplots/every first z axis line/.append style={#1}},
    second x axis line style/.style={/pgfplots/every second x axis line/.append style={#1}},
    second y axis line style/.style={/pgfplots/every second y axis line/.append style={#1}},
    second z axis line style/.style={/pgfplots/every second z axis line/.append style={#1}}
}

\makeatletter
\def\pgfplots@drawaxis@outerlines@separate@onorientedsurf#1#2{%
    \if2\csname pgfplots@#1axislinesnum\endcsname
    \else
    \scope[/pgfplots/every outer #1 axis line,
        #1discont,decoration={pre length=\csname #1disstart\endcsname, post length=\csname #1disend\endcsname}]
        \pgfplots@ifaxisline@B@onorientedsurf@should@be@drawn{0}{%
            \draw [/pgfplots/every first #1 axis line] decorate {
                \pgfextra
                \pgfplotspointonorientedsurfaceabsetupfor{#2}{#1}{\pgfplotspointonorientedsurfaceN}%
                \pgfplots@drawgridlines@onorientedsurf@fromto{\csname pgfplots@#2min\endcsname}%
                \endpgfextra 
                };
        }{}%
        \pgfplots@ifaxisline@B@onorientedsurf@should@be@drawn{1}{%
            \draw [/pgfplots/every second #1 axis line] decorate {
                \pgfextra
                \pgfplotspointonorientedsurfaceabsetupfor{#2}{#1}{\pgfplotspointonorientedsurfaceN}%
                \pgfplots@drawgridlines@onorientedsurf@fromto{\csname pgfplots@#2max\endcsname}%
                \endpgfextra 
                };
        }{}%
    \endscope
    \fi
}%
\makeatother

\newtheorem{proposition}{Proposition}

\newcommand{\paragraphb}[1]{\vspace{0.03in} \noindent{\bf #1} }
\newcommand{\paragraphe}[1]{\vspace{0.03in} \noindent{\em #1} }

\usepackage{colortbl}

\makeatletter
\newcommand*\titleheader[1]{\gdef\@titleheader{#1}}
\AtBeginDocument{%
  \let\st@red@title\@title
  \def\@title{%
    \bgroup\normalfont\large\centering\@titleheader\par\egroup
    \vskip1.5em\st@red@title}
}
\makeatother

\title{Membership Privacy for Machine Learning Models Through Knowledge Transfer}
\titleheader{\em To Appear in the 35th AAAI Conference on Artificial Intelligence, 2021}
\author {
        Virat Shejwalkar\quad
        Amir Houmansadr\\
}
\affiliations {
    University of Massachusetts Amherst \\
    \{vshejwalkar, amir\}@cs.umass.edu
}

\usepackage[switch]{lineno}

\begin{document}

\maketitle
% !TEX root = main.tex

\begin{abstract}

Large capacity machine learning (ML) models are prone to membership inference attacks (MIAs), which aim to infer whether the target sample is a member of the target model's training dataset.
The serious privacy concerns due to the membership inference have motivated multiple defenses against MIAs, e.g., differential privacy and adversarial regularization. 
Unfortunately, these defenses produce ML models with unacceptably low classification performances.

Our work proposes a new defense, called \emph{distillation for membership privacy} (DMP), against MIAs that preserves the utility of the resulting models significantly better than prior defenses. 
DMP leverages knowledge distillation to train ML models with membership privacy.
We provide a novel criterion to tune the data used for knowledge transfer in order to amplify the membership privacy of DMP.

Our extensive evaluation shows that DMP provides significantly better tradeoffs between membership privacy and classification accuracies compared to state-of-the-art MIA defenses.
For instance, DMP achieves $\sim$100\% accuracy improvement over adversarial regularization for DenseNet trained on CIFAR100, for similar membership privacy (measured using MIA risk): when the MIA risk is 53.7\%, adversarially regularized DenseNet is 33.6\% accurate, while DMP-trained DenseNet is 65.3\% accurate.

\end{abstract}
% !TEX root = main.tex

\vspace*{-1em}
\section{Introduction}\label{introduction}

\begin{comment}
- DMP overview not needed
- 
\end{comment}

The remarkable performance of machine learning (ML) in solving many classification tasks has facilitated its adoption in various domains ranging from recommendation systems to critical health-care management.
Many ML-as-a-Service platforms (e.g., Google API, Amazon AWS) enable novice data owners to train ML models and release the models either as a blackbox prediction API or as model parameters that can be accessed in whitebox fashion.

ML models are often trained on data with sensitive user information such as clinical records and personal photos.  Hence, ML models trained using sensitive data can leak private information about their data owners.
This has been demonstrated through various inference attacks~\cite{fredrikson2015model,hitaj2017deep,carlini2018secret}
% (Fredrikson et al.~\citeyear{fredrikson2015model}, Hitaj et al.~\citeyear{hitaj2017deep}, Carlini et al.~\citeyear{carlini2018secret})
, and most notably the \emph{membership inference attack} (MIA)~\cite{shokri2017membership} which is the focus of our work.
An MIA adversary with a blackbox or whitebox access to a target model aims to determine if a given target sample belonged to the private training data of the target model or not.
% The attack performance significantly improves with a whitebox access to the trained models~\cite{nasr2019comprehensive}.
% This serious privacy concern due to membership inference is shown to be effective against multiple ML models and ML-as-a-service platforms in both the blackbox and whitebox settings \cite{shokri2017membership,melis2019exploiting}.
MIAs are able to distinguish the members from non-members by \emph{learning} the behavior of the target model on member versus non-member inputs. 
They use different features of the target model for this classification, e.g., model predictions \cite{shokri2017membership}, model loss, and gradients of the model parameters for given input~\cite{nasr2019comprehensive}.
MIAs are particularly more effective against deep neural networks~\cite{shokri2017membership,salem2019ml}, because, with their large capacities, such models can better memorize their training data.

Recent work has investigated several defenses against  membership inference attacks.
In order to provide the worst case privacy guarantees, \emph{Differential Privacy} (DP) based defenses add very large amounts of noise to the learning objective or model outputs~\cite{papernot2017semi,chaudhuri2011differentially}
% (Papernot et al.~\citeyear{papernot2017semi}, Chaudhuri et al.~\citeyear{chaudhuri2011differentially})
. This results in  models with unacceptable tradeoffs between privacy and utility~\cite{jayaraman2019evaluating}, therefore questioning their use in practice.
Sablayrolles et al.~\cite{sablayrolles2019white} showed that membership privacy is a weaker notion of privacy than DP, which improves with generalization of ML models.
Similarly, Nasr et al.~\cite{nasr2018machine} proposed \emph{adversarial regularization} targeted to defeat MIAs by improving the target model's generalization.
However, as we demonstrate, the adversarial regularization and other state-of-the-art regularizations, including label smoothing~\cite{szegedy2016rethinking} and dropout~\cite{srivastava2014dropout}, fail to provide acceptable membership privacy-utility tradeoffs (simply called `tradeoffs' here onward).
Memguard~\cite{jia2019memguard}, a blackbox defense, improves model utility, but it cannot protect the model from whitebox MIAs and even the simple threshold based MIAs~\cite{yeom2018privacy}.
In summary, \emph{existing defenses against MIAs offer poor tradeoffs between model utility and membership privacy}.

To this end, our work proposes a defense against MIAs that significantly improves the tradeoffs compared to prior defenses. 
That is, for a given degree of membership privacy  (i.e., MIA resistance), our defense produces models with significantly higher classification performances compared to prior defenses.
Our defense, called \emph{\textbf{D}istillation for \textbf{M}embership \textbf{P}rivacy} (DMP), leverages  \emph{knowledge distillation}~\cite{hinton2014distilling}, which transfers the knowledge of large models to smaller models, and is primarily used for model compression.
Intuitively, DMP protects membership privacy by thwarting the access of the resulting models to the private training data.
The first \emph{pre-distillation} phase of DMP trains an \emph{unprotected} model on the private training data without any privacy protection.
Next, in \emph{distillation} phase, DMP selects/generates reference data and transfers the knowledge of the unprotected model into predictions of the reference data.
In the final \emph{post-distillation} phase, DMP trains a \emph{protected} model on the  reference data labeled in the previous phase.
Unlike conventional distillation, we use the same architectures for the unprotected and protected models.

Similar to adversarial regularization and PATE, DMP assumes access to a possibly sensitive and ``unlabeled'' \emph{reference data} drawn from the same distribution as the ``labeled'' private training data, and uses such reference data to train its final models; the reference data is not publicly available. 
This is a  highly realistic assumption as typical model generating entities (e.g., banks) possess huge amounts of ``unlabeled'' data (but limited labeled data due to the expensive labeling process).
Furthermore, we show that this assumption can be relaxed by synthesizing reference data using generator networks~\cite{micaelli2019zero}.
While some prior work~\cite{papernot2017semi} combined distillation and DP to protect data privacy, our work is \emph{the first} to study the promise of knowledge distillation as the sole technique to train membership privacy-preserving models. 
Our key contributions are summarized below:

\vspace*{-.2em}
\begin{itemize}
	\item[-] We propose a defense against MIAs, called \emph{\textbf{D}istillation for \textbf{M}embership \textbf{P}rivacy} (DMP). 
	% Figure~\ref{fig:dml_blocks} gives an overview of DMP.
	
	\item[-] Given an unprotected model trained on a private training data and a reference sample, we provide a novel result that the lower the entropy of prediction of the model on the reference sample, the lower the sensitive membership information in the prediction. We use this result to  select/generate appropriate reference data so as to improve the membership privacy due to DMP.
	
	\item[-] We perform an extensive evaluation of DMP to show the state-of-the-art tradeoffs between membership privacy and model accuracy of DMP. For instance, at a fixed high degrees of membership privacy, DMP achieves 30\% to 140\% higher classification accuracies compared to state-of-the-art defenses across various classification tasks.
\end{itemize}

% !TEX root = main.tex
\section{Related Work}\label{related}

\paragraphb{Membership inference attacks.}\label{related:meminf}
\cite{shokri2017membership} introduced membership inference attacks (MIAs).
Given a target model trained on a private training data and a target sample, MIA adversary aims to infer whether the target sample is a member of the private training data.
\cite{shokri2017membership} proposed to train a neural network to distinguish the features of the target model on members and non-members. They assumed a partial access to the private trainin data.
\cite{salem2019ml} relaxed this assumptions and showed the transferability of MIAs across datasets.
These works relied on the blackbox features of target models, e.g., model predictions to mount MIAs.
\cite{nasr2019comprehensive} proposed to use whitebox features of target models, e.g., model gradients, along with the blackbox features, to further enhance the MIA accuracy.
Above works used generalization gap (i.e., difference in train and test accuracy) of target models to mount strong MIAs. 
The more recent MIA literature focuses on deriving features that can better distinguish the behavior of target models on members and non-members~\cite{leino2019stolen,song2020systematic}.

\paragraphb{Defenses against membership inference attacks. }\label{related:defenses}
MIAs exploit differences in behaviors of target models on members and non-members.
Regularization techniques, including dropout and label smoothing, reduce the difference in terms of accuracies of the target model on members and non-members, and mitigate MIAs to some extent~\cite{shokri2017membership}.
\cite{nasr2018machine} proposed adversarial regularization (AdvReg) tailored to defeat MIAs. AdvReg simultaneously trains the target and attack models in a game theoretic manner, and regularizes the target model using the accuracy of the attack model. 
The final target models that use above regularization defenses can be deployed in whitebox manner, i.e., similar to DMP, they are \emph{whitebox defenses}. Hence, we thoroughly compare our DMP defense with all these regularization techniques.
However, as shown in~\cite{song2020systematic} and seen from the original work~\cite{nasr2018machine}, AdvReg is not an effective defense, because it either fails to mitigate MIA or incurs large drops in model utility (classification accuracy).
Jia et al.~\shortcite{jia2019memguard} proposed MemGuard, a blackbox defense that adds noise to the output of the target model such that the noisy output is both accurate and fools the given MIA attack model.
However, MemGuard does not defend against the simplest of threshold based attacks~\cite{yeom2018privacy,sablayrolles2019white}. We omit MemGuard and other blackbox defenses, e.g., top-k predictions~\cite{shokri2017membership}, from evaluations.

Differential privacy based defenses such as DP-SGD~\cite{abadi2016deep} and PATE~\cite{papernot2017semi} are whitebox defenses and provide strong theoretical membership privacy guarantees.
% Target models trained in differentially private (DP) manner, using defenses such as , are resistant to MIAs~\cite{yeom2018privacy}.
However, as~\cite{jayaraman2019evaluating} show\textemdash and we confirm in our work\textemdash target models trained using DP-SGD and PATE have prohibitively low classification accuracies rendering them unusable.
% We compare our as these defenses are whitebox defenses.

\section{Preliminaries}\label{preliminaries}

\paragraphb{Knowledge distillation.}\label{prelim:distil}
\cite{bucilua2006model} and \cite{ba2014deep} proposed knowledge distillation, which uses the outputs of a large teacher model to train a smaller student model, in order to \emph{compress} large models to smaller models.
The outputs used for distillation can vary, e.g., 
\cite{hinton2014distilling} use class probabilities generated by the teacher as the outputs, while \cite{romero2014fitnets} use the intermediate activations along with class probabilities of the teacher.
It is well established that \emph{knowledge distillation produces students with accuracies similar to their teachers}~\cite{crowley2018moonshine,zagoruyko2016paying}. This also allows DMP to produce highly accurate target models.
Note that, although we use term ``distillation'', DMP uses teacher and student models of the same sizes, because DMP is not concerned with the size of the resulting model.

\paragraphb{Membership inference attacks. }
Below we give the threat model and MIA methodology that we consider in this work.

\paragraphb{\em{Threat model.} }
The primary \emph{goal} of the adversary is to infer the membership of a target sample $(\textbf{x},y)$ in the private training data $D_\mathsf{tr}$ of a target model $\theta$.
Our DMP defense uses private, unlabeled reference data $X_\mathsf{ref}$ for knowledge transfer, which itself could be privacy sensitive, hence, we consider a secondary goal to infer membership of a target sample in $X_\mathsf{ref}$.
Following the previous works, we assume a strong adversary with the \emph{knowledge} of: target model parameters (the strongest whitebox case), half of the members of $D_\mathsf{tr}$ and equal number of non-members. Similarly, to assess the MIA risk to $X_\mathsf{ref}$, we assume that the adversary has half of the members of $X_\mathsf{ref}$ and the equal number of non-members. Note that, the assumptions on the partial availability of private $D_\mathsf{tr}$ and private $X_\mathsf{ref}$ facilitates the assessment of defenses under a very strong adversary.
The adversary can compute various whitebox and blackbox features of the target model and train an attack model. The adversary \emph{cannot poison} $X_\mathsf{ref}$ as it is not publicly available.

\paragraphb{\em {Methodology.} }
Consider a target model $\theta$ and a sample $(\textbf{x},y)$. 
MIAs exploit the differences in the behavior of $\theta$ on members and non-members of the private $D_\mathsf{tr}$.
Therefore, MIAs train a binary attack model to classify target samples into members and non-members.
Such attack models can be neural networks~\cite{shokri2017membership,salem2019ml} or simple thresholding functions where threshold is tuned for maximum attack performance~\cite{yeom2018privacy,sablayrolles2019white,song2020systematic}.
% Let $h: F(X,Y,\theta)\rightarrow [0,1]$ be the inference model.
The adversary computes various features of $\theta$ for given $(\mathbf{x},y)$, e.g., prediction $\theta(\mathbf{x},y)$, $\theta$'s loss on $(\mathbf{x},y)$, and the gradients of the loss.
The adversary combines these features to form $F(\mathbf{x},y,\theta)$.
The attack model $h$ takes $F(\mathbf{x},y,\theta)$ as its input and outputs the probability that $(\mathbf{x},y)$ is a member of $D_\mathsf{tr}$.
Let $\text{Pr}_{D_\mathsf{tr}}$ and $\text{Pr}_{\text{\textbackslash} {D_\mathsf{tr}}}$ be the conditional probabilities of the members and non-members of ${D_\mathsf{tr}}$, respectively.
Hence, the expected gain of the attack model for the above setting is given by:
\begin{align}\label{exp_gain}
G^{\theta}(h)&=\underset{\substack{(\mathbf{x},y)\\ \sim \text{Pr}_{D_\mathsf{tr}}}}{\mathbb{E}} [\text{log}(h(F))]+\underset{\substack{(\mathbf{x},y)\\ \sim \text{Pr}_{\text{\textbackslash} D_\mathsf{tr}}}}{\mathbb{E}} [\text{log}(1-h(F))]
\end{align}
In practice, the adversary knows only a finite set of the members $D$ and non-members $D'^A$ required to train $h$, hence computes the above gain empirically as:
\begin{align}\label{emp_gain}
G^{\theta}_{D^A, D'^A}(h)= \sum_{\substack{(\mathbf{x},y)\\ \in D^A}} \frac{\text{log}(h(F))}{|D^A|} + \sum_{\substack{(\mathbf{x},y)\\ \in D'^A}} \frac{\text{log}(1-h(F))}{|D'^A|}
\end{align}
Finally, the adversary solves for $h^*$ that maximizes~\eqref{emp_gain}.

% !TEX root = main.tex

\section{Our Proposed Defense: DMP}\label{dmp}

Now, we present our defense \emph{Distillation For Membership Privacy (DMP)},
which is  motivated by the poor membership privacy-utility tradeoffs provided by existing MIA defenses (\S~\ref{related}).
First, we give an intuition behind DMP and detail the DMP training. Finally, to achieve the desired tradeoffs, we give a criterion to tune the selection or generation (e.g., using GANs) of reference data used in DMP.

\paragraphb{Notations. }\label{dmp:notations}
$D_\textsf{tr}$ is a \emph{private} training data. 
An ML model trained on $D_\textsf{tr}$ without any privacy protections is called \emph{unprotected} model, denoted by $\theta_\textsf{up}$. 
An ML model is called \emph{protected} model, denoted by $\theta_\textsf{p}$, if it protects $D_\textsf{tr}$ from MIAs. 
For knowledge transfer, DMP uses an \emph{unlabeled and possibly private reference dataset} which is \emph{disjoint} from $D_\textsf{tr}$; as the reference data is unlabeled, we denote it by $X_\mathsf{ref}$.
We denote the soft label of $\theta$ on $\mathbf{x}$, i.e., $\theta(\mathbf{x})$, by $\theta^\mathbf{x}$.

\paragraphb{Main intuition of DMP. }\label{dmp:intuition}
\cite{sablayrolles2019white} show that $\theta$ trained on a sample $z$ (short for $(\mathbf{x},y)$) provides $(\epsilon,\delta)$ membership privacy to $z$ if the expected loss of the models not trained on $z$ is $\epsilon$-close to the loss of $\theta$ on $z$, with probability at least $1-\delta$.
They assume a posterior distribution of the parameters trained on a given data $D=\{z_1,..,z_n\}$ to be: 
\begin{equation}\label{eq:post_assumption}
\mathbb{P}(\theta|z_1,...,z_n)\propto \text{exp}(\sum^n_{i=1} \ell(\theta,z_i))
\end{equation}
Consider a neighboring dataset $D'=\{z_1,..,z'_j,..,z_n\}$ of $D$, which is obtained by modifying at most one sample of $D$~\cite{ding2018detecting}. 
\cite{sablayrolles2019white} show that, to provide membership privacy to $z_j$, the log of the ratio of probabilities of obtaining the same $\theta$ from $D$ and $D'$ should be bounded, i.e., \eqref{eq:prob_ratio} should be bounded.
\begin{align}\label{eq:prob_ratio}
\text{log}\Big|\frac{\mathbb{P}(\theta|D)}{\mathbb{P}(\theta|D')}\Big| = |\ell(\theta,z_j)-\ell(\theta,z'_j)|
\end{align}

\eqref{eq:prob_ratio} implies that, if $\theta$ was indeed trained on $z_j$, then to provide membership privacy to $z_j$, the loss of $\theta$ on $z_j$ should be same as the loss on any non-member sample $z'_j$.

\emph{DMP is a strong meta-regularization} technique built on this intuition. DMP aims to protect its target models against the membership inference attacks that exploit the gap between the target model's losses on the members and non-members, by reducing the gap.

DMP achieves this via knowledge transfer and restricts the direct access of $\theta_\textsf{p}$ to the private $D_\mathsf{tr}$, which significantly reduces the membership information leakage to $\theta_\textsf{p}$.
However, unlike existing knowledge transfer, DMP proposes an entropy-based criterion to select/generate $X_\mathsf{ref}$. Simply put, soft labels of the unprotected model $\theta_\mathsf{up}$ on $X_\mathsf{ref}$ should have low entropy and the $X_\mathsf{ref}$ should be far from decision boundaries of $\theta_\mathsf{up}$, i.e., far from $D_\mathsf{tr}$, in the input feature space.
\emph{Intuitively, such samples are easy to classify and none of the members of $D_\textsf{tr}$ significantly affects their predictions, and therefore, these predictions do not leak membership information of any particular member.}

\begin{figure}
\centering
\includegraphics[scale=.8]{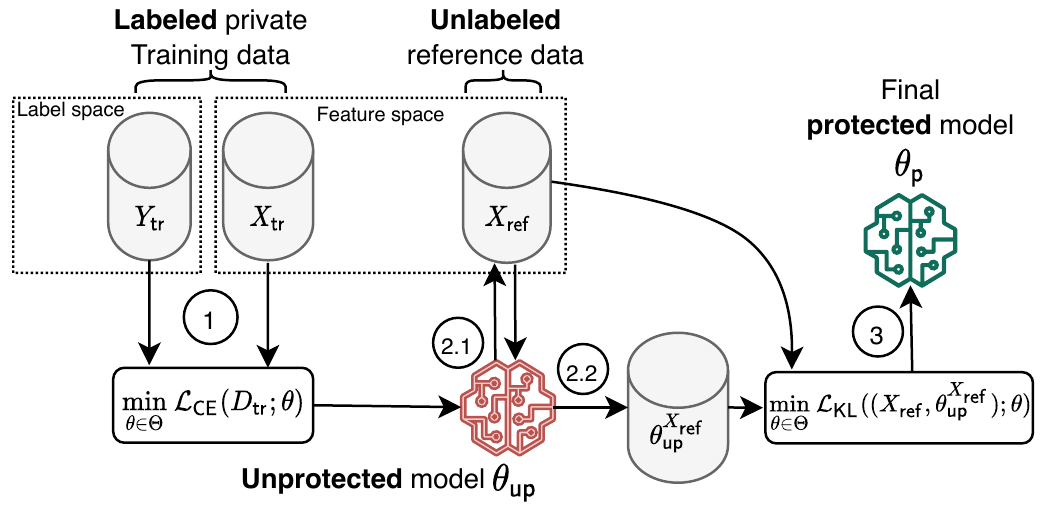}
% \vspace*{-1em}
\caption{\emph{\textbf{D}istillation for \textbf{M}embership \textbf{P}rivacy} (DMP) defense. (1) In \emph{pre-distillation} phase, DMP trains an unprotected model $\theta_\mathsf{up}$ on the private training data without any privacy protection. (2.1) In \emph{distillation} phase, DMP uses $\theta_\mathsf{up}$ to select/generate appropriate reference data $X_\mathsf{ref}$ that minimizes membership privacy leakage. (2.2) Then, DMP transfers the knowledge of $\theta_\mathsf{up}$ by computing predictions of $\theta_\mathsf{up}$ on $X_\mathsf{ref}$, denoted by $\theta^{X_\mathsf{ref}}_\mathsf{up}$. (3) In \emph{post-distillation} phase, DMP trains the final protected model $\theta_\mathsf{p}$ on $(X_\mathsf{ref},\theta^{X_\mathsf{ref}}_\mathsf{up})$.}
\label{fig:dml_blocks}
\vspace*{-1em}
\end{figure}

\paragraphb{Details of the DMP technique.}\label{dmp:description}
We now detail the three phases of our DMP defense depicted in Figure~\ref{fig:dml_blocks}.
In \emph{pre-distillation phase} (step (1) in Figure~\ref{fig:dml_blocks}), DMP trains $\theta_\textsf{up}$ on the private training data, $D_\textsf{tr}$, using standard SGD optimizer, e.g., Adam.
Such unprotected $\theta_\textsf{up}$ is highly susceptible to MIA due to large generalization error, i.e., difference between train and test accuracies~\cite{shokri2017membership,yeom2018privacy}.

Next, in \emph{distillation phase} (step (2.1) in Figure~\ref{fig:dml_blocks}), DMP obtains $X_\textsf{ref}$ required to transfer the knowledge of $\theta_\textsf{up}$ in $\theta_\textsf{p}$. Note that, $X_\textsf{ref}$ is \emph{unlabeled} and cannot be used directly for any learning.
Then, we compute soft labels of $X_\textsf{ref}$, i.e., $\theta^{X_\mathsf{ref}}_\mathsf{up}=\theta_\textsf{up}(X_\textsf{ref})$ (step (2.2) in Figure~\ref{fig:dml_blocks}).
There are two key factors of the distillation phase that allow us to tune DMP and achieve the desired privacy-utility tradeoffs.
First, the lower the entropy of predictions $\theta^{X_\mathsf{ref}}_\mathsf{up}$, the lower the membership leakage through $X_\mathsf{ref}$ and vice-versa. Such low entropy predictions are characteristics of the members of $D_\textsf{tr}$, however, non-members with low entropy can be obtained (or generated using GANs~\cite{micaelli2019zero}) due to large input feature space.
Second, using higher softmax temperatures to compute $\theta^{X_\mathsf{ref}}_\mathsf{up}$ reduces membership leakage, but may reduce accuracy of the final model, and vice-versa.

Finally, in \emph{Post-distillation phase} (step (3) in Figure~\ref{fig:dml_blocks}), DMP trains a protected model $\theta_\textsf{p}$ on $(X_\textsf{ref},\theta^{X_\mathsf{ref}}_\mathsf{up})$ using Kullback-Leibler divergence loss defined in~\eqref{kld_datum}. 
In~\eqref{kld_datum}, $\overline{\mathbf{y}}$ is the target soft label.
The final $\theta_\textsf{p}$ is obtained by solving~\eqref{kld_emprical_risk_min}.

\vspace{-.5em}
\begin{align}
\label{kld_datum}
\mathcal{L}_{\scaleto{\textsf{KL}}{4pt}}(\mathbf{x},\overline{\mathbf{y}})&= \sum^{\mathbf{c}-1}_{i=0}\overline{\mathbf{y}}_i\ \text{log}\Big(\frac{\overline{\mathbf{y}}_i}{\theta_\mathsf{p}(\mathbf{x})_i} \Big)\\
\label{kld_emprical_risk_min}
\theta_\textsf{p}=\underset{\theta}{\text{argmin}}&\ \frac{1}{|X_\textsf{ref}|}\sum_{(\mathbf{x},\overline{\mathbf{y}})\in(X_{\mathsf{ref}},\theta^{X_\mathsf{ref}}_\mathsf{up})} \mathcal{L}_{\scaleto{\textsf{KL}}{4pt}}(\mathbf{x},\overline{\mathbf{y}})
\end{align}

Due to KL-divergence loss in~\eqref{kld_emprical_risk_min}, the resulting model, $\theta_\textsf{p}$, perfectly learns the behavior of $\theta_\textsf{up}$ on the $X_\mathsf{ref}$.
Furthermore, $X_\mathsf{ref}$ being a representative non-member data, i.e., test data, we expect that the test accuracies of $\theta_\textsf{p}$ and $\theta_\textsf{up}$ are close, and that the final DMP models will not suffer significant accuracy reductions~\cite{ba2014deep,romero2014fitnets}.
\\

\paragraphb{Fine-tuning the DMP defense.}\label{dmp:tune}
As mentioned before, the appropriate choice of reference data $X_\mathsf{ref}$ is important to achieve the desired privacy-utility tradeoffs in DMP.
In this section, we show that $X_\mathsf{ref}$ with low entropy predictions of unprotected model $\theta_\mathsf{up}$ strengthens membership privacy and derive an entropy-based criterion to select/generate 
$X_\mathsf{ref}$.

\begin{proposition}\label{prop:entropy}
Consider $\theta_\mathsf{up}$ trained on a private $D_\mathsf{tr}$. Then, the membership leakage about $D_\mathsf{tr}$ through predictions $\theta_\mathsf{up}(X_\mathsf{ref})$ can be reduced by selecting/generating $X_\mathsf{ref}$ that are far from $D_\mathsf{tr}$ in input feature space with respect to some $L_p$ distance and whose predictions, $\theta_\mathsf{up}(X_\mathsf{ref})$, have low entropies.
\end{proposition}

\paragraphe{Sketch of proof of Proposition~\ref{prop:entropy}.}
Due to space limitations, we defer the detailed proof to Appendix and provide its sketch here.
Consider two training datasets $D_\textsf{tr}$ and $D'_\textsf{tr}$ such that $D'_\textsf{tr}\leftarrow D_\textsf{tr}-z$, and $X_\textsf{ref}$.
Then, the log of the ratio of the posterior probabilities of learning the exact same parameters $\theta_\textsf{p}$ using DMP is given by \eqref{eq:obj_ratio}.
Observe that, $\mathcal{R}$ is an extension of \eqref{eq:prob_ratio} to the setting of DMP, where $\theta_\mathsf{p}$ is trained via the knowledge transferred using $(X_\textsf{ref},\theta^{X_\textsf{ref}}_\textsf{up})$, instead of directly training on $D_\textsf{tr}$.
\cite{sablayrolles2019white} argue to reduce this ratio to improve membership privacy.
Hence, we want to obtain $X_\mathsf{ref}$ which reduces $\mathcal{R}$ when $D_\textsf{tr}$, $D'_\textsf{tr}$, and $\theta_\mathsf{p}$ are kept constant.
We note that, although similar in appearance to differential privacy, $\mathcal{R}$ is defined only for the given private dataset, $D_\textsf{tr}$.
\begin{equation}\label{eq:obj_ratio}
\mathcal{R}=\Big|\text{log}\ \Big({\text{Pr}(\theta_\textsf{p}|D_\textsf{tr},X_\textsf{ref})}/{\text{Pr}(\theta_\textsf{p}|D'_\textsf{tr},X_\textsf{ref})}\Big)\Big|
\end{equation}

Next, we modify $\mathcal{R}$ as:
\begin{align}
\label{eq:obj_ratio1}
& \mathcal{R}= \Big|-\frac{1}{T}\sum_{\mathbf{x}\in X_\textsf{ref}} \mathcal{L}_{\scaleto{\textsf{KL}}{4pt}}((\mathbf{x},\theta^{\mathbf{x}}_\textsf{up});\theta_\textsf{p}) - \mathcal{L}_{\scaleto{\textsf{KL}}{4pt}}((\mathbf{x},\theta'^{\mathbf{x}}_\textsf{up});\theta_\textsf{p})\Big| \\
\label{eq:obj_ratio2}
&\leq \frac{1}{T} \sum_{\mathbf{x}\in X_\textsf{ref}}\Big| \mathcal{L}_{\scaleto{\textsf{KL}}{4pt}}(\theta^{\mathbf{x}}_\textsf{up}\Vert\theta^{\mathbf{x}}_\textsf{p}) - \mathcal{L}_{\scaleto{\textsf{KL}}{4pt}}(\theta'^{\mathbf{x}}_\textsf{up}\Vert\theta^{\mathbf{x}}_\textsf{p})\Big|
\end{align}

\noindent where $\theta_\textsf{up}$ and $\theta'_\textsf{up}$ are trained on $D_\textsf{tr}$ and $D'_\textsf{tr}$, respectively.
Note that, \eqref{eq:obj_ratio1} holds due to the assumption in \eqref{eq:post_assumption} and the KL-divergence loss used to train $\theta_\mathsf{p}$ in DMP.
 % minimizes   between predictions of $\theta_\textsf{p}$ and $\theta_\textsf{up}$ on the reference data.
\eqref{eq:obj_ratio2} follows from \eqref{eq:obj_ratio1} because $|a+b|\leq|a|+|b|$.
Therefore, minimizing \eqref{eq:obj_ratio2} implies minimizing \eqref{eq:obj_ratio}.
Thus, to improve membership privacy due to $\theta_\mathsf{p}$, $X_\mathsf{ref}$ is obtained by solving~\eqref{eq:ref_obj}.
\begin{align}
\label{eq:ref_obj}
X^*_\textsf{ref}=\underset{X_\textsf{ref}\in X}{\text{argmin}}\Big(\frac{1}{T}\sum_{\mathbf{x}\in X_\textsf{ref}}  \big|\mathcal{L}_{\scaleto{\textsf{KL}}{4pt}}&(\theta^{\mathbf{x}}_\textsf{up}\Vert\theta^{\mathbf{x}}_\textsf{p}) -\mathcal{L}_{\scaleto{\textsf{KL}}{4pt}}(\theta'^{\mathbf{x}}_\textsf{up}\Vert\theta^{\mathbf{x}}_\textsf{p})\big|\Big)
\end{align}

The objective of \eqref{eq:ref_obj} is minimized when $\theta^{\mathbf{x}}_\mathsf{up} = \theta'^{\mathbf{x}}_\mathsf{up}\ \ \forall\mathbf{x}\in X_\mathsf{ref}$ and is very intuitive: It implies that, $z$ (i.e., $D_\mathsf{tr}-D'_\mathsf{tr}$) enjoys stronger membership privacy when the reference data, $X_\mathsf{ref}$, are such that \emph{the distributions of outputs of $\theta_\mathsf{up}$ and $\theta'_\mathsf{up}$ on $X_\mathsf{ref}$   are not affected by the presence of $z$ in $D_\mathsf{tr}$}.

Next, we simplify \eqref{eq:ref_obj} by replacing $\mathcal{L}_{\scaleto{\textsf{KL}}{4pt}}$ with closely related cross-entropy loss $\mathcal{L}_{\scaleto{\textsf{CE}}{4pt}}$. 
This simplification can be easily validated using $X_\mathsf{ref}$ whose ground truth labels are known.
Specifically, we randomly sample  $D_\mathsf{tr}$ and $X_\mathsf{ref}$ from Purchase100 dataset, and compute $\theta_\mathsf{up}$ and $\theta_\mathsf{p}$ using DMP. 
Next, for some $z\in D_\mathsf{tr}$, we train $\theta'_\mathsf{up}$ on $D'_\mathsf{tr}$. 
Then, for each $\mathbf{x}\in X_\mathsf{ref}$, we compute $\Delta\mathcal{L}_{\scaleto{\textsf{KL}}{4pt}}$ as in~\eqref{eq:ref_obj} and use the available ground truth label of $\mathbf{x}$ to compute $\Delta\mathcal{L}_{\scaleto{\textsf{CE}}{4pt}}$.
Finally, we show that $\Delta\mathcal{L}_{\scaleto{\textsf{KL}}{4pt}}$ and $\Delta\mathcal{L}_{\scaleto{\textsf{CE}}{4pt}}$ are strongly correlated for all $z\in D_\mathsf{tr}$.

Next, we use the linear approximation given by~\cite{koh2017understanding} for the difference in $\mathcal{L}_{\scaleto{\textsf{CE}}{4pt}}$ of a pair of models trained with and without a sample to simplify \eqref{eq:ref_obj}.
Then the result of Proposition~\ref{prop:entropy} follows after a few simple mathematical manipulations.

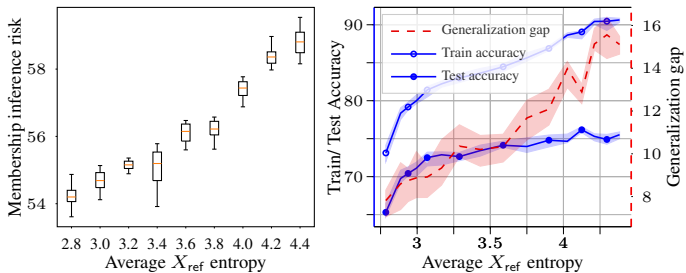
\begin{figure}[t!]
\centering
\hspace*{-2em}
\begin{tabular}{cc}

\subfloat{% This file was created by matplotlib2tikz v0.7.4.
\begin{tikzpicture}

\definecolor{color0}{rgb}{1,0.498039215686275,0.0549019607843137}
\definecolor{color1}{rgb}{0.172549019607843,0.627450980392157,0.172549019607843}
\pgfmathsetlengthmacro\MajorTickLength{
      \pgfkeysvalueof{/pgfplots/major tick length} * 0.15
    }
\begin{axis}[
height=4.5cm,
width=5cm,
tick align=outside,
tick pos=left,
x grid style={white!69.01960784313725!black},
y grid style={white!69.01960784313725!black},
xticklabel style={font=\tiny, rotate=0.0},
yticklabel style={font=\tiny},
x grid style={white!69.01960784313725!black},
xlabel={Average $X_\mathsf{ref}$ entropy},
xlabel style={font=\scriptsize, at={(axis description cs:0.5,-0.09)},anchor=north},
xmin=0.5, xmax=9.5,
xtick style={color=black, font=\tiny},
xtick={1,2,3,4,5,6,7,8,9},
xticklabels={2.8,3.0,3.2,3.4,3.6,3.8,4.0,4.2,4.4},
y grid style={white!69.01960784313725!black},
scaled y ticks={real:0.01},
ytick scale label code/.code={},
ylabel={Membership inference risk},
ylabel style={font=\scriptsize, at={(axis description cs:-0.09,.5)},anchor=south},
ymin=0.533157467532468, ymax=0.598336038961039,
ytick style={color=black, font=\tiny},
major tick length=\MajorTickLength,
]
\addplot [black]
table {%
0.85 0.540635146103896
1.15 0.540635146103896
1.15 0.54398336038961
0.85 0.54398336038961
0.85 0.540635146103896
};
\addplot [black]
table {%
1 0.540635146103896
1 0.53612012987013
};
\addplot [black]
table {%
1 0.54398336038961
1 0.548701298701299
};
\addplot [black]
table {%
0.925 0.53612012987013
1.075 0.53612012987013
};
\addplot [black]
table {%
0.925 0.548701298701299
1.075 0.548701298701299
};
\addplot [black]
table {%
1.85 0.544795048701299
2.15 0.544795048701299
2.15 0.549247402597403
1.85 0.549247402597403
1.85 0.544795048701299
};
\addplot [black]
table {%
2 0.544795048701299
2 0.541193181818182
};
\addplot [black]
table {%
2 0.549247402597403
2 0.551339285714286
};
\addplot [black]
table {%
1.925 0.541193181818182
2.075 0.541193181818182
};
\addplot [black]
table {%
1.925 0.551339285714286
2.075 0.551339285714286
};
\addplot [black]
table {%
2.85 0.550468506493506
3.15 0.550468506493506
3.15 0.552637743506494
2.85 0.552637743506494
2.85 0.550468506493506
};
\addplot [black]
table {%
3 0.550468506493506
3 0.548904220779221
};
\addplot [black]
table {%
3 0.552637743506494
3 0.553571428571429
};
\addplot [black]
table {%
2.925 0.548904220779221
3.075 0.548904220779221
};
\addplot [black]
table {%
2.925 0.553571428571429
3.075 0.553571428571429
};
\addplot [black]
table {%
3.85 0.546875
4.15 0.546875
4.15 0.555348701298701
3.85 0.555348701298701
3.85 0.546875
};
\addplot [black]
table {%
4 0.546875
4 0.539163961038961
};
\addplot [black]
table {%
4 0.555348701298701
4 0.557832792207792
};
\addplot [black]
table {%
3.925 0.539163961038961
4.075 0.539163961038961
};
\addplot [black]
table {%
3.925 0.557832792207792
4.075 0.557832792207792
};
\addplot [black]
table {%
4.85 0.55861112012987
5.15 0.55861112012987
5.15 0.563666801948052
4.85 0.563666801948052
4.85 0.55861112012987
};
\addplot [black]
table {%
5 0.55861112012987
5 0.556
};
\addplot [black]
table {%
5 0.563666801948052
5 0.564732142857143
};
\addplot [black]
table {%
4.925 0.556
5.075 0.556
};
\addplot [black]
table {%
4.925 0.564732142857143
5.075 0.564732142857143
};
\addplot [black]
table {%
5.85 0.56045
6.15 0.56045
6.15 0.56447849025974
5.85 0.56447849025974
5.85 0.56045
};
\addplot [black]
table {%
6 0.56045
6 0.556209415584416
};
\addplot [black]
table {%
6 0.56447849025974
6 0.565746753246753
};
\addplot [black]
table {%
5.925 0.556209415584416
6.075 0.556209415584416
};
\addplot [black]
table {%
5.925 0.565746753246753
6.075 0.565746753246753
};
\addplot [black]
table {%
6.85 0.572161525974026
7.15 0.572161525974026
7.15 0.576247970779221
6.85 0.576247970779221
6.85 0.572161525974026
};
\addplot [black]
table {%
7 0.572161525974026
7 0.568790584415584
};
\addplot [black]
table {%
7 0.576247970779221
7 0.577719155844156
};
\addplot [black]
table {%
6.925 0.568790584415584
7.075 0.568790584415584
};
\addplot [black]
table {%
6.925 0.577719155844156
7.075 0.577719155844156
};
\addplot [black]
table {%
7.85 0.581748051948052
8.15 0.581748051948052
8.15 0.585122646103896
7.85 0.585122646103896
7.85 0.581748051948052
};
\addplot [black]
table {%
8 0.581748051948052
8 0.579748376623377
};
\addplot [black]
table {%
8 0.585122646103896
8 0.589691558441558
};
\addplot [black]
table {%
7.925 0.579748376623377
8.075 0.579748376623377
};
\addplot [black]
table {%
7.925 0.589691558441558
8.075 0.589691558441558
};
\addplot [black]
table {%
8.85 0.584872159090909
9.15 0.584872159090909
9.15 0.590959821428571
8.85 0.590959821428571
8.85 0.584872159090909
};
\addplot [black]
table {%
9 0.584872159090909
9 0.581574675324675
};
\addplot [black]
table {%
9 0.590959821428571
9 0.595373376623377
};
\addplot [black]
table {%
8.925 0.581574675324675
9.075 0.581574675324675
};
\addplot [black]
table {%
8.925 0.595373376623377
9.075 0.595373376623377
};
\addplot [color0]
table {%
0.85 0.542002435064935
1.15 0.542002435064935
};
\addplot [color1, dashed, mark size=3, mark options={solid}]
table {%
1 0.542351785714286
};
\addplot [color0]
table {%
1.85 0.546875
2.15 0.546875
};
\addplot [color1, dashed, mark size=3, mark options={solid}]
table {%
2 0.54743948051948
};
\addplot [color0]
table {%
2.85 0.551542207792208
3.15 0.551542207792208
};
\addplot [color1, dashed, mark size=3, mark options={solid}]
table {%
3 0.551421136363636
};
\addplot [color0]
table {%
3.85 0.551948051948052
4.15 0.551948051948052
};
\addplot [color1, dashed, mark size=3, mark options={solid}]
table {%
4 0.550871558441558
};
\addplot [color0]
table {%
4.85 0.56148538961039
5.15 0.56148538961039
};
\addplot [color1, dashed, mark size=3, mark options={solid}]
table {%
5 0.560972987012987
};
\addplot [color0]
table {%
5.85 0.562195616883117
6.15 0.562195616883117
};
\addplot [color1, dashed, mark size=3, mark options={solid}]
table {%
6 0.56212
};
\addplot [color0]
table {%
6.85 0.574370941558441
7.15 0.574370941558441
};
\addplot [color1, dashed, mark size=3, mark options={solid}]
table {%
7 0.574690032467532
};
\addplot [color0]
table {%
7.85 0.583603896103896
8.15 0.583603896103896
};
\addplot [color1, dashed, mark size=3, mark options={solid}]
table {%
8 0.583848961038961
};
\addplot [color0]
table {%
8.85 0.588085551948052
9.15 0.588085551948052
};
\addplot [color1, dashed, mark size=3, mark options={solid}]
table {%
9 0.587967532467532
};
\end{axis}

\end{tikzpicture}}
\hspace*{-.5em}
\subfloat{% This file was created by matplotlib2tikz v0.7.4.
\begin{tikzpicture}

\definecolor{color0}{rgb}{0.976470588235294,0.450980392156863,0.0235294117647059}
\definecolor{color1}{rgb}{0.0823529411764706,0.690196078431373,0.101960784313725}
\pgfmathsetlengthmacro\MajorTickLength{
      \pgfkeysvalueof{/pgfplots/major tick length} * 0.15
    }
\begin{axis}[
height=4.5cm,
width=5cm,
grid=both,
xticklabel style = {font=\tiny},
yticklabel style = {font=\tiny},
legend cell align={left},
legend style={at={(0.03,0.97)}, nodes={scale=0.9}, fill opacity=0.8, anchor=north west, font=\tiny, draw=white!80.0!black},
tick align=outside,
tick pos=left,
minor xtick={2.75,3,...,4.25},
minor ytick={65,70,...,90},
x grid style={white!69.01960784313725!black},
xlabel={Average $X_\mathsf{ref}$ entropy},
xlabel style={font=\scriptsize, at={(axis description cs:0.5,-0.09)},anchor=north},
xmajorgrids,
xmin=2.70595481395721, xmax=4.46344015598297,
xtick style={color=black},
y grid style={white!69.01960784313725!black},
ylabel={Train/ Test Accuracy},
ylabel style={font=\scriptsize, at={(axis description cs:-0.08,.5)},anchor=south},
ymajorgrids,
ymin=63.4048509873863, ymax=92.2884518455335,
ytick style={color=black},
major tick length=\MajorTickLength,
separate axis lines,
% first x axis line style=red,
% second x axis line style={ultra thick, dashed},
first y axis line style={semithick, blue},
second y axis line style={semithick, red!89.80392156862746!black, dashed},
]
\addlegendimage{/pgfplots/refstyle=gen_gap} \addlegendentry{Generalization gap}
\path [draw=blue, fill=blue, opacity=0.2]
(axis cs:2.78584051132202,71.9751602564103)
--(axis cs:2.78584051132202,74.1386217948718)
--(axis cs:2.8870632648468,79.0564903846154)
--(axis cs:2.93812131881714,79.8076923076923)
--(axis cs:2.99454736709595,80.7491987179487)
--(axis cs:3.06869053840637,81.8810096153846)
--(axis cs:3.16676163673401,82.5620993589744)
--(axis cs:3.28857612609863,83.5336538461538)
--(axis cs:3.43108367919922,84.2447916666667)
--(axis cs:3.58782410621643,84.8257211538462)
--(axis cs:3.75045156478882,86.3080929487179)
--(axis cs:3.90056324005127,87.2095352564103)
--(axis cs:4.02654266357422,88.9322916666667)
--(axis cs:4.12503385543823,89.2127403846154)
--(axis cs:4.21155023574829,90.7451923076923)
--(axis cs:4.29680204391479,90.9755608974359)
--(axis cs:4.38355445861816,90.8553685897436)
--(axis cs:4.38355445861816,90.3245192307692)
--(axis cs:4.38355445861816,90.3245192307692)
--(axis cs:4.29680204391479,89.2528044871795)
--(axis cs:4.21155023574829,90.0540865384615)
--(axis cs:4.12503385543823,88.1510416666667)
--(axis cs:4.02654266357422,88.3413461538462)
--(axis cs:3.90056324005127,86.6085737179487)
--(axis cs:3.75045156478882,85.2163461538462)
--(axis cs:3.58782410621643,83.8341346153846)
--(axis cs:3.43108367919922,83.193108974359)
--(axis cs:3.28857612609863,82.3818108974359)
--(axis cs:3.16676163673401,81.7808493589744)
--(axis cs:3.06869053840637,80.478766025641)
--(axis cs:2.99454736709595,79.5973557692308)
--(axis cs:2.93812131881714,78.7760416666667)
--(axis cs:2.8870632648468,77.5140224358974)
--(axis cs:2.78584051132202,71.9751602564103)
--cycle;

\path [draw=blue, fill=blue, opacity=0.2]
(axis cs:2.78584051132202,64.7177419354839)
--(axis cs:2.78584051132202,66.491935483871)
--(axis cs:2.8870632648468,70.8467741935484)
--(axis cs:2.93812131881714,71.2903225806452)
--(axis cs:2.99454736709595,71.8951612903226)
--(axis cs:3.06869053840637,73.2258064516129)
--(axis cs:3.16676163673401,73.2661290322581)
--(axis cs:3.28857612609863,73.1854838709677)
--(axis cs:3.43108367919922,74.1532258064516)
--(axis cs:3.58782410621643,74.6774193548387)
--(axis cs:3.75045156478882,75.1209677419355)
--(axis cs:3.90056324005127,75.4838709677419)
--(axis cs:4.02654266357422,75.6048387096774)
--(axis cs:4.12503385543823,76.5725806451613)
--(axis cs:4.21155023574829,75.6451612903226)
--(axis cs:4.29680204391479,75.5645161290323)
--(axis cs:4.38355445861816,75.8870967741936)
--(axis cs:4.38355445861816,75)
--(axis cs:4.38355445861816,75)
--(axis cs:4.29680204391479,74.2741935483871)
--(axis cs:4.21155023574829,74.7177419354839)
--(axis cs:4.12503385543823,75.8064516129032)
--(axis cs:4.02654266357422,74.4758064516129)
--(axis cs:3.90056324005127,73.9112903225806)
--(axis cs:3.75045156478882,73.1854838709677)
--(axis cs:3.58782410621643,73.8709677419355)
--(axis cs:3.43108367919922,72.7016129032258)
--(axis cs:3.28857612609863,72.0967741935484)
--(axis cs:3.16676163673401,71.5322580645161)
--(axis cs:3.06869053840637,71.4112903225806)
--(axis cs:2.99454736709595,70.0403225806452)
--(axis cs:2.93812131881714,70.0806451612903)
--(axis cs:2.8870632648468,69.1935483870968)
--(axis cs:2.78584051132202,64.7177419354839)
--cycle;

\addplot [semithick, mark=o, mark size=1, mark repeat=2, blue]
table {%
2.78584051132202 73.1169871794872
2.8870632648468 78.3553685897436
2.93812131881714 79.1716746794872
2.99454736709595 79.9829727564103
3.06869053840637 81.4152644230769
3.16676163673401 82.216546474359
3.28857612609863 82.997796474359
3.43108367919922 83.6288060897436
3.58782410621643 84.4701522435898
3.75045156478882 85.6470352564103
3.90056324005127 86.8840144230769
4.02654266357422 88.6368189102564
4.12503385543823 89.0625
4.21155023574829 90.4346955128205
4.29680204391479 90.4597355769231
4.38355445861816 90.635016025641
};
\addlegendentry{Train accuracy}
\addplot [semithick, mark=*, mark size=1, mark repeat=2, blue]
table {%
2.78584051132202 65.3024193548387
2.8870632648468 69.758064516129
2.93812131881714 70.4233870967742
2.99454736709595 71.1088709677419
3.06869053840637 72.5
3.16676163673401 72.883064516129
3.28857612609863 72.6411290322581
3.43108367919922 73.4274193548387
3.58782410621643 74.133064516129
3.75045156478882 73.9717741935484
3.90056324005127 74.7983870967742
4.02654266357422 74.6572580645161
4.12503385543823 76.1693548387097
4.21155023574829 75.3024193548387
4.29680204391479 74.8991935483871
4.38355445861816 75.5241935483871
};
\addlegendentry{Test accuracy}
\end{axis}

\begin{axis}[
height=4.5cm,
width=5cm,
xticklabel style = {font=\tiny},
yticklabel style = {font=\tiny},
% legend cell align={left},
% legend style={at={(.53,.15)}, anchor=north west, font=\scriptsize, draw=white!80.0!black},
axis y line=right,
tick align=outside,
x grid style={white!69.01960784313725!black},
xmin=2.70595481395721, xmax=4.46344015598297,
xtick pos=left,
xtick style={color=black},
y grid style={white!69.01960784313725!black},
ylabel={Generalization gap},
ylabel style={font=\scriptsize, at={(axis description cs:1.24,.5)},anchor=south},
ymin=6.58965635339124, ymax=16.8449357681969,
ytick pos=right,
ytick style={color=black},
major tick length=\MajorTickLength,
separate axis lines,
% first y axis line style=red,
second y axis line style={red, dashed},
]
\path [draw=red!89.80392156862746!black, fill=red!89.80392156862746!black, opacity=0.2]
(axis cs:2.78584051132202,7.05580541770058)
--(axis cs:2.78584051132202,8.29908498759305)
--(axis cs:2.8870632648468,9.58068393300249)
--(axis cs:2.93812131881714,9.31483922663358)
--(axis cs:2.99454736709595,10.097898573201)
--(axis cs:3.06869053840637,9.5019773573201)
--(axis cs:3.16676163673401,10.6592483457403)
--(axis cs:3.28857612609863,11.3567514474773)
--(axis cs:3.43108367919922,10.8576948924731)
--(axis cs:3.58782410621643,10.9144308312655)
--(axis cs:3.75045156478882,12.5016154880066)
--(axis cs:3.90056324005127,13.0177962158809)
--(axis cs:4.02654266357422,14.3162608560794)
--(axis cs:4.12503385543823,13.2560483870968)
--(axis cs:4.21155023574829,15.8661600496278)
--(axis cs:4.29680204391479,16.3787867038875)
--(axis cs:4.38355445861816,15.4929823200993)
--(axis cs:4.38355445861816,14.6582919768404)
--(axis cs:4.38355445861816,14.6582919768404)
--(axis cs:4.29680204391479,14.4544173904053)
--(axis cs:4.21155023574829,14.8496303763441)
--(axis cs:4.12503385543823,12.3445900537634)
--(axis cs:4.02654266357422,13.0994106699752)
--(axis cs:3.90056324005127,11.3453137923904)
--(axis cs:3.75045156478882,10.6462598221671)
--(axis cs:3.58782410621643,9.61745244003309)
--(axis cs:3.43108367919922,9.42049214226633)
--(axis cs:3.28857612609863,9.85893558726221)
--(axis cs:3.16676163673401,8.53475237799834)
--(axis cs:3.06869053840637,7.95770006203475)
--(axis cs:2.99454736709595,7.95414598842018)
--(axis cs:2.93812131881714,7.76616780397022)
--(axis cs:2.8870632648468,7.39279621588089)
--(axis cs:2.78584051132202,7.05580541770058)
--cycle;

\addplot [semithick, dashed, red!89.80392156862746!black]
table {%
2.78584051132202 7.81456782464846
2.8870632648468 8.59730407361457
2.93812131881714 8.74828758271299
2.99454736709595 8.87410178866833
3.06869053840637 8.91526442307692
3.16676163673401 9.33348195822995
3.28857612609863 10.3566674421009
3.43108367919922 10.2013867349049
3.58782410621643 10.3370877274607
3.75045156478882 11.6752610628619
3.90056324005127 12.0856273263027
4.02654266357422 13.9795608457403
4.12503385543823 12.8931451612903
4.21155023574829 15.1322761579818
4.29680204391479 15.560542028536
4.38355445861816 15.1108224772539
};\label{gen_gap}
\end{axis}

\end{tikzpicture}}

\end{tabular}
\vspace*{-1em}
\caption{The lower the entropy of predictions of unprotected model on $X_\mathsf{ref}$, the higher the membership privacy.}
\label{fig:hypothesis_eval_1}
\vspace*{-1.75em}
\end{figure}

\paragraphe{Empirical verification of Proposition~\ref{prop:entropy}.}
We randomly pick $D_\mathsf{tr}$ of size 10k from Purhcase100 data and train $\theta_\mathsf{up}$. Then, we sort the rest of Purhcase100 data based on entropy of the predictions of $\theta_\mathsf{up}$ on the data. We form first $X_\mathsf{ref}$ using the first 10k data with the lowest entropies, second $X_\mathsf{ref}$ using the following 10k data, and so on. Finally we train multiple protected models, $\theta_\mathsf{p}$'s, using each of the $X_\mathsf{ref}$'s.
Figure \ref{fig:hypothesis_eval_1} (left) shows the increase in the MIA risk and Figure \ref{fig:hypothesis_eval_1} (right) shows the increase in the classification performance of $\theta_\textsf{p}$ with the increase in average entropy of the $X_\mathsf{ref}$ used.
This tradeoff is because, although the higher entropy predictions contain more useful information \cite{nayak2019zero,hinton2014distilling} and lead to high accuracy of $\theta_\mathsf{p}$, they also contain higher membership information about $D_\mathsf{tr}$ and lead to higher MIA risk.

% !TEX root = main.tex
\section{Experimental Setup}\label{exp_setup}

\subsection{Datasets and target model architectures}
We use four datasets and corresponding model architectures that are consistent with the previous works~\cite{shokri2017membership,nasr2019comprehensive,nasr2018machine,salem2019ml}.

\paragraphb{Purchase}~\cite{purchase} is a 100 class classification task with 197,324 binary feature vectors of length 600; each dimension corresponds to a product and its value states if corresponding customer purchased the product; the corresponding label represents the shopping habit of the customer.

\paragraphb{Texas}~\cite{texas} is dataset of patient records. It is a 100 class classification task with 67,300 binary feature vectors of length 6,170 each; each dimension corresponds to symptoms and its value states if corresponding patient has the symptom or not; the label represents the treatment given to the patient.
For Purchase and Texas we use fully connected (FC) networks.

\paragraphb{CIFAR10 and CIFAR100}~\cite{krizhevsky2009learning} are popular image classification datasets, each has size 50k and $32\times 32$ color images. We use Alexnet, DenseNet-12 (with 0.77M parameters), and DenseNet-19 (with 25.6M parameters) models for CIFAR100, and Alexnet for CIFAR10.
Following previous works, \emph{we measure the test accuracy of the target models as their utility}.

\paragraphb{Sizes of dataset splits. }
The dataset splits are given in Table~\ref{tab:data_sizes}. For Purchase and Texas tasks, we use $D_\mathsf{ref}$ of size 10k and \emph{select} $X_\mathsf{ref}$ of size 10k from the remaining data using our entropy-based criterion. For CIFAR datasets, we use $D_\mathsf{ref}$ of size 25k and due to small sizes of these datasets, use the entire remaining 25k data as $X_\mathsf{ref}$.
The `Attack training' (described shortly) column shows the MIA adversary's knowledge of members and non-members of $D_\mathsf{tr}$. Following all the previous works, we assume that the adversary knows 50\% of $D_\mathsf{tr}$.
Further experimental details are provided in Appendix.

\begin{table}[h]
\fontsize{8.5}{9}\selectfont{}
\begin{center}
\begin{tabular}{ |c|c|c|c|c| } 
\hline
\multirow{2}{*}{Dataset}& \multicolumn{2}{c|}{DMP training} & \multicolumn{2}{c|}{Attack training} \\ \cline{2-5}
& $|D_\textsf{tr}|$ & $|X_\textsf{ref}|$ &  $|D|$ & $|D'|$ \\ \hline
Purchase (P) & 10000 & 10000 & 5000 & 5000 \\
Texas (T) & 10000 & 10000 & 5000 & 5000 \\
CIFAR100 (C100) &  25000 & 25000 & 12500 & 8000 \\ 
CIFAR10 (C10)&  25000 & 25000 & 12500 & 8000 \\ \hline 
\end{tabular}
\vspace*{-.6em}
\caption{All the dataset splits are disjoint. $D$, $D'$ data are the members and non-members of $D_\textsf{tr}$ known to MIA adversary.}
\label{tab:data_sizes}

\end{center}
\vspace*{-2.5em}
\end{table}

\vspace{-.5em}
\subsection{Membership inference attacks}\label{setup:attacks}
We briefly review the four MIAs we use for evaluations. Following previous works, \emph{we use the accuracy of MIAs on target models as a measure of their membership privacy}.

\paragraphb{Bounded loss (BL) attack}~\cite{yeom2018privacy} decides membership using a threshold on the target model's loss on the target sample. When 0-1 loss is used, the attack accuracy is simply the difference in training and test accuracy of target model. We denote BL attack accuracy by $A_\mathsf{bl}$.

\paragraphb{NN attack}~\cite{salem2019ml} uses a \emph{shadow dataset} $d_s$ drawn from the same distribution as $D_\mathsf{tr}$. The attacker splits $d_s$ in $d'_s$ and $d''_s$, trains a \emph{shadow model} $\theta_s$ on $d'_s$, computes predictions of $\theta_s$ on $d'_s$ and $d''_s$, labels the predictions of $d'_s$ as members and that of $d''_s$ as non-members, and trains binary attack model on the predictions. We denote NN attack accuracy by $A_\textsf{nn}$. Due to their small sizes, DMP cannot be evaluated with CIFAR datasets, hence we omit NN attack evaluation for CIFAR datasets.

\paragraphb{NSH attacks}~\cite{nasr2019comprehensive} are similar to NN attacks. They concatenate various whitebox (e.g., model gradients) and/or blackbox (e.g., model loss, predictions) features of target model, while NN attack  uses only the target model predictions. We denote whitebox and blackbox NSH attack accuracies by $A_\textsf{wb}$ and $A_\textsf{bb}$, respectively. For NN and NSH attacks, we use the same attack models as the original works.

% !TEX root = main.tex
\begin{table}
\vspace*{-.6em}

\fontsize{8.5}{9}\selectfont{}
\begin{center}
\setlength{\extrarowheight}{0.02cm}

\begin{tabular}{ |c|c|c|c|c|c|c| } 
\hline

{Dataset} & \multicolumn{6}{c|}{\multirow{2}{*}{No defense}}  \\

and & \multicolumn{6}{c|}{} \\ \cline{2-7}

model & $E_\textsf{gen}$ & $A_\textsf{test}$ & $A_\textsf{wb}$ & $A_\textsf{bb}$ & $A_\textsf{bl}$ & $A_\textsf{nn}$ \\ \hline

P-FC & 24.0 & 76.0  & 77.1 & 76.8  & 63.1 & 60.5\\ \cline{1-7}

T-FC & 51.3 & 48.7 & 84.0 & 82.2 & 76.1 & 71.9\\ \cline{1-7}

C100-A & 63.2 & 36.8  & 90.3 & 91.3 & 81.8 & N/A\\ \cline{2-7}

C100-D12  & 33.8 & {65.2} & 72.2 & 71.8 & 67.5 & N/A \\ \cline{2-7}

C100-D19 &  34.4 & 65.5 & 82.3 & 81.6 & 68.1 & N/A  \\ \cline{1-7}

C10-A & 32.5 & 67.5 & 77.9 & 77.5 & 66.4 & N/A \\ \cline{1-7}

\end{tabular}
\vspace*{-.4em}
\caption{Models trained without any defenses have high test accuracies, $A_\mathsf{test}$, but their high generalization errors, $E_\mathsf{gen}$ (i.e., $A_\mathsf{train}-A_\mathsf{test}$) facilitate strong MIAs (\S~\ref{setup:attacks}). ``N/A'' means the attack is not evaluated due to lack of data. }
\label{table:baseline}
\end{center}
\vspace*{-1em}
\end{table}

\begin{table*}
\fontsize{8.5}{9}\selectfont{}
\begin{center}
\setlength{\extrarowheight}{0.02cm}
\vspace*{-.2em}
\begin{tabular}{ |c|c|c|c|c|c|c|c|c|c|c|c|c|c| } 
\hline

{Dataset} & \multicolumn{6}{c|}{Adversarial regularization (AdvReg)} & \multicolumn{7}{c|}{{DMP}} \\ \cline{2-14}

and & \multirow{2}{*}{$E_\textsf{gen}$} & \multirow{2}{*}{$A_\textsf{test}$} & \multicolumn{4}{c|}{Attack accuracy} & \multirow{2}{*}{$E_\textsf{gen}$} & \multirow{2}{*}{$A_\textsf{test}$} & \multirow{2}{*}{$A^{+}_\textsf{test}$} & \multicolumn{4}{c|}{Attack accuracy} \\ \cline{4-7}\cline{11-14}

model & & & $A_\textsf{wb}$ & $A_\textsf{bb}$ & $A_\textsf{bl}$ & $A_\textsf{nn}$ & & & & $A_\textsf{wb}$ & $A_\textsf{bb}$ & $A_\textsf{bl}$ & $A_\textsf{nn}$ \\ \hline
\hline

Purchase + FC & 9.7 &  56.5 & 55.8 & 55.4 & 54.9 & 50.1 &  10.1 & 74.1 & +\textbf{31.2}\% & 55.3 & 55.1 & 55.2 & 50.2  \\ \cline{1-14}

Texas + FC & 6.1 & 33.5 &  58.2 & 57.9 & 54.1 & 50.8 & 7.1 & 48.6 & +\textbf{45.1}\% & 55.3 & 55.4 & 53.6 & 50.0  \\ \cline{1-14}

CIFAR100 + Alexnet & 6.9 & 19.7 & 54.3 & 54.0 & 53.5 & N/A & 6.5 & {35.7} & +\textbf{81.2}\% & 55.7 & 55.6 & 53.3 & N/A \\ \cline{2-14}

CIFAR100 + DenseNet-12 & 5.5 &  26.5 &  51.4 & 51.3 & 52.8 & N/A & 3.6 & 63.1 & +\textbf{138.1}\% & 53.7 & 53.0 & 51.8 & N/A \\ \cline{2-14}

CIFAR100 + DenseNet-19 & 7.2 & 33.9 &  54.2 & 53.4 & 53.6 & N/A & 7.3 & 65.3 & +\textbf{92.6}\% & 54.7 & 54.4 & 53.7 & N/A \\ \cline{1-14}

CIFAR10 + Alexnet & 4.2 & 53.4 & 51.9 & 51.2 & 52.1 & N/A & 3.1 & 65.0 & +\textbf{21.7}\% & 51.3 & 50.6 & 51.6 & N/A \\ \cline{1-14}

\end{tabular}
\vspace*{-.6em}
\caption{Comparing test accuracy ($A_\mathsf{test}$) and generalization error ($E_\mathsf{gen}$) of DMP and Adversarial Regularization, for near-equal, low MIA risks (high membership privacy). 
$A^{+}_\textsf{test}$ shows \emph{the \% increase} in $A_\textsf{test}$ of DMP over Adversarial Regularization. 
}
\label{table:performance_comparison}
\end{center}
\vspace*{-1em}
\end{table*}

\section{Experiments}\label{exp}

\subsection{Comparison with regularization techniques} \label{exp:regularizations}
Regularization improves the generalization of ML models, and hence, reduce the MIA risk~\cite{shokri2017membership}.
Hence, we compare DMP with five regularization defeses, including state-of-the-art MIA defense\textemdash adeversarial regularization~\cite{nasr2018machine}.
In all tables, $E_\textsf{gen}$ is generalization error, i.e., ($A_\textsf{train}-A_\textsf{test}$), where $A_\textsf{train}$ and $A_\textsf{test}$ are train and test accuracies of the target model, respectively. 
$A^+_\mathsf{test}$ gives the \% increase in $A_\mathsf{test}$ due to DMP over the other regularizers.
$A_\textsf{wb}$, $A_\textsf{bb}$, $A_\textsf{bl}$, $A_\textsf{nn}$ are accuracies of various attacks discussed in the previous section.

Table~\ref{table:baseline} shows accuracies of models trained without any defense; CIFAR models have lower than state-of-the-art accuracies due to smaller training datasets.

\subsubsection{Comparison with adversarial regularization (AdvReg).} 
Table \ref{table:performance_comparison} compares $A_\mathsf{test}$ of DMP and AdvReg models, for similar MIA accuracies (i.e., membership privacy). As expected, these models also have similar $E_\mathsf{gen}$'s.
However, $A_\mathsf{test}$ of DMP models is significantly higher than AdvReg models; $A^+_\mathsf{test}$ column shows the \% increase in $A_\mathsf{test}$ due to DMP over AdvReg:
Accuracy improvements due to DMP over AdvReg are close to 100\% for CIFAR-100, and 20\% to 45\% for other datasets.
AdvReg uses accuracy of an MIA model to regularize and train its target models to fool the MIA model.
However, AdvReg allows its target models to directly access $D_\mathsf{tr}$. Hence, to effectively fool the MIA model, it puts relatively large weight on the regularization-loss term. This reduces the impact of the loss on main task and reduces the accuracy of AdvReg models.
DMP uses appropriate reference data to transfer the knowledge of $D_\mathsf{tr}$ to its target models without allowing them direct access.
Hence, DMP significantly outperforms AdvReg in terms of privacy-utility tradeoffs.

\begin{table}
\fontsize{8}{9}\selectfont{}

\begin{center}
\setlength{\extrarowheight}{0.03cm}
\hspace{-2em}

\begin{tabular}{ |c|c|c|c|c|c|c| } 
\hline
\multicolumn{7}{|c|}{{Purchase + FC} (DMP's $A_\textsf{test}$ = 74.1)} \\ \hline
{Regularizer} & {$E_\textsf{gen}$} & {$A_\textsf{test}$} & {$A^{+}_\textsf{test}$} & $A_\textsf{wb}$ & $A_\textsf{bb}$ & $A_\textsf{bl}$ \\ \hline
WD & 10.3 & 42.5 & +\textbf{74.4}\% & 54.9 & 55.4 & 55.2 \\ \hline
WD + DR & 9.1 & 42.1 & +\textbf{76.0}\% & 56.4 & 56.8 & 54.6 \\ \hline
WD + LS & 12.3 & 42.0 & +\textbf{76.4}\% & 57.2 & 57.0 & 56.2 \\ \hline
% WD + CP & 12.5 & 43.4 & +\textbf{30.7} & 56.4 & 56.4 & 56.3 \\ \hline %\Xcline{1-11}{.7pt}
\hline

\multicolumn{7}{|c|}{{Texas + FC} (DMP's $A_\textsf{test}$ = 48.6)} \\ \hline
{Regularizer} & {$E_\textsf{gen}$} & {$A_\textsf{test}$} & {$A^{+}_\textsf{test}$} & $A_\textsf{wb}$ & $A_\textsf{bb}$ & $A_\textsf{bl}$ \\ \hline
WD & 5.0 & 22.5 & +\textbf{116}\% & 58.3 & 57.7 & 52.5 \\ \hline
WD + DR & 6.1 & 14.2 & +\textbf{242}\% & 63.1 & 62.6 & 53.1 \\ \hline
WD + LS & 8.3 & 37.3 & +\textbf{30}\% & 61.7 & 61.0 & 54.2 \\ \hline
% WD + CP & -- & -- & -- & -- & -- & --\\ \hline %\Xcline{1-11}{.7pt}
\hline

\multicolumn{7}{|c|}{{CIFAR100 + DenseNet-12} (DMP's $A_\textsf{test}$ = 63.1)} \\ \hline
{Regularizer} & {$E_\textsf{gen}$} & {$A_\textsf{test}$} & {$A^{+}_\textsf{test}$} & $A_\textsf{wb}$ & $A_\textsf{bb}$ & $A_\textsf{bl}$ \\ \hline
WD & 4.0 & 26.3 & +\textbf{140}\% & 49.9 & 49.7 & 52.0 \\ \hline
WD + DR & 3.7 & 32.3 & +\textbf{95.4}\% & 51.2 & 51.0 & 51.9 \\ \hline
WD + LS & 2.7 & 13.0 & +\textbf{385}\% & 51.0 & 51.4 & 51.4 \\ \hline
% WD + CP & -- & -- & -- & -- & -- & -- \\ \hline %\Xcline{1-11}{.7pt}
\hline

\multicolumn{7}{|c|}{{CIFAR10 + Alexnet} (DMP's $A_\textsf{test}$ = 65.0)} \\ \hline
{Regularizer} & {$E_\textsf{gen}$} & {$A_\textsf{test}$} & {$A^{+}_\textsf{test}$} & $A_\textsf{wb}$ & $A_\textsf{bb}$ & $A_\textsf{bl}$ \\ \hline
WD & 4.1  & 45.9 & +\textbf{41.6}\% & 52.4 & 52.5 & 52.1 \\ \hline
WD + DR & 3.2 & 44.7 & +\textbf{45.4}\% & 51.9 & 51.7 & 51.6 \\ \hline
WD + LS & 4.8 & 53.2 & +\textbf{22.2}\% & 53.8 & 53.0 & 52.4 \\ \hline
% WD + CP & -- & -- & & -- & -- & --\\ \hline

\end{tabular}
\vspace*{-.6em}
\caption{Evaluating three state-of-the-art regularizers, with similar, low MIA risks (high membership privacy) as DMP. 
$A^{+}_\textsf{test}$ shows \emph{the \% increase} in $A_\textsf{test}$ due to DMP over the corresponding regularizers. 
}
\label{table:regularization_comparison}

\end{center}
\vspace*{-2em}
\end{table}

\subsubsection{Comparison with other regularizers.}
Next, we compare DMP with four state-of-the-art regularizers: weight decay (WD), dropout \cite{srivastava2014dropout} (DR), label smoothing \cite{szegedy2016rethinking} (LS), and confidence penalty \cite{pereyra2017regularizing} (CP). Due to the poor MIA resistance of CP, we defer its results to Appendix.

Table~\ref{table:regularization_comparison} shows the results, when MIA risks of regularized models is close that of DMP models (Table~\ref{table:performance_comparison}).
We note that, in all the cases, $A_\mathsf{test}$ of DMP are significantly higher (up to 385\% increase as $A^+_\mathsf{test}$ column specifies) than $A_\mathsf{test}$ of other regularizers.
This is because, these regularizers aim to improve the test accuracies of target models, but are not designed to reduce MIA risk. Thus, to reduce MIA risk, these regularization techniques add large, suboptimal noises during training, and hurt the utility of resulting models.

\begin{table}[h]
\fontsize{8.5}{9}\selectfont{}
\setlength{\extrarowheight}{0cm}
\begin{center}

\begin{tabular}{ |c|c|c|c|c| } 
\hline
\multirow{2}{*}{Defense} & Privacy & \multirow{2}{*}{$E_\mathsf{gen}$} & \multirow{2}{*}{$A_\mathsf{test}$} & \multirow{2}{*}{$A_\mathsf{wb}$} \\

& budget $(\epsilon)$  &  &   &  \\ \hline

No defense & -- & 32.5 & 67.5 & 77.9 \\ \hline
DMP & -- & \cellcolor[gray]{0.8}3.10 & \cellcolor[gray]{0.8}65.0 & \cellcolor[gray]{0.8}51.3 \\ \hline

\multirow{4}{*}{ DP-SGD } & 198.5 & \cellcolor[gray]{0.8}3.60 & \cellcolor[gray]{0.8}52.2 & \cellcolor[gray]{0.8}51.7 \\ 
& 50.2 & 1.30 & 36.9 & 50.2\\ 
& 12.5 & 0.30 & 31.7 & 50.0 \\  
& 6.8 & -1.60 & 29.4 & 49.9 \\ 

 \hline

\end{tabular}
\vspace{-.6em}
\caption{DP-SGD versus DMP for CIFAR10 and Alexnet. For low MIA risk of $\sim51.3$\%, DMP achieves 24.5\% higher $A_\mathsf{test}$ than of DP-SGD (12.8\% absolute increase in $A_\mathsf{test}$).}
\label{table:dp_sgd}
\end{center}
\vspace*{-1.2em}
\end{table}

\subsection{Comparison with differentially private defenses}\label{exp:dp}

\subsubsection{Comparison with DP-SGD.}
Following the methodology of~\cite{jayaraman2019evaluating}, we compare DMP and DP-SGD~\cite{abadi2016deep} using the empirically observed tradeoffs between membership privacy (MIA resistance) and $A_\mathsf{test}$ of models.
We use only CIFAR10 for these experiments, as the DP-SGD achieves prohibitively low accuracies on difficult tasks such as Texas and CIFAR100. We evaluate MIA risk using the whitebox NSH attack.
Table \ref{table:dp_sgd} shows the results of  Alexnet trained on CIFAR10 using DMP and DP-SGD with different privacy budgets $\epsilon$'s; -ve $E_\mathsf{gen}$ implies $A_\mathsf{train}$ is lower than $A_\mathsf{test}$.
DP-SGD incurs significant (35\%) loss in $A_\mathsf{test}$ at lower $\epsilon$ (12.5) to provide strong membership privacy.
At higher $\epsilon$, $A_\mathsf{test}$ of DP-SGD increases, but at the cost of very high generalization error, which facilitates stronger MIAs. 
Note that, further increase in privacy budget, $\epsilon$, does not improve tradeoff of DP-SGD.
More importantly, {for low MIA risk of $\sim$ 51.3\%, DMP models have 12.8\% higher $A_\mathsf{test}$ (i.e., 24.5\% improvement) than DP-SGD models}, which shows the superior tradeoffs due to DMP.

\subsubsection{Comparison with PATE.} PATE~\cite{papernot2017semi}, a semi-supervized learning technique, requires a compatible pair of generator and disciminator to achieve acceptable performances. Hence, we use CIFAR10 dataset and, instead of Alexnet, use the generator-discriminator pair from~\cite{salimans2016improved}, which has state-of-the-art performances.
PATE trains a set of teachers, computes hard labels of each teacher on some $X_\mathsf{ref}$, aggregates the labels for each $\mathbf{x}\in X_\mathsf{ref}$ using majority voting, adds DP noise to the aggregate, and finally trains its  target model on the noisy aggregate.

We train ensembles of 5, 10, and 25 teachers using $D_\mathsf{tr}$ of sise 25k.
We use the optimized confident-GNMax (GNMax) aggregation scheme of \cite{papernot2018scalable} to label $X_\mathsf{ref}$
We present a subset of results in Table \ref{table:pate} and defer comprehensive comparison to Appendix.
At low $\epsilon$'s ($<$10), GNMax hardly produces any labels, hence, the final target model has very low $A_\mathsf{test}$, but at higher $\epsilon$'s ($>$1000), PATE target model has acceptable $A_\mathsf{test}$.
However, PATE cannot achieve performances close to DMP, as it divides $D_\mathsf{tr}$ among its teachers. Such teachers have significantly lower accuracies and their ensemble cannot achieve the accuracy close to that of the unprotected model of DMP, which is trained on the entire $D_\mathsf{tr}$. Hence, the quality of knowledge transferred in DMP is always higher than that in PATE.

\begin{table}[h]
\fontsize{8.5}{9}\selectfont{}
\begin{center}

\begin{tabular}{ |c|c|c|c|c|c| } 
\hline
% \multirow{2}{*}
\centering
{\# of} & Queries & Privacy & \multicolumn{2}{c|}{Target model} & \multirow{2}{*}{$A_\mathsf{wb}$} \\

Teachers & answered & budget $(\epsilon)$ & $E_\mathsf{gen}$ & $A_\mathsf{test}$ & \\ \hline

\multirow{2}{*}{5} & 49 & 195.9 & 31.4 & 33.9 & 49.1 \\
& 1163 & 11684  &65.4 & 68.1 &49.0  \\ \hline

\multirow{2}{*}{10} & 23 & 42.9 &39.1  & 38.3 & 50.1\\
& 1527 & 6535 & 63.9& 65.2 & 49.8 \\ \hline

\multirow{2}{*}{25} & 108 & 183.5 & 53.8& 55.7 & 49.0 \\
& 4933 & 1794.1 & 57.8& 60.3 & 48.6\\ \hline

\end{tabular}
\vspace*{-0.6em}
\caption{
Comparing PATE with DMP.
DMP has $E_\mathsf{gen}$, $A_\mathsf{test}$, and $A_\mathsf{wb}$ of 1.19\%, 76.79\%, and 50.8\%, respectively.
PATE has low accuracy even at high privacy budgets, as it divides data among teachers and produces low accuracy ensembles.
}
\label{table:pate}
\end{center}
\vspace*{-2.5em}
\end{table}

\subsection{Discussions}
Below, we provide further key insights in to DMP defense and defer their detailed discussion to Appendix.

\subsubsection{Hyperparameter selection in DMP. } \emph{Increasing} the temperature of softmax layer of the unprotected model, $\theta_\mathsf{up}$, used to transfer the knowledge of $\theta_\mathsf{up}$, can further reduce the membership leakage of $D_\mathsf{tr}$. This is because, at higher softmax temperatures, predictions of $\theta_\mathsf{up}$ have uniform distribution over all classes and contain no useful information for MIAs.
Similarly, reducing the size of $X_\mathsf{ref}$ reduces MIA risk due to DMP, but comes at the cost of reduction in $A_\mathsf{test}$.

\subsubsection{Privacy risk to reference data ($X_\mathsf{ref}$). }
We evaluate the privacy risk to $X_\mathsf{ref}$, as it can be of sensitive nature, e.g., in case of Texas medical records dataset.
Our results in appendix show that given the final DMP model, $\theta_\mathsf{p}$, and a target sample, MIA adversary (who mounts BL, NN, or NSH attacks) cannot decide if the sample belonged to $X_\mathsf{ref}$ with sufficient confidence.
This is expected, because DMP trains its $\theta_\mathsf{p}$ on noisy, soft-labels of $X_\mathsf{ref}$, which do not contain any sensitive information about $X_\mathsf{ref}$ and its ground-truth labels, which is necessary for MIAs to succeed~\cite{yeom2018privacy}.
We provide detailed results in Appendix.

\subsubsection{DMP with synthetic reference data ($X_\mathsf{ref}$). }
Following previous works~\cite{papernot2018scalable,papernot2017semi}, including the state-of-the-art MIA defense AdvReg~\cite{nasr2018machine}, we assume availability of $X_\mathsf{ref}$.
However, in privacy sensitive domains such as patient medical records, $X_\mathsf{ref}$ may not be available. 
Hence, we show that the assumption can be relaxed by using $X_\mathsf{ref}$ {synthesized} from private $D_\mathsf{tr}$ to train DMP models. 
For CIFAR10, we use  DC-GAN to generate synthetic $X_\mathsf{ref}$ of sizes 12.5k, 25k, and 37.5k from $D_\mathsf{tr}$ of size 25k. We then train three DMP models and evaluate their MIA risk using whitebox NSH attack. We note that for 12.5k, 25k, and 37.5k synthetic $X_\mathsf{ref}$ samples, ($E_\mathsf{gen}$, $A_\mathsf{test}$, $A_\mathsf{wb}$) of DMP are (2.1, 53.0, 50.3), (3.5, 56.8, 51.3), and (5.0, 57.5, 52.1), respectively. Note that, \emph{DMP outperforms existing defenses even with synthetic $X_\mathsf{ref}$} (Tables~\ref{table:performance_comparison},~\ref{table:regularization_comparison}).

\subsubsection{Adaptive attack on DMP. }
In DMP, the reference data, $X_\mathsf{ref}$, is selected such that the predictions of DMP's unprotected model $\theta_\mathsf{up}$ on $X_\mathsf{ref}$ have low entropies. Due to memorization, predictions of $\theta_\mathsf{up}$ on $D_\mathsf{tr}$ also have low entropies. Hence, an adaptive adversary may exploit this peculiar $X_\mathsf{ref}$ selection in DMP. Based on this intuition, we investigate the possibility of an adaptive MIA, which labels a target sample as a member if the sample is close to some $X_\mathsf{ref}$ datum in feature space. However, such attack has accuracy close to random guess. This is because, we observe that the proximity of two samples in feature space has no correlation with the entropy of predictions of given $\theta_\mathsf{up}$ on those samples, which is the selection criterion of DMP. We leave further investigation of adaptive attacks on DMP to future work.

% !TEX root = main.tex
\section{Conclusions}

We proposed Distillation for Membership Privacy (DMP), a knowledge distillation based defense against membership inference attacks that significantly improves the membership privacy-model utility tradeoffs compared to state-of-the-art defenses. We provided a novel criterion to generate/select reference data in DMP and achieve the desired tradeoffs. Our extensive evaluation demonstrated the state-of-the-art privacy-utility tradeoffs of DMP.

\vspace{1em} 
\paragraphb{Acknowledgments.}
This work was supported in part by NSF grant CPS-1739462.

\bibliography{ml_privacy}

\appendix

% !TEX root = main.tex

\section{Fine tuning the DMP defense\\(Missing details)}\label{analysis:ref_choice}

We propose a fine tuning technique to select/generate appropriate reference data, $X_\mathsf{ref}$, and achieve the desired privacy-utility tradeoffs using our distillation for membership privacy (DMP) defense.
The technique depends on the result given in Proposition~\ref{prop:entropy}; we provide a detailed proof of the results below.

\paragraphb{Detailed proof of Proposition~\ref{prop:entropy}. }

\paragraphe{Deriving the objective for desired $X_\textsf{ref}$. }
Consider two training datasets $D_\textsf{tr}$ and $D'_\textsf{tr}$ such that $D'_\textsf{tr}\leftarrow D_\textsf{tr}-z$, and $X_\textsf{ref}$.
Then, the log of the ratio of the posterior probabilities of learning the exact same parameters $\theta_\textsf{p}$ using DMP is given by \eqref{eq:obj_ratio}.
Observe that, $\mathcal{R}$ is an extension of \eqref{eq:prob_ratio} to the setting of DMP, where $\theta_\mathsf{p}$ is trained via the knowledge transferred using $(X_\textsf{ref},\theta^{X_\textsf{ref}}_\textsf{up})$, instead of directly training on $D_\textsf{tr}$.
\cite{sablayrolles2019white} argue to reduce this ratio to improve membership privacy.
Hence, we want to obtain $X_\mathsf{ref}$ which reduces the ratio $\mathcal{R}$ when $D_\textsf{tr}$, $D'_\textsf{tr}$, and $\theta_\mathsf{p}$ are kept constant.
We note that, although similar in appearance to differential privacy, $\mathcal{R}$ is defined only for the given private dataset, $D_\textsf{tr}$.
\begin{equation}\label{eq:obj_ratio}
\mathcal{R}=\Big|\text{log}\ \frac{\text{Pr}(\theta_\textsf{p}|D_\textsf{tr},X_\textsf{ref})}{\text{Pr}(\theta_\textsf{p}|D'_\textsf{tr},X_\textsf{ref})}\Big|
\end{equation}

Next, we modify $\mathcal{R}$ as:
\begin{align}
\label{eq:obj_ratio1}
& \mathcal{R}= \Big|-\frac{1}{T}\sum_{\mathbf{x}\in X_\textsf{ref}} \mathcal{L}_{\scaleto{\textsf{KL}}{4pt}}((\mathbf{x},\theta^{\mathbf{x}}_\textsf{up});\theta_\textsf{p}) - \mathcal{L}_{\scaleto{\textsf{KL}}{4pt}}((\mathbf{x},\theta'^{\mathbf{x}}_\textsf{up});\theta_\textsf{p})\Big| \\
\label{eq:obj_ratio2}
&\leq \frac{1}{T} \sum_{\mathbf{x}\in X_\textsf{ref}}\Big| \mathcal{L}_{\scaleto{\textsf{KL}}{4pt}}(\theta^{\mathbf{x}}_\textsf{up}\Vert\theta^{\mathbf{x}}_\textsf{p}) - \mathcal{L}_{\scaleto{\textsf{KL}}{4pt}}(\theta'^{\mathbf{x}}_\textsf{up}\Vert\theta^{\mathbf{x}}_\textsf{p})\Big|
\end{align}

\noindent where $\theta_\textsf{up}$ and $\theta'_\textsf{up}$ are trained on $D_\textsf{tr}$ and $D'_\textsf{tr}$, respectively.
Note that, \eqref{eq:obj_ratio1} holds due to the assumption in \eqref{eq:post_assumption} and the KL-divergence loss used to train $\theta_\mathsf{p}$ in DMP.
\eqref{eq:obj_ratio2} follows from \eqref{eq:obj_ratio1} because $|a+b|\leq|a|+|b|$.
Therefore, minimizing \eqref{eq:obj_ratio2} implies minimizing \eqref{eq:obj_ratio}.
Thus, to improve membership privacy due to $\theta_\mathsf{p}$, $X_\mathsf{ref}$ is obtained by solving~\eqref{eq:ref_obj}.
\begin{align}
\label{eq:ref_obj}
X^*_\textsf{ref}=\underset{X_\textsf{ref}\in X}{\text{argmin}}\Big(\frac{1}{T}\sum_{\mathbf{x}\in X_\textsf{ref}}  \big|\mathcal{L}_{\scaleto{\textsf{KL}}{4pt}}&(\theta^{\mathbf{x}}_\textsf{up}\Vert\theta^{\mathbf{x}}_\textsf{p}) -\mathcal{L}_{\scaleto{\textsf{KL}}{4pt}}(\theta'^{\mathbf{x}}_\textsf{up}\Vert\theta^{\mathbf{x}}_\textsf{p})\big|\Big)
\end{align}

The objective of \eqref{eq:ref_obj} is minimized when $\theta^{\mathbf{x}}_\mathsf{up} = \theta'^{\mathbf{x}}_\mathsf{up}\ \ \forall\mathbf{x}\in X_\mathsf{ref}$ and is very intuitive: It implies that, $z$ (i.e., $D_\mathsf{tr}-D'_\mathsf{tr}$) enjoys stronger membership privacy when the reference data, $X_\mathsf{ref}$, are such that \emph{the distributions of outputs of $\theta_\mathsf{up}$ and $\theta'_\mathsf{up}$ on $X_\mathsf{ref}$   are not affected by the presence of $z$ in $D_\mathsf{tr}$}.
\\

\paragraphe{Simplifying the objective. }
Next, we simplify \eqref{eq:ref_obj} by replacing $\mathcal{L}_{\scaleto{\textsf{KL}}{4pt}}$ with closely related cross-entropy loss $\mathcal{L}_{\scaleto{\textsf{CE}}{4pt}}$. 
The simplified objective is given by \eqref{eq:ref_obj2}.
\begin{align}\label{eq:ref_obj2}
X^*_\textsf{ref}=\underset{X_\textsf{ref}\in X}{\text{argmin}}\sum_{\substack{z'=(\mathbf{x},y)\\ \in (X_\textsf{ref},Y_\textsf{ref})}} \frac{1}{T} \big|\mathcal{L}_{\scaleto{\textsf{CE}}{4pt}}(z';\theta'_\textsf{up}) - \mathcal{L}_{\scaleto{\textsf{CE}}{4pt}}(z';\theta_\textsf{up})\big| 
\end{align}
where $\mathcal{L}_{\scaleto{\textsf{CE}}{4pt}}$ is cross-entropy loss and $z'$ is not the same as $z\leftarrow D_\mathsf{tr}-D'_\mathsf{tr}$.
For clarity of presentation, here onward, we denote $\mathcal{L}_{\scaleto{\textsf{CE}}{4pt}}$ by $\mathcal{L}$.

\begin{figure}[t]
\centering
\includegraphics[height=7.5cm,width=7.5cm,trim=3cm 3cm 3cm 3cm]{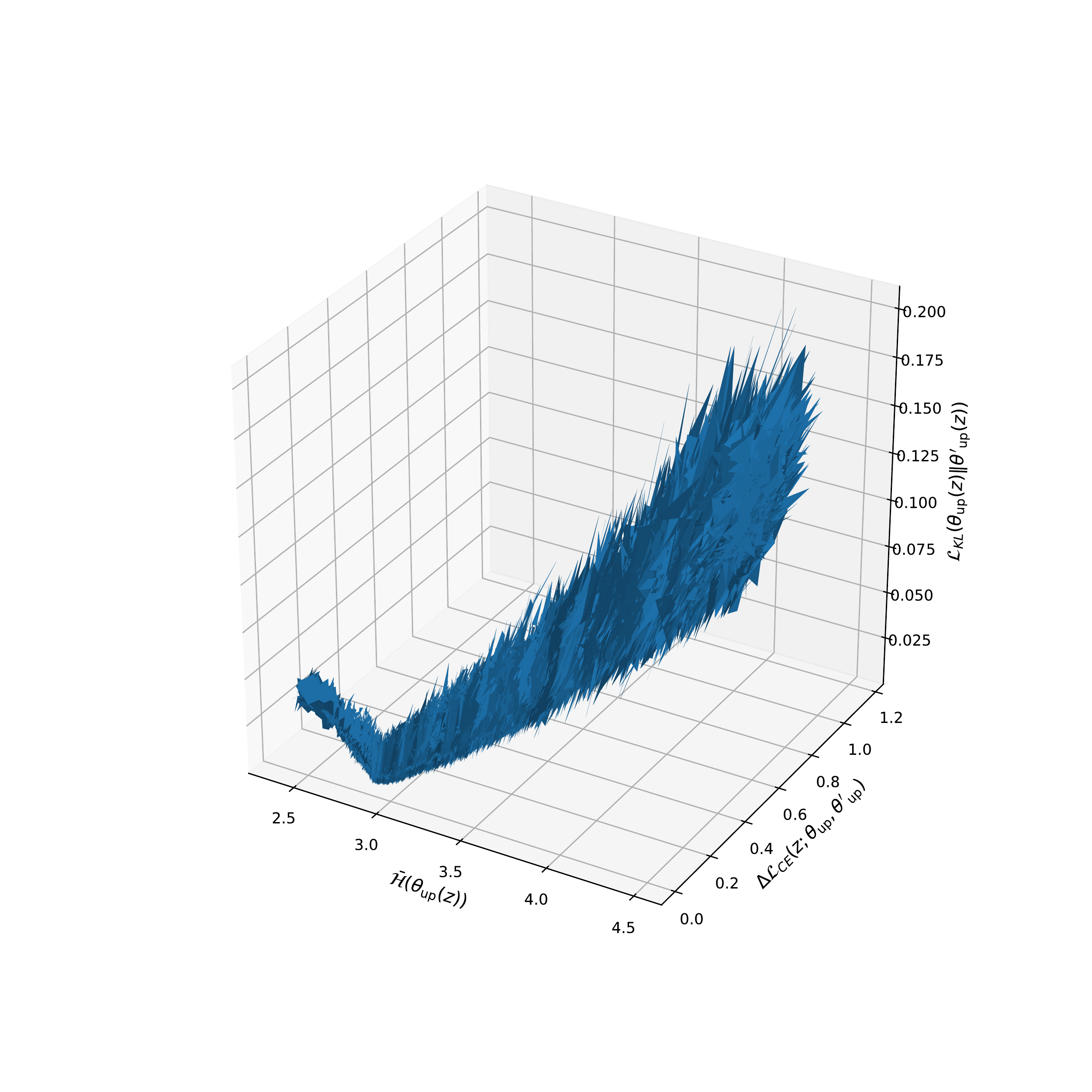}
\caption{Empirical validation of simplification of \eqref{eq:ref_obj} to \eqref{eq:ref_obj2}: Increase in $\Delta\mathcal{L}_{\scaleto{\textsf{CE}}{4pt}}$ increases $\Delta\mathcal{L}_{\scaleto{\textsf{KL}}{4pt}}$, and that of \eqref{eq:ref_obj} to \eqref{eq:final_ref_obj}: Increase in $\mathcal{H}(\theta_\textsf{up}(z))$ increases $\Delta\mathcal{L}_{\scaleto{\textsf{KL}}{4pt}}$.}
\label{fig:kl_to_ce}
\vspace*{-1em}
\end{figure}

Next, we assume that ground truth labels $Y_\textsf{ref}$ of $X_\textsf{ref}$ are available. Note that $X_\textsf{ref}$ is unlabeled dataset, but  \emph{only to empirically demonstrate the validity of the simplification of \eqref{eq:ref_obj} to \eqref{eq:ref_obj2}, we assume that ground truth labels of $X_\textsf{ref}$ are available}.
We validate the simplification in Figure \ref{fig:kl_to_ce}: for any given reference sample, the lower the difference between cross-entropy losses, $\Delta\mathcal{L}$, the lower the corresponding difference between KL-divergence losses; and vice-versa. 
Note that, to select/generate a reference sample, we do not need the exact difference between cross-entropy or KL-divergence losses for the sample, but only the difference for the sample relative to the other samples.
Hence, although the difference between cross-entropy losses is not exactly the same as difference between KL-divergence losses, their strong positive correlation is sufficient to make  the reduction  \eqref{eq:ref_obj} $\rightarrow$ \eqref{eq:ref_obj2} useful in our task.
\\

\vspace{.5em}
\paragraphe{Deriving the final objective to select/generate $X_\mathsf{ref}$. }
Next, to avoid repetitive training, we simplify the term for each sample in \eqref{eq:ref_obj2} using the results of~\cite{koh2017understanding}.
More specifically, they propose a linear approximation to the difference in cross-entropy losses of a pair of models trained with and without a specific sample in their training data.
We note that this is the exact setting of our problem.
If $\theta$ and $\theta_{-z}$ are two models trained with and without a member $z$, then the difference in cross-entropy losses of the two models on some test sample $z_\mathsf{test}=(\mathbf{x}_\mathsf{test},y_\mathsf{test})$ is approximated as:

\begin{equation}\label{koh_result}
|\mathcal{L}(z_\text{test},\theta_{-z}) - \mathcal{L}(z_\mathsf{test},\theta)| \simeq |\nabla_{\theta}\mathcal{L}(z_\mathsf{test},\theta)H^{-1}_{\theta}\nabla_{\theta}\mathcal{L}(z,\theta)|
\end{equation}

\noindent where $H_\theta$ is the Hessian matrix that is  defined as $H_\theta=\frac{1}{n}\sum_{z\in D_\textsf{tr}}\nabla^2_\theta(z,\theta)$.
Substituting \eqref{koh_result} in \eqref{eq:ref_obj2} simplifies the objective in \eqref{eq:ref_obj} to:

\begin{align}
\label{eq:sim_ref_obj}
X^*_\textsf{ref}=\underset{X_\textsf{ref}\in X}{\text{argmin}}\sum_{\substack{z'=(\mathbf{x},y)\\ \in (X_\textsf{ref},Y_\textsf{ref})}} \frac{1}{T}|\nabla_{\theta}\mathcal{L}(z',\theta_\textsf{up})H^{-1}_{\theta}\nabla_{\theta}\mathcal{L}(z',\theta_\textsf{up})|
\end{align}

Note that, for a given member $z$, $H^{-1}_{\theta}\nabla_{\theta}\mathcal{L}(z',\theta)$ in \eqref{eq:sim_ref_obj} remains constant and the minimization reduces to minimizing the gradient $\nabla_{\theta}\mathcal{L}(z_\textsf{p},\theta_\textsf{up})$.
The lower the loss $\mathcal{L}(z',\theta_\textsf{up})$, the smaller the gradient $\nabla_{\theta}\mathcal{L}(z',\theta_\textsf{up})$.
Therefore objective \eqref{eq:sim_ref_obj} further simplifies as:

\begin{align}
\label{eq:sim_ref_obj1}
X^*_\textsf{ref}=\underset{X_\textsf{ref}\in X}{\text{argmin}}\ \frac{1}{T}\sum_{\substack{z'=(\mathbf{x}',y)\\ \in (X_\textsf{ref},Y_\textsf{ref})}} \mathcal{L}{\scaleto{\textsf{CE}}{4pt}}(z',\theta_\textsf{up})
\end{align}

Note that, in practice, it is not possible to solve the objective in \eqref{eq:sim_ref_obj1} as it is. Because, we cannot compute the loss without the ground truth labels of $X_\textsf{ref}$; recall that $X_\textsf{ref}$ is \emph{unlabeled}.
However, as the loss involved here is the cross-entropy loss, minimizing the loss is equivalent to minimizing the entropy of prediction $\theta_\textsf{up}(\mathbf{x}')$.
This gives us the final objective as:
\begin{align}
\label{eq:final_ref_obj}
X^*_\textsf{ref}=\underset{X_\textsf{ref}\in X}{\text{argmin}}\ \frac{1}{T}\sum_{\mathbf{x}'\in X_\textsf{ref}} \mathcal{H}(\theta_\textsf{up}(\mathbf{x}'))
\end{align}
where, $\mathcal{H}(\mathbf{v})\triangleq \sum_{i}-\mathbf{v}_i\text{log}(\mathbf{v}_i)$ is the entropy of $\mathbf{v}$.
This provides the result of Proposition~\ref{prop:entropy}.

Proposition~\ref{prop:entropy} states that, \emph{using the reference data with low entropy predictions of $\theta_\textsf{up}$ strengthens the membership resistance of $\theta_\textsf{p}$, and vice versa.}
In Figure~\ref{fig:kl_to_ce}, we empirically validate the reductions \eqref{eq:ref_obj} $\rightarrow$ \eqref{eq:sim_ref_obj1}$\rightarrow$ \eqref{eq:final_ref_obj}.
Specifically, we show that, for a given $\theta_\mathsf{up}$, the lower the cross-entropy loss of reference data sample, the lower the entropy of prediction of $\theta_\mathsf{up}$ on the sample, i.e.,  \eqref{eq:sim_ref_obj1}$\rightarrow$ \eqref{eq:final_ref_obj}.
Then, we show that, the difference between cross-entropy losses of two models $\theta_\textsf{up}$ and $\theta'_\textsf{up}$, trained on neighboring datasets, on a sample increases with the increase in cross-entropy loss of their prediction on the sample, i.e., \eqref{eq:ref_obj2} $\rightarrow$ \eqref{eq:sim_ref_obj1}.
This, in combination with the reduction \eqref{eq:ref_obj} $\rightarrow$ \eqref{eq:ref_obj2} demonstrated in Figure \ref{fig:kl_to_ce}, completes the validation of \eqref{eq:ref_obj} $\rightarrow$ \eqref{eq:sim_ref_obj1}.
Figure~\ref{fig:hypothesis_eval_1} validates our hypothesis.

% !TEX root = main.tex
\section{Missing Details of Experimental Setup}\label{exp_setup}

\subsection{Computing environment}\label{setup:computing}

We will make our code and all the relevant datasets (all the datasets used are already available online) publicly available upon acceptance of the submission.
We perform all of our experiments using PyTorch 1.2 framework on TitanX GPU of 12GB memory.
All the experimental results in the paper are average of three runs of the corresponding experimental setting.

\subsection{Target model architectures}\label{setup:architecutres}

Unlike conventional distillation \cite{hinton2014distilling}, DMP uses  same architectures for  unprotected and protected models. 
Needless to say, using a lower-capacity architecture for the protected model will further improve privacy protection at the cost of reducing  utility (prediction accuracy).   
The details of the architectures for all the datasets is given in Table \ref{table:exp_setup}.
For Purchase-100 and Texas-100, the fully connected network has hidden layers of sizes \{1024, 512, 256, 128\}.
For CIFAR-100, we choose two DenseNet models to assess the efficacy of DMP  for two models with equivalent performance, but significantly different capacities.
In Table \ref{table:exp_setup}, DenseNet12 corresponds to DenseNet-BC (L=100, k=12) and DenseNet19 corresponds to DenseNet-BC (L=190, k=40).
For the comparison with PATE using CIFAR-10, we use the generator and discriminator architectures used in \cite{salimans2016improved}.
% !TEX root = main.tex
\begin{table*}
% \hspace{-2em}
\fontsize{8.5}{9}\selectfont{}

\begin{center}

\begin{tabular}{ |c|c|c|c|c|c|c|c|c|c|c|c| } 
\hline

\multicolumn{4}{|c|}{5 Teachers} & \multicolumn{4}{c|}{10 Teachers} & \multicolumn{4}{c|}{25 Teachers} \\ \hline

Queries & Privacy & GNMax & Student & Queries & Privacy & GNMax & Student & Queries & Privacy & GNMax & Student \\
answered & bound $\epsilon$ & accuracy & accuracy & answered & bound $\epsilon$ & accuracy & accuracy & answered & bound $\epsilon$ & accuracy & accuracy \\ \hline

0 & 4.6 & -- & -- & 0 & 9 & -- & -- & 0 & 8.43 & -- & -- \\ \hline
49 & 195.9 & 79.6 & 33.93 & 23 & 42.87 & 56.5 & 38.28 & 108 & 183.5 & 95.4 & 55.7 \\ \hline
127 & 281.6 & 69.3 & 49.89 & 358 & 409.5 & 67.0 & 57.59 & 357 & 231.3  & 83.9& 56.14 \\ \hline
679 & 1283.7 & 70.3 & 58.04& 1128 & 1092.5 & 66.13 & 60.94 & 1130 & 508.9 & 83.8 & 58.26 \\ \hline
1163 & 11684 & 91.1 & 68.08 & 1527 & 6535 & 93.1 & 65.18 & 4933 & 1794.1 & 74.0 & 60.27 \\ \hline

\end{tabular}
\end{center}
\caption{Evaluation of PATE using the discriminator architecture in \cite{salimans2016improved} trained on CIFAR10. 
The corresponding DMP-trained model has 77.98\% and 76.79\% accuracies on the training and test data, and 50.8\% membership inference accuracy.
Comparison of results clearly show the superior membership privacy-model utility tradeoffs of DMP over PATE.
}
\label{table:pate_comparison}

\end{table*}

\section{Detailed comparison with PATE}\label{appendix:pate_details}

In this section, we detail the experimental comparison between PATE \cite{papernot2018scalable,papernot2017semi}  and our DMP defense for CIFAR10 classification task.
The motivation of this comparison is to show that the DMP-trained models achieve significantly better tradeoffs between membership privacy (i.e., resistance to membership inference attacks) and classification accuracy than the PATE-trained models.

PATE relies on semi-supervised learning that uses a large unlabeled dataset.
PATE computes the labels of a subset of the unlabeled data using an ensemble of teachers.
Each of the teachers is trained on a disjoint set of the private training dataset; all sets have the same size.
Semi-supervised learning involves an unstable game between a generator $G$ and a discriminator $D$. Hence, the architectures of $G$ and $D$ should be compatible for effective learning.
Therefore, instead of AlexNet, which we use in the rest of our CIFAR10 experiments, we use the the pair of discriminator and generator architectures proposed in \cite{salimans2016improved} due to its state-of-the-art classification performance.
Finally, PATE uses its discriminator as the classification model.
For both PATE and DMP, we use the same 25,000 data of CIFAR10 as the private training and the rest of 25,000 data the unlabeled reference data.
The accuracy of the discriminator trained on the entire private training data is 97.65\% and 79.6\% on training and test data, respectively.

We use the 25,000 \emph{training} data to train three ensembles of sizes 5, 10 and 25 teachers.
Each of the teachers in all the ensembles have disjoint and equal-sized training data.
The accuracy, \emph{without adding any noise to labels}, of the corresponding ensembles on the 25000 \emph{reference} samples is 64.92\%, 60.1\% and 54.52\%, respectively.
We use the confident-GNMax (GNMax) aggregation scheme to add DP noise to the aggregate of the votes (i.e., hard labels) of the teachers on the unlabeled reference data.
GNMax labels samples based on remaining privacy budget, hence, it may not label all the reference data samples.
GNMax aggregation scheme is similar to the sparse vector technique \cite{dwork2014algorithmic} and outputs a label only if the noisy version of the votes count of the label crosses a noisy version of a fixed threshold.
Table \ref{table:pate_comparison} details the accuracy of the GNMax aggregation  for different number of teachers and privacy levels $(\epsilon,\delta)$. 
We use $\delta$ of $10^{-4}$ as the order of the size of the reference data is $10^{4}$ \cite{papernot2018scalable}.

Note that, the DMP-trained discriminator has training, test, and attack accuracies of 77.98\%, 76.79\%, and 50.8\%, respectively.
Table \ref{table:pate_comparison} shows results for PATE with teacher ensembles of different sizes: 
At low $\epsilon$ values, GNMax cannot provide many labels, and therefore, PATE suffers significant accuracy degradations.
While at high $\epsilon$ values ($>$1000), GNMax performs better, but does not outperform DMP.
The reason for this is as follows:
At very high $\epsilon$'s, PATE is just a knowledge transfer based semi-supervised learning, while DMP is knowledge transfer based supervised learning.
DMP does not divide its training data among teacher, and therefore, the predictions of the unprotected model used in DMP to train the protected model are more useful in terms of both the quality and quantity.
Therefore, DMP-trained models have significantly higher accuracy than PATE-trained model, for similar membership inference risk, i.e., DMP achieves significantly better membership privacy-model utility tradeoffs.

% !TEX root = main.tex
\section{Missing Discussion Details}

In the last section of main paper, we provide various insights in to our DMP defense based on our extensive evaluation. We provide the missing details of the discussions below.

\subsection{Hyperparameter selection in DMP}\label{exp:param_variation}

\subsubsection{The temperature of the  softmax layer.}\label{exp:param_variation:temp}

The softmax temperature, $T$,  of the unprotected model, $\theta_\textsf{up}$, plays an important role in the amount of knowledge transferred from the unprotected to protected model in DMP.
Our results in Table~\ref{table:acc_vs_temp} confirm our analytical understanding of the use of the softmax temperature: increasing the temperature for AlexNet trained on CIFAR100 dataset reduces the classification accuracy of the final protected model, $\theta_\textsf{p}$, but also strengthens the its membership inference resistance.
Therefore, the softmax temperature $T$ should be chosen depending on the desired privacy-utility tradeoff.
Table \ref{table:exp_setup} shows the temperatures used in our experiments.

\begin{table}[h]
\fontsize{8.5}{10}\selectfont{}
\setlength{\extrarowheight}{0cm}

\begin{center}

\begin{tabular}{ |c|c|c|c|c| } 
\cline{1-5}

\multirow{2}{*}{Defense} & Softmax & Training & Test & Attack \\

& $T$ & Accuracy & Accuracy  & Accuracy  \\ \hline

No defense & n/a & 100 & 36.8 & 91.3 \\ \hline

\multirow{4}{*}{ DMP }& 2 & 46.6 & 37.3 & 57.4  \\ % \cline{2-5}
& 4  & 42.2 & 35.7 & 55.6   \\ % \cline{2-5}
& 6 & 36.4 & 32.8 & 52.5  \\ % \cline{2-5}
& 8 & 12.1 & 12.3 &  51.7 \\  \hline

\end{tabular}
\end{center}
\vspace{-1em}
\caption{Effect of the softmax temperature on DMP: For a fixed $X_\textsf{ref}$, increasing the temperature of softmax layer of $\theta_\textsf{up}$ reduces  $\mathcal{R}$ in \eqref{eq:obj_ratio}, which strengthens the membership privacy.}
\label{table:acc_vs_temp}
\end{table}

\begin{table}[h]
\fontsize{8.5}{9}\selectfont{}
\begin{center}

\begin{tabular}{ |c|c|c|c|c| }
\hline
{Combination} &\multirow{2}{*}{Dataset}& \multirow{2}{*}{Architecture} & \multirow{2}{*}{$|\theta|$} & \multirow{2}{*}{$T$} \\
acronym& & & & \\ \hline
P-FC & Purchase & Fully Connected & 1.32M & 1.0 \\ \hline
T-FC & Texas & Fully Connected & 1.32M & 1.0 \\ \hline
C100-A & \multirow{3}{*}{CIFAR100} & AlexNet & 2.47M & 4.0 \\ 
C100-D12 &  & DenseNet12 & 0.77M & 4.0 \\ 
C100-D19 &  & DenseNet19 & 25.6M & 1.0 \\ \hline
C10-A & CIFAR10 & AlexNet & 2.47M & 1.0 \\ \hline 
\end{tabular}
\end{center}
\vspace{-1em}
\caption{Temperature of the softmax layers for the different combinations of dataset and network architecture used to produce the results in Table 3 of the main paper.}
\label{table:exp_setup}
\end{table}

\subsubsection{The size of reference data.}\label{exp:param_variation:publen}

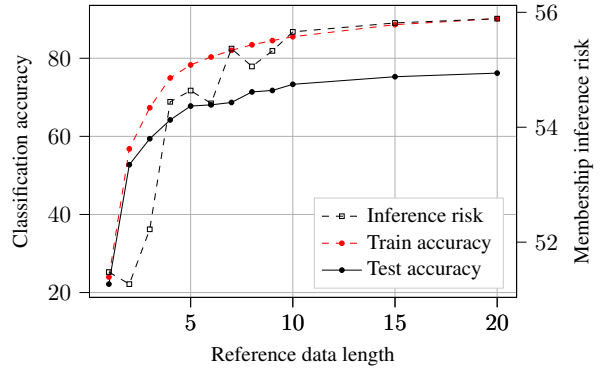
\begin{figure}
\centering
\resizebox{8cm}{5cm}{% This file was created by matplotlib2tikz v0.7.4.
\begin{tikzpicture}

\begin{axis}[
height=6cm,
width=8cm,
legend cell align={left},
legend style={font=\small,at={(0.97,0.03)}, anchor=south east, draw=white!80.0!black},
tick align=outside,
tick pos=left,
x grid style={white!69.01960784313725!black},
xlabel={Reference data length},
xlabel style={font=\small},
xmajorgrids,
xmin=0.0499999999999999, xmax=20.95,
xtick style={color=black},
y grid style={white!69.01960784313725!black},
ylabel={Classification accuracy},
ylabel style={font=\small},
ymajorgrids,
ymin=18.7665209690696, ymax=93.5396716713955,
ytick style={color=black}
]
\addlegendimage{/pgfplots/refstyle=inf_risk}\addlegendentry{Inference risk}
\addplot [line width=0.44000000000000006pt, red, dashed, mark=*, mark size=1, mark options={solid}]
table {%
1 23.9850427350427
2 56.7942040598291
3 67.331063034188
4 74.9565972222222
5 78.2919337606838
6 80.325186965812
7 82.0345886752137
8 83.450186965812
9 84.5219017094017
10 85.5101495726496
15 88.6017628205128
20 90.1408920940171
};
\addlegendentry{Train accuracy}
\addplot [line width=0.44000000000000006pt, black, mark=*, mark size=1, mark options={solid}]
table {%
1 22.1653005464481
2 52.7322404371585
3 59.3920765027322
4 64.207650273224
5 67.7595628415301
6 68.0669398907104
7 68.681693989071
8 71.379781420765
9 71.7554644808743
10 73.3265027322404
15 75.2732240437158
20 76.1953551912568
};
\addlegendentry{Test accuracy}
\end{axis}

\begin{axis}[
height=6cm,
width=8cm,
% legend cell align={left},
% legend style={font=\footnotesize, at={(0.97,0.19)}, anchor=south east, draw=white!80.0!black},
axis y line=right,
tick align=outside,
x grid style={white!69.01960784313725!black},
xmin=0.0499999999999999, xmax=20.95,
xtick pos=left,
xtick style={color=black},
y grid style={white!69.01960784313725!black},
ylabel={Membership inference risk},
ylabel style={font=\small},
ymin=51.0388122294372, ymax=56.1206574675325,
ytick pos=right,
ytick style={color=black},
]
\addplot [line width=0.44000000000000006pt, black, dashed, mark=square, mark size=1, mark options={solid}]
table {%
1 51.4794913419913
2 51.2698051948052
3 52.223538961039
4 54.4421536796537
5 54.6383116883117
6 54.4150974025974
7 55.3688311688312
8 55.057683982684
9 55.3282467532468
10 55.6596861471861
15 55.8152597402597
20 55.8896645021645
};\label{inf_risk}
% \addlegendentry{Inference risk}
\end{axis}

\end{tikzpicture}}
\vspace*{-.8em}
\caption{Increasing reference data size, $|X_\textsf{ref}|$, increases accuracy of $\theta_\textsf{p}$, but also increases $\mathcal{R}$ in \eqref{eq:obj_ratio}, which increases the membership inference risk due to $\theta_\textsf{p}$.}
\label{fig:cls_acc_vs_publen}
\vspace*{-.5em}
\end{figure}

In DMP, the more the reference data, the looser the bound on $\mathcal{R}$ in \eqref{eq:obj_ratio}, and therefore, weaker the membership resistance of the corresponding $\theta_\textsf{p}$.
To validate this, we quantify the classification accuracy and the membership inference risk of $\theta_\textsf{p}$ with increasing the amount of $X_\textsf{ref}$.
We use Purchase-100 data and vary $|X_\textsf{ref}|$ as shown in Figure \ref{fig:cls_acc_vs_publen}; we fix the softmax $T$ of $\theta_\textsf{up}$ at 1.0.
$\theta_\textsf{up}$ used here has train accuracy, test accuracy, and membership inference risk of 99.9\%, 77.0\% and 77.1\%, respectively.
Initially, the test accuracy of $\theta_\textsf{p}$ increases with $|X_\textsf{ref}|$  due to the useful knowledge transferred.
But, beyond the test accuracy of $\theta_\textsf{up}$, its predictions essentially insert noise in the training data of $\theta_\textsf{p}$, therefore the gain from increasing the size of reference data slows down.
Although this noise  marginalizes the increase in the test performance of $\theta_\textsf{p}$, it also prevents $\theta_\textsf{p}$ from learning more about $D_\textsf{tr}$ and prevents further inference risk. 
This is shown by the train accuracy and membership inference risk curves in Figure \ref{fig:cls_acc_vs_publen}.
Therefore, size of reference data should be selected based on the desired tradeoffs of the final model.

\subsection{Privacy risk to reference data ($X_\mathsf{ref}$)}\label{exp:ref_risk}

\begin{table}
\fontsize{8.5}{9}\selectfont{}
\begin{center}

\begin{tabular}{ |c|c|c|c|c| } 
\hline
{Dataset} & Test & Reference data & \multirow{2}{*}{$A_\textsf{wb}$} & \multirow{2}{*}{$A_\textsf{bb}$}  \\
\& model & acc. ($A_\textsf{test}$) & acc. ($A_\textsf{ref}$) & & \\ \hline

P-FC & 74.1 & 80.8 & 53.1 & 52.6 \\ \cline{1-5}

T-FC & 48.6 & 52.0 & 52.2 & 52.0 \\ \cline{1-5}

C100-A & 35.7 & 35.9 & 50.9 & 50.5 \\ \cline{2-5}

C100-D12  & 63.1 & 65.1 & 53.0 & 52.2 \\ \cline{2-5}

C10-A & 65.0 & 66.7 & 53.9 & 52.7 \\ \cline{1-5}

\end{tabular}
\end{center}
\vspace*{-1.0em}
\caption{DMP does not pose membership inference risk to the possibly sensitive reference data. $A_\mathsf{ref}$ and $A_\mathsf{test}$ are accuracies of protected model, $\theta_\mathsf{p}$, on $X_\mathsf{ref}$ and $D_\mathsf{test}$, respectively.}
\label{table:ref_risk}
\end{table}

The reference data used in DMP can be of sensitive nature.
For instance, for Texas-100, the reference data used are unlabeled, sensitive patients' records, and  therefore, at the risk of privacy breach.
However, we quantitatively show that \textbf{DMP does not pose membership inference risk to its reference data}.
The results are given in Table \ref{table:ref_risk}.
We note that, for any combination of model and dataset, the membership inference risk to the reference data due to DMP is close to 50\%, which is a random guess.
The intuition here is as follows.
$\theta_\mathsf{p}$ is trained on noisy soft-labels of $\theta_\mathsf{up}$ on $X_\mathsf{ref}$, and therefore, compared to an arbitrary test data, the influence of $X_\mathsf{ref}$ on $\theta_\mathsf{p}$ is not unique, which membership inference attacks exploit~\cite{long2018understanding,shokri2017membership,salem2019ml,nasr2019comprehensive}.
For Purchase-100 and Texas-100, the accuracy of $\theta_\mathsf{p}$ on $X_\mathsf{ref}$ is much higher than on  $D_\mathsf{test}$, because for these datasets, $X_\mathsf{ref}$ contains easy-to-classify samples.

% !TEX root = main.tex
\section{Statistical Indistinguishability due to DMP}\label{appendix:dmp_stats}

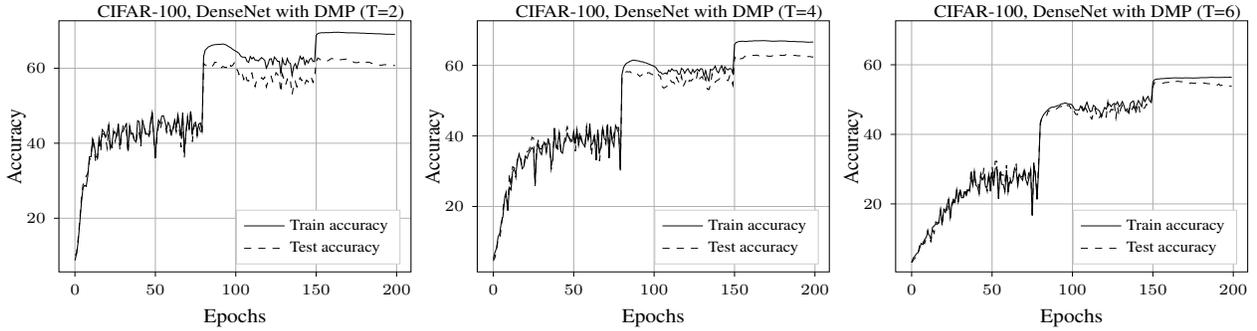
\begin{figure*}
\vspace*{-2em}
\centering
\resizebox{17cm}{2.5cm}{

\begin{tabular}{ccc}

\hspace{-1.5em}
\subfloat{% This file was created by matplotlib2tikz v0.7.5.
\begin{tikzpicture}

\pgfmathsetlengthmacro\MajorTickLength{
      \pgfkeysvalueof{/pgfplots/major tick length} * 0.5
    }
\begin{axis}[
height=5.6cm,
width=6.8cm,
legend cell align={left},
legend style={at={(0.97,0.03)}, anchor=south east, draw=white!80.0!black,font=\scriptsize},
tick align=outside,
tick pos=left,
x grid style={lightgray!92.02614379084967!black},
xlabel={Epochs},
xlabel style={font=\small},
ylabel style={font=\small,at={(axis description cs:-0.07,.5)},anchor=south},
% ylabel near ticks,
xmajorgrids,
grid=both,
xmin=-9.95, xmax=208.95,
xtick style={color=black},
xticklabel style={color=black,font=\scriptsize},
yticklabel style={color=black,font=\scriptsize},
y grid style={lightgray!92.02614379084967!black},
ylabel={Accuracy},
ymajorgrids,
ymin=5.69308322622108, ymax=72.644802377892,
ytick style={color=black},
major tick length=\MajorTickLength
]
\addplot [line width=0.44000000000000006pt, black]
table {%
0 8.73634318766067
1 10.4956619537275
2 13.6166452442159
3 18.7660668380463
4 24.0400064267352
5 28.0767994858612
6 28.7596401028278
7 28.4824871465296
8 31.3745179948586
9 36.411471722365
10 36.3793380462725
11 41.155205655527
12 39.6690231362468
13 36.3391709511568
14 36.2267030848329
15 41.4283419023136
16 39.3677699228792
17 43.6214652956298
18 43.0350257069409
19 41.6050771208226
20 42.6253213367609
21 44.8947622107969
22 38.6206619537275
23 39.9301092544987
24 43.0832262210797
25 42.9346079691517
26 42.0067480719794
27 45.1397814910026
28 39.415970437018
29 42.922557840617
30 43.1876606683805
31 44.3525064267352
32 45.1678984575835
33 38.0743894601542
34 44.5935089974293
35 42.3682519280206
36 42.540970437018
37 40.7133676092545
38 40.327763496144
39 46.0073907455013
40 45.7262210796915
41 40.1911953727506
42 46.0033740359897
43 44.3404562982005
44 42.3762853470437
45 42.5490038560411
46 42.0308483290488
47 45.7623714652956
48 48.0719794344473
49 41.9023136246787
50 36.0981683804627
51 42.7337724935733
52 46.1319087403599
53 45.2201156812339
54 47.2365038560411
55 47.348971722365
56 45.8386889460154
57 43.8745179948586
58 45.545469151671
59 44.6095758354756
60 42.4646529562982
61 44.0793701799486
62 46.2082262210797
63 44.280205655527
64 47.610057840617
65 44.8304948586118
66 42.1875
67 48.1764138817481
68 37.4879498714653
69 42.6775385604113
70 46.5255462724936
71 44.0432197943445
72 48.4375
73 42.540970437018
74 45.10764781491
75 45.5896529562982
76 42.9426413881748
77 45.0674807197943
78 45.3968508997429
79 43.3643958868895
80 63.1386568123393
81 64.8296915167095
82 65.243412596401
83 65.6892673521851
84 65.8901028277635
85 66.0588046272494
86 66.2114395886889
87 66.2837403598972
88 66.2917737789203
89 66.3801413881748
90 66.3640745501285
91 66.4082583547558
92 66.4644922879177
93 66.3640745501285
94 66.2034061696658
95 65.9302699228792
96 65.6691838046273
97 65.4362146529563
98 65.0626606683805
99 64.7855077120823
100 64.5404884318766
101 64.2914524421594
102 63.9701156812339
103 63.2832583547558
104 62.9699550128535
105 62.9579048843188
106 62.8896208226221
107 61.821176092545
108 61.5761568123393
109 63.1105398457584
110 62.4799164524422
111 62.9217544987147
112 62.1545629820051
113 62.1666131105398
114 61.9256105398458
115 62.3915488431877
116 62.8454370179949
117 62.5120501285347
118 62.1063624678663
119 60.0377570694087
120 60.993733933162
121 62.4116323907455
122 62.556233933162
123 61.0419344473008
124 62.9860218508997
125 62.4959832904884
126 61.8131426735219
127 61.8332262210797
128 58.7082262210797
129 62.5200835475578
130 61.1503856041131
131 60.881266066838
132 62.5843508997429
133 61.5199228791774
134 61.2829370179949
135 58.0856362467866
136 60.4313946015424
137 61.2186696658098
138 59.9815231362468
139 61.5480398457584
140 63.1748071979434
141 62.379498714653
142 62.8936375321337
143 61.0339010282776
144 62.1304627249357
145 61.3954048843188
146 62.379498714653
147 62.6887853470437
148 61.7850257069409
149 61.7689588688946
150 68.7861503856041
151 69.2119215938303
152 69.3926735218509
153 69.4448907455013
154 69.456940874036
155 69.4489074550129
156 69.5131748071979
157 69.4529241645244
158 69.4890745501285
159 69.4971079691517
160 69.5372750642673
161 69.5935089974293
162 69.5734254498715
163 69.5854755784062
164 69.6015424164524
165 69.5453084832905
166 69.5372750642673
167 69.5011246786632
168 69.4689910025707
169 69.4489074550129
170 69.4248071979435
171 69.4368573264781
172 69.4328406169666
173 69.4328406169666
174 69.428823907455
175 69.4127570694087
176 69.3926735218509
177 69.3766066838046
178 69.3324228791774
179 69.3444730077121
180 69.3444730077121
181 69.3002892030848
182 69.2641388174807
183 69.2721722365039
184 69.2561053984576
185 69.2400385604113
186 69.1998714652956
187 69.1878213367609
188 69.1717544987146
189 69.1797879177378
190 69.1556876606684
191 69.1476542416452
192 69.0914203084833
193 69.0994537275064
194 69.0392030848329
195 69.0673200514139
196 69.0633033419023
197 69.0432197943445
198 69.0351863753213
199 69.0351863753213
};
\addlegendentry{Train accuracy}
\addplot [line width=0.44000000000000006pt, black, dashed]
table {%
0 10.15625
1 10.6026785714286
2 14.6205357142857
3 19.53125
4 25.1116071428571
5 28.90625
6 30.1339285714286
7 31.25
8 31.9196428571429
9 35.3794642857143
10 36.3839285714286
11 40.8482142857143
12 38.28125
13 34.5982142857143
14 35.9375
15 40.0669642857143
16 37.8348214285714
17 42.4107142857143
18 42.5223214285714
19 41.6294642857143
20 38.7276785714286
21 44.1964285714286
22 38.0580357142857
23 38.28125
24 43.1919642857143
25 40.5133928571429
26 42.6339285714286
27 43.6383928571429
28 38.0580357142857
29 41.7410714285714
30 44.9776785714286
31 44.3080357142857
32 43.0803571428571
33 37.2767857142857
34 43.3035714285714
35 43.8616071428571
36 42.4107142857143
37 40.5133928571429
38 40.1785714285714
39 43.5267857142857
40 42.4107142857143
41 39.2857142857143
42 44.9776785714286
43 44.6428571428571
44 42.6339285714286
45 41.1830357142857
46 41.7410714285714
47 43.4151785714286
48 48.4375
49 41.1830357142857
50 35.2678571428571
51 41.40625
52 44.8660714285714
53 43.5267857142857
54 45.5357142857143
55 45.6473214285714
56 44.8660714285714
57 42.8571428571429
58 44.3080357142857
59 43.6383928571429
60 41.9642857142857
61 43.6383928571429
62 45.7589285714286
63 43.0803571428571
64 46.2053571428571
65 43.3035714285714
66 38.9508928571429
67 47.4330357142857
68 36.1607142857143
69 42.7455357142857
70 47.0982142857143
71 43.6383928571429
72 48.9955357142857
73 40.4017857142857
74 44.4196428571429
75 44.8660714285714
76 41.9642857142857
77 44.6428571428571
78 43.1919642857143
79 42.0758928571429
80 61.2723214285714
81 60.8258928571429
82 60.8258928571429
83 60.6026785714286
84 60.15625
85 60.3794642857143
86 60.8258928571429
87 61.6071428571429
88 61.4955357142857
89 60.6026785714286
90 60.4910714285714
91 60.15625
92 60.4910714285714
93 60.2678571428571
94 60.15625
95 61.0491071428571
96 61.3839285714286
97 60.4910714285714
98 60.7142857142857
99 61.3839285714286
100 59.7098214285714
101 59.5982142857143
102 58.3705357142857
103 59.2633928571429
104 58.4821428571429
105 57.03125
106 58.3705357142857
107 56.6964285714286
108 56.5848214285714
109 58.0357142857143
110 58.7053571428571
111 56.6964285714286
112 55.1339285714286
113 56.8080357142857
114 56.5848214285714
115 57.03125
116 57.4776785714286
117 57.03125
118 57.5892857142857
119 55.0223214285714
120 56.4732142857143
121 57.2544642857143
122 57.8125
123 56.3616071428571
124 56.8080357142857
125 58.8169642857143
126 58.3705357142857
127 57.4776785714286
128 53.5714285714286
129 58.3705357142857
130 56.9196428571429
131 55.5803571428571
132 57.03125
133 55.9151785714286
134 56.5848214285714
135 53.0133928571429
136 55.2455357142857
137 56.4732142857143
138 55.0223214285714
139 56.25
140 57.4776785714286
141 56.6964285714286
142 57.2544642857143
143 55.9151785714286
144 56.8080357142857
145 56.3616071428571
146 57.7008928571429
147 57.4776785714286
148 56.25
149 57.4776785714286
150 61.6071428571429
151 62.7232142857143
152 62.5
153 62.0535714285714
154 61.8303571428571
155 61.8303571428571
156 62.0535714285714
157 61.9419642857143
158 62.1651785714286
159 62.5
160 62.6116071428571
161 62.2767857142857
162 62.1651785714286
163 62.2767857142857
164 62.3883928571429
165 62.2767857142857
166 62.2767857142857
167 62.3883928571429
168 62.3883928571429
169 62.5
170 62.2767857142857
171 62.2767857142857
172 62.1651785714286
173 61.71875
174 61.4955357142857
175 61.4955357142857
176 61.4955357142857
177 61.71875
178 61.8303571428571
179 61.6071428571429
180 61.3839285714286
181 61.6071428571429
182 61.6071428571429
183 61.6071428571429
184 61.3839285714286
185 61.2723214285714
186 61.1607142857143
187 61.0491071428571
188 61.0491071428571
189 60.8258928571429
190 60.8258928571429
191 60.8258928571429
192 60.9375
193 60.9375
194 60.9375
195 60.9375
196 60.8258928571429
197 60.9375
198 60.7142857142857
199 60.7142857142857
};
\addlegendentry{Test accuracy}
\end{axis}

\node at ({$(current bounding box.south west)!0.6!(current bounding box.south east)$}|-{$(current bounding box.south west)!1.07!(current bounding box.north west)$})[
  scale=0.8,
  anchor=north,
  text=black,
  rotate=0.0
]{CIFAR-100, DenseNet with DMP (T=2)};
\end{tikzpicture}
}

&
\hspace{-2em}
\subfloat{% This file was created by matplotlib2tikz v0.7.5.
\begin{tikzpicture}
\pgfmathsetlengthmacro\MajorTickLength{
      \pgfkeysvalueof{/pgfplots/major tick length} * 0.5
    }
\begin{axis}[
height=5.6cm,
width=6.8cm,
legend cell align={left},
legend style={at={(0.97,0.03)}, anchor=south east, draw=white!80.0!black,font=\scriptsize},
tick align=outside,
tick pos=left,
x grid style={lightgray!92.02614379084967!black},
xlabel={Epochs},
xlabel style={font=\small},
ylabel style={font=\small,at={(axis description cs:-0.07,.5)},anchor=south},
xmajorgrids,
grid=both,
xmin=-9.95, xmax=208.95,
xtick style={color=black},
xticklabel style={color=black,font=\scriptsize},
yticklabel style={color=black,font=\scriptsize},
y grid style={lightgray!92.02614379084967!black},
ylabel={Accuracy},
ymajorgrids,
ymin=1.45334591902314, ymax=72.644802377892,
ytick style={color=black},
major tick length=\MajorTickLength
]
\addplot [line width=0.44000000000000006pt, black]
table {%
0 5.08917095115681
1 6.95292416452442
2 8.13785347043702
3 11.4154884318766
4 12.3553984575835
5 15.5004820051414
6 17.1754498714653
7 22.1883033419023
8 23.5419344473008
9 18.8303341902314
10 25.4217544987147
11 24.7148136246787
12 29.8762853470437
13 30.5028920308483
14 29.2416452442159
15 32.7321658097686
16 34.9011889460154
17 33.1258033419023
18 30.7157776349614
19 31.3263174807198
20 34.3107326478149
21 33.6198586118252
22 35.0377570694087
23 35.8892994858612
24 36.672557840617
25 36.1021850899743
26 26.3777313624679
27 35.6683804627249
28 37.6646850899743
29 35.4795951156812
30 34.7204370179949
31 37.6646850899743
32 38.0141388174807
33 35.2948264781491
34 38.8616645244216
35 37.6807519280206
36 31.5392030848329
37 33.342705655527
38 42.1674164524422
39 36.8171593830334
40 37.5241002570694
41 40.8981362467866
42 35.6241966580977
43 35.0980077120823
44 38.0261889460154
45 38.5965616966581
46 40.7776349614396
47 39.5525385604113
48 39.034383033419
49 38.4840938303342
50 41.2355398457584
51 38.002088688946
52 41.8259961439589
53 37.3594151670951
54 40.0224935732648
55 38.4439267352185
56 37.620501285347
57 36.9256105398458
58 41.3439910025707
59 38.9500321336761
60 43.4567802056555
61 37.5080334190231
62 38.0101221079692
63 39.8618251928021
64 33.1137532133676
65 41.6934447300771
66 35.0980077120823
67 37.0179948586118
68 42.1714331619537
69 39.3356362467866
70 38.5804948586118
71 43.0470758354756
72 33.7443766066838
73 42.6775385604113
74 38.5001606683805
75 37.0420951156812
76 42.9627249357326
77 41.9063303341902
78 40.7655848329049
79 30.6474935732648
80 57.40279562982
81 59.3669665809769
82 60.2144922879177
83 60.640263496144
84 60.8772493573265
85 61.1624357326478
86 61.34720437018
87 61.4797557840617
88 61.4918059125964
89 61.3391709511568
90 61.3030205655527
91 61.2026028277635
92 61.0017673521851
93 61.0499678663239
94 60.788881748072
95 60.7165809768638
96 60.5117287917738
97 60.3349935732648
98 60.1261246786632
99 59.9534061696658
100 59.7324871465296
101 59.4714010282776
102 58.7363431876607
103 58.1418701799486
104 58.0976863753213
105 57.5554305912596
106 57.6116645244216
107 57.8647172236504
108 57.5514138817481
109 58.6600257069409
110 58.2543380462725
111 59.0335796915167
112 58.5395244215938
113 58.4993573264781
114 58.5837082262211
115 57.5192802056555
116 57.5192802056555
117 59.1661311053985
118 58.8046272493573
119 57.7723329048843
120 58.5997750642674
121 59.0697300771208
122 58.1539203084833
123 57.5996143958869
124 58.844794344473
125 57.9651349614396
126 58.5716580976864
127 58.9452120822622
128 58.3748393316195
129 59.1018637532134
130 58.9612789203085
131 57.0412917737789
132 58.5355077120823
133 56.7038881748072
134 55.6876606683805
135 58.8849614395887
136 58.6600257069409
137 59.0416131105398
138 59.7967544987147
139 58.4230398457584
140 57.8928341902314
141 59.1018637532134
142 59.9011889460154
143 58.4591902313625
144 59.6119858611825
145 58.9693123393316
146 58.7202763496144
147 59.1339974293059
148 59.0255462724936
149 57.756266066838
150 65.9182197943445
151 66.4926092544987
152 66.6653277634961
153 66.77779562982
154 66.8380462724936
155 66.8581298200514
156 66.834029562982
157 66.77779562982
158 66.8259961439589
159 66.9023136246787
160 66.9023136246787
161 66.9183804627249
162 66.9625642673522
163 67.0027313624679
164 66.9625642673522
165 66.9746143958869
166 66.9946979434447
167 66.9987146529563
168 67.0268316195373
169 67.0067480719794
170 66.9625642673522
171 66.9505141388175
172 66.9183804627249
173 66.8541131105398
174 66.8741966580977
175 66.862146529563
176 66.8862467866324
177 66.9023136246787
178 66.9264138817481
179 66.9625642673522
180 66.9023136246787
181 66.9384640102828
182 66.9023136246787
183 66.8460796915167
184 66.7898457583548
185 66.7697622107969
186 66.7737789203085
187 66.7978791773779
188 66.7536953727506
189 66.713528277635
190 66.7014781491003
191 66.7095115681234
192 66.6934447300771
193 66.685411311054
194 66.6171272493573
195 66.6050771208226
196 66.5729434447301
197 66.5930269922879
198 66.6211439588689
199 66.6090938303342
};
\addlegendentry{Train accuracy}
\addplot [line width=0.44000000000000006pt, black, dashed]
table {%
0 4.57589285714286
1 5.46875
2 7.58928571428571
3 10.8258928571429
4 10.8258928571429
5 13.9508928571429
6 18.9732142857143
7 22.5446428571429
8 24.1071428571429
9 18.8616071428571
10 29.3526785714286
11 25.4464285714286
12 31.25
13 31.0267857142857
14 28.3482142857143
15 33.59375
16 34.375
17 34.1517857142857
18 30.9151785714286
19 31.3616071428571
20 34.4866071428571
21 35.4910714285714
22 35.8258928571429
23 36.9419642857143
24 39.6205357142857
25 39.3973214285714
26 25.78125
27 37.2767857142857
28 38.6160714285714
29 36.3839285714286
30 36.2723214285714
31 38.8392857142857
32 37.5
33 37.6116071428571
34 39.6205357142857
35 37.2767857142857
36 30.5803571428571
37 33.3705357142857
38 42.4107142857143
39 36.9419642857143
40 36.2723214285714
41 40.2901785714286
42 35.9375
43 37.3883928571429
44 37.9464285714286
45 40.9598214285714
46 42.5223214285714
47 39.6205357142857
48 38.3928571428571
49 38.1696428571429
50 39.7321428571429
51 35.7142857142857
52 40.5133928571429
53 38.6160714285714
54 41.40625
55 38.7276785714286
56 37.3883928571429
57 37.1651785714286
58 42.7455357142857
59 39.3973214285714
60 41.7410714285714
61 34.375
62 38.28125
63 40.7366071428571
64 33.8169642857143
65 41.9642857142857
66 32.9241071428571
67 36.4955357142857
68 42.8571428571429
69 39.0625
70 37.7232142857143
71 43.9732142857143
72 34.5982142857143
73 42.6339285714286
74 38.1696428571429
75 40.5133928571429
76 42.7455357142857
77 42.2991071428571
78 42.4107142857143
79 30.2455357142857
80 56.3616071428571
81 57.7008928571429
82 58.0357142857143
83 58.2589285714286
84 58.1473214285714
85 58.2589285714286
86 57.8125
87 57.5892857142857
88 57.9241071428571
89 57.9241071428571
90 57.3660714285714
91 56.9196428571429
92 57.1428571428571
93 57.5892857142857
94 57.3660714285714
95 57.03125
96 57.9241071428571
97 57.9241071428571
98 57.5892857142857
99 57.3660714285714
100 57.7008928571429
101 58.1473214285714
102 57.8125
103 57.1428571428571
104 56.8080357142857
105 54.0178571428571
106 53.5714285714286
107 53.7946428571429
108 54.1294642857143
109 55.2455357142857
110 55.9151785714286
111 55.2455357142857
112 55.2455357142857
113 54.5758928571429
114 56.5848214285714
115 55.46875
116 54.9107142857143
117 57.03125
118 54.0178571428571
119 54.9107142857143
120 56.0267857142857
121 54.9107142857143
122 55.8035714285714
123 55.5803571428571
124 56.1383928571429
125 56.1383928571429
126 55.46875
127 57.9241071428571
128 56.0267857142857
129 56.5848214285714
130 56.8080357142857
131 55.0223214285714
132 56.5848214285714
133 53.3482142857143
134 53.125
135 54.3526785714286
136 57.4776785714286
137 56.1383928571429
138 55.2455357142857
139 55.6919642857143
140 55.9151785714286
141 56.3616071428571
142 56.9196428571429
143 56.8080357142857
144 58.4821428571429
145 55.8035714285714
146 55.6919642857143
147 57.03125
148 56.25
149 54.4642857142857
150 62.3883928571429
151 61.1607142857143
152 61.3839285714286
153 61.71875
154 61.9419642857143
155 61.9419642857143
156 62.3883928571429
157 62.7232142857143
158 62.7232142857143
159 62.9464285714286
160 62.8348214285714
161 62.8348214285714
162 62.8348214285714
163 62.8348214285714
164 62.9464285714286
165 62.7232142857143
166 62.7232142857143
167 62.6116071428571
168 62.8348214285714
169 62.8348214285714
170 62.7232142857143
171 62.6116071428571
172 62.6116071428571
173 62.6116071428571
174 62.8348214285714
175 62.8348214285714
176 62.6116071428571
177 62.8348214285714
178 62.9464285714286
179 62.8348214285714
180 62.8348214285714
181 62.8348214285714
182 63.0580357142857
183 63.0580357142857
184 62.9464285714286
185 62.9464285714286
186 62.9464285714286
187 62.8348214285714
188 62.8348214285714
189 62.7232142857143
190 62.7232142857143
191 62.7232142857143
192 62.7232142857143
193 62.7232142857143
194 62.7232142857143
195 62.5
196 62.5
197 62.5
198 62.3883928571429
199 62.3883928571429
};
\addlegendentry{Test accuracy}
\end{axis}

\node at ({$(current bounding box.south west)!0.6!(current bounding box.south east)$}|-{$(current bounding box.south west)!1.07!(current bounding box.north west)$})[
  scale=0.8,
  anchor=north,
  text=black,
  rotate=0.0
]{CIFAR-100, DenseNet with DMP (T=4)};
\end{tikzpicture}
}

&
\hspace{-2em}
\subfloat{% This file was created by matplotlib2tikz v0.7.5.
\begin{tikzpicture}
\pgfmathsetlengthmacro\MajorTickLength{
      \pgfkeysvalueof{/pgfplots/major tick length} * 0.5
    }
\begin{axis}[
height=5.6cm,
width=6.8cm,
legend cell align={left},
legend style={at={(0.97,0.03)}, anchor=south east, draw=white!80.0!black,font=\scriptsize},
tick align=outside,
tick pos=left,
x grid style={lightgray!92.02614379084967!black},
xlabel={Epochs},
xlabel style={font=\small},
ylabel style={font=\small,at={(axis description cs:-0.07,.5)},anchor=south},
xmajorgrids,
grid=both,
xmin=-9.95, xmax=208.95,
xtick style={color=black},
xticklabel style={color=black,font=\scriptsize},
yticklabel style={color=black,font=\scriptsize},
y grid style={lightgray!92.02614379084967!black},
ylabel={Accuracy},
ymajorgrids,
ymin=0.458909061696658, ymax=72.644802377892,
ytick style={color=black},
major tick length=\MajorTickLength
]
\addplot [line width=0.44000000000000006pt, black]
table {%
0 3.18525064267352
1 4.48264781491003
2 4.95661953727506
3 5.6956940874036
4 5.9607969151671
5 7.33049485861183
6 8.65199228791774
7 8.33065552699229
8 9.70437017994859
9 10.3309768637532
10 11.6765745501285
11 10.9977506426735
12 9.88913881748072
13 13.2832583547558
14 13.6527956298201
15 14.2753856041131
16 14.1428341902314
17 15.1028277634961
18 18.2559447300771
19 14.592705655527
20 18.4889138817481
21 17.5731041131105
22 19.6095758354756
23 21.3608611825193
24 15.922236503856
25 20.1277313624679
26 21.3206940874036
27 20.5093187660668
28 24.0480398457584
29 20.609736503856
30 23.586118251928
31 21.7785989717224
32 21.7745822622108
33 23.3571658097686
34 22.4172557840617
35 24.0118894601542
36 23.8351542416452
37 27.5425771208226
38 25.6306233933162
39 29.3661632390745
40 25.381587403599
41 24.8915488431877
42 22.8430269922879
43 28.0928663239075
44 26.9199871465296
45 27.9683483290488
46 24.8071979434447
47 24.8995822622108
48 28.3258354755784
49 23.6343187660668
50 29.1372107969152
51 25.2811696658098
52 30.1172879177378
53 30.6193766066838
54 21.493412596401
55 29.0930269922879
56 25.6466902313625
57 25.5743894601542
58 23.4857005141388
59 29.6875
60 23.4214331619537
61 26.5022493573265
62 25.9840938303342
63 26.8717866323907
64 30.3261568123393
65 29.5509318766067
66 25.2369858611825
67 28.5266709511568
68 27.5385604113111
69 26.7994858611825
70 29.5830655526992
71 27.0485218508997
72 25.5623393316195
73 27.3778920308483
74 29.7959511568123
75 16.6733611825193
76 31.25
77 27.7675128534704
78 21.3006105398458
79 30.2779562982005
80 43.3724293059126
81 44.7099935732648
82 45.6057197943445
83 46.2001928020566
84 46.698264781491
85 47.1802699228792
86 47.5337403598972
87 47.8189267352185
88 47.7827763496144
89 47.9474614395887
90 48.0759961439589
91 48.2928984575835
92 48.3129820051414
93 48.7548200514139
94 48.8110539845758
95 48.9757390745501
96 48.9877892030848
97 48.7990038560411
98 48.7066195372751
99 48.6664524421594
100 48.2969151670951
101 47.4011889460154
102 46.9553341902314
103 47.07985218509
104 46.931233933162
105 48.0077120822622
106 47.9836118251928
107 46.7906491002571
108 48.1322300771208
109 49.2167416452442
110 47.7385925449871
111 46.5335796915167
112 46.6661311053985
113 48.3330655526992
114 48.0036953727506
115 47.2244537275064
116 47.0999357326478
117 46.5335796915167
118 45.0273136246787
119 46.5054627249357
120 47.7265424164524
121 47.2365038560411
122 48.558001285347
123 48.6945694087404
124 47.9675449871465
125 48.6062017994859
126 48.9315552699229
127 48.5539845758355
128 49.3251928020566
129 47.9715616966581
130 48.4133997429306
131 49.4256105398458
132 48.2848650385604
133 46.7826156812339
134 47.3570051413882
135 46.1359254498715
136 48.8793380462725
137 49.3934768637532
138 49.5661953727506
139 47.0959190231362
140 48.4053663239075
141 49.7067802056555
142 48.911471722365
143 47.9916452442159
144 50.6707904884319
145 49.4095437017995
146 50.2932197943445
147 50.9037596401028
148 49.6224293059126
149 49.1283740359897
150 55.0008033419023
151 55.7157776349614
152 55.8242287917738
153 55.9366966580977
154 55.924646529563
155 56.0210475578406
156 56.0130141388175
157 56.04514781491
158 56.0973650385604
159 56.1335154241645
160 56.1335154241645
161 56.1897493573265
162 56.2178663239075
163 56.2419665809769
164 56.2459832904884
165 56.2419665809769
166 56.2058161953727
167 56.2419665809769
168 56.193766066838
169 56.2138496143959
170 56.25
171 56.2419665809769
172 56.2299164524422
173 56.2258997429306
174 56.221883033419
175 56.1897493573265
176 56.2098329048843
177 56.2299164524422
178 56.2178663239075
179 56.278116966581
180 56.306233933162
181 56.3022172236504
182 56.306233933162
183 56.2901670951157
184 56.3343508997429
185 56.3504177377892
186 56.3825514138817
187 56.3785347043702
188 56.3745179948586
189 56.3343508997429
190 56.3624678663239
191 56.3383676092545
192 56.3544344473008
193 56.3544344473008
194 56.3865681233933
195 56.4026349614396
196 56.3664845758355
197 56.3946015424165
198 56.4468187660668
199 56.4307519280206
};
\addlegendentry{Train accuracy}
\addplot [line width=0.44000000000000006pt, black, dashed]
table {%
0 3.125
1 3.79464285714286
2 4.46428571428571
3 4.57589285714286
4 5.69196428571429
5 8.14732142857143
6 9.48660714285714
7 9.15178571428571
8 8.59375
9 10.0446428571429
10 12.3883928571429
11 10.0446428571429
12 9.15178571428571
13 12.5
14 11.6071428571429
15 14.84375
16 14.5089285714286
17 14.5089285714286
18 18.1919642857143
19 13.5044642857143
20 18.4151785714286
21 18.4151785714286
22 18.1919642857143
23 20.7589285714286
24 16.0714285714286
25 19.4196428571429
26 20.7589285714286
27 19.9776785714286
28 23.3258928571429
29 21.4285714285714
30 24.6651785714286
31 22.4330357142857
32 21.09375
33 24.5535714285714
34 22.9910714285714
35 25.1116071428571
36 23.7723214285714
37 30.1339285714286
38 25.4464285714286
39 28.4598214285714
40 25.4464285714286
41 26.2276785714286
42 24.7767857142857
43 28.7946428571429
44 27.4553571428571
45 30.46875
46 25.78125
47 23.6607142857143
48 28.90625
49 25.6696428571429
50 30.8035714285714
51 27.1205357142857
52 32.2544642857143
53 32.2544642857143
54 22.65625
55 29.9107142857143
56 28.2366071428571
57 28.3482142857143
58 26.3392857142857
59 31.1383928571429
60 24.4419642857143
61 27.9017857142857
62 25.78125
63 27.0089285714286
64 30.5803571428571
65 31.3616071428571
66 27.1205357142857
67 29.0178571428571
68 27.9017857142857
69 27.7901785714286
70 30.46875
71 27.7901785714286
72 26.6741071428571
73 29.3526785714286
74 30.46875
75 16.8526785714286
76 31.8080357142857
77 28.6830357142857
78 22.8794642857143
79 32.1428571428571
80 42.7455357142857
81 45.0892857142857
82 45.2008928571429
83 45.3125
84 45.4241071428571
85 46.7633928571429
86 46.7633928571429
87 46.6517857142857
88 47.0982142857143
89 47.0982142857143
90 47.3214285714286
91 47.65625
92 47.9910714285714
93 47.9910714285714
94 48.5491071428571
95 48.1026785714286
96 48.5491071428571
97 48.2142857142857
98 48.3258928571429
99 47.4330357142857
100 47.9910714285714
101 46.6517857142857
102 45.7589285714286
103 45.6473214285714
104 46.5401785714286
105 48.1026785714286
106 47.3214285714286
107 46.09375
108 47.65625
109 46.875
110 44.9776785714286
111 45.2008928571429
112 44.4196428571429
113 46.09375
114 47.4330357142857
115 45.6473214285714
116 44.3080357142857
117 45.0892857142857
118 45.7589285714286
119 44.53125
120 45.3125
121 44.9776785714286
122 49.21875
123 46.2053571428571
124 47.2098214285714
125 46.9866071428571
126 48.8839285714286
127 44.7544642857143
128 48.1026785714286
129 45.3125
130 47.4330357142857
131 48.3258928571429
132 47.2098214285714
133 46.7633928571429
134 46.5401785714286
135 46.5401785714286
136 48.4375
137 47.2098214285714
138 48.1026785714286
139 45.8705357142857
140 47.2098214285714
141 48.9955357142857
142 46.09375
143 46.3169642857143
144 47.4330357142857
145 47.65625
146 48.6607142857143
147 48.4375
148 48.3258928571429
149 49.1071428571429
150 53.6830357142857
151 54.9107142857143
152 54.7991071428571
153 54.7991071428571
154 54.7991071428571
155 54.6875
156 54.6875
157 54.7991071428571
158 54.9107142857143
159 54.9107142857143
160 55.0223214285714
161 54.9107142857143
162 54.6875
163 55.0223214285714
164 55.1339285714286
165 55.2455357142857
166 55.2455357142857
167 55.1339285714286
168 55.1339285714286
169 55.1339285714286
170 54.7991071428571
171 54.7991071428571
172 54.7991071428571
173 54.6875
174 54.7991071428571
175 54.6875
176 54.6875
177 54.7991071428571
178 54.9107142857143
179 54.7991071428571
180 54.7991071428571
181 54.9107142857143
182 54.6875
183 54.6875
184 54.5758928571429
185 54.5758928571429
186 54.3526785714286
187 54.2410714285714
188 54.1294642857143
189 54.0178571428571
190 54.1294642857143
191 54.4642857142857
192 54.0178571428571
193 54.1294642857143
194 53.7946428571429
195 53.90625
196 53.90625
197 53.90625
198 53.90625
199 53.90625
};
\addlegendentry{Test accuracy}
\end{axis}

\node at ({$(current bounding box.south west)!0.6!(current bounding box.south east)$}|-{$(current bounding box.south west)!1.07!(current bounding box.north west)$})[
  scale=0.8,
  anchor=north,
  text=black,
  rotate=0.0
]{CIFAR-100, DenseNet with DMP (T=6)};
\end{tikzpicture}
}
     
\end{tabular}
}

\vspace{-1em}
\caption{Impact of softmax temperature on training of $\theta_\textsf{p}$: 
Increase in the temperature of softmax layer of $\theta_\textsf{up}$ reduces $\Delta\mathcal{L}_{\scaleto{\textsf{KL}}{4pt}}$ in \eqref{eq:obj_ratio2}, and hence, the ratio $\mathcal{R}$ in \eqref{eq:obj_ratio}. This improves the membership privacy and generalization of $\theta_\textsf{p}$.}

\label{fig:training}
\vspace*{-2em}
\end{figure*}

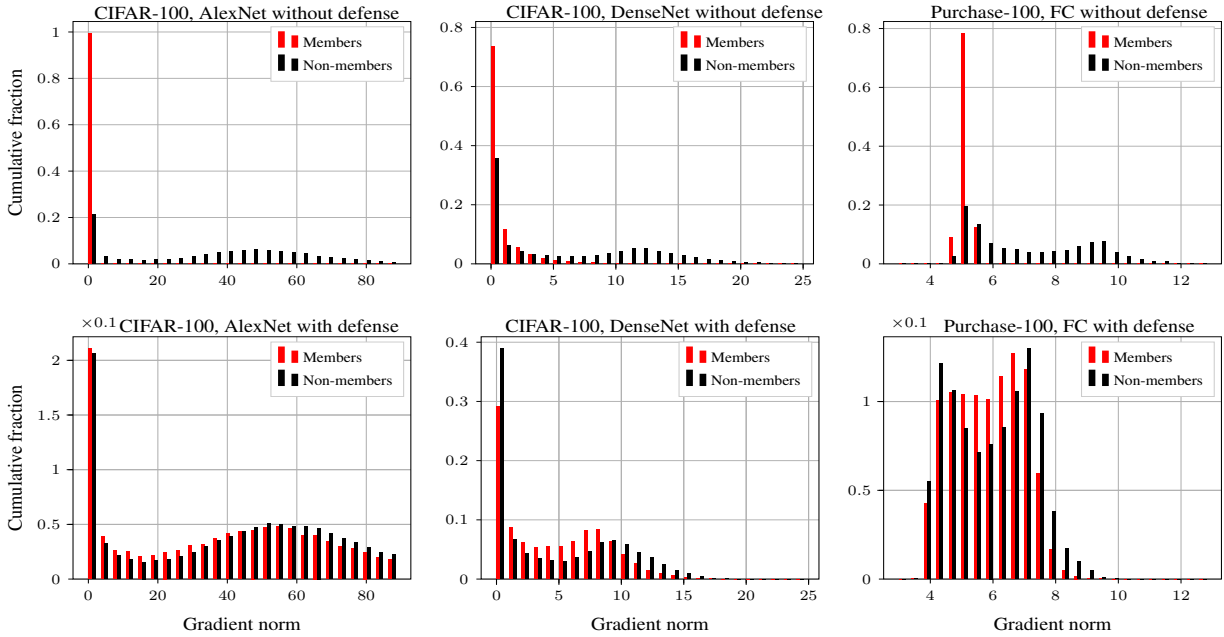
\begin{figure*}
\centering

\resizebox{16cm}{4.5cm}{
\begin{tabular}{ccc}

\hspace{-3em}
\subfloat{% This file was created by matplotlib2tikz v0.7.4.
\begin{tikzpicture}
\pgfmathsetlengthmacro\MajorTickLength{
      \pgfkeysvalueof{/pgfplots/major tick length} * 0.5
    }
\begin{axis}[
height=5.75cm,
width=6.9cm,
legend cell align={left},
legend style={draw=white!80.0!black,font=\scriptsize},
tick align=outside,
tick pos=left,
grid=both,
ticks=both,
x grid style={white!69.01960784313725!black},
ylabel={Cumulative fraction},
ylabel style={font=\small},
x label style={at={(axis description cs:0.5,-0.13)}},
xmajorgrids,
xmin=-4.42000007629395, xmax=92.8200016021729,
xtick style={color=black},
xticklabel style={color=black,font=\scriptsize},
yticklabel style={color=black,font=\scriptsize},
y grid style={white!69.01960784313725!black},
ymajorgrids,
ymin=0, ymax=1.045044,
ytick style={color=black},
major tick length=\MajorTickLength
]
\draw[fill=red,draw opacity=0] (axis cs:0,0) rectangle (axis cs:1,0.99528);
\addlegendimage{ybar,ybar legend,fill=red,draw opacity=0};
\addlegendentry{Members}

\draw[fill=red,draw opacity=0] (axis cs:3.59999990463257,0) rectangle (axis cs:4.59999990463257,0.00152);
\draw[fill=red,draw opacity=0] (axis cs:7.19999980926514,0) rectangle (axis cs:8.19999980926514,0.00016);
\draw[fill=red,draw opacity=0] (axis cs:10.8000001907349,0) rectangle (axis cs:11.8000001907349,4e-05);
\draw[fill=red,draw opacity=0] (axis cs:14.3999996185303,0) rectangle (axis cs:15.3999996185303,0);
\draw[fill=red,draw opacity=0] (axis cs:18,0) rectangle (axis cs:19,0.00016);
\draw[fill=red,draw opacity=0] (axis cs:21.6000003814697,0) rectangle (axis cs:22.6000003814697,0.00012);
\draw[fill=red,draw opacity=0] (axis cs:25.2000007629395,0) rectangle (axis cs:26.2000007629395,0.00012);
\draw[fill=red,draw opacity=0] (axis cs:28.7999992370605,0) rectangle (axis cs:29.7999992370605,0);
\draw[fill=red,draw opacity=0] (axis cs:32.4000015258789,0) rectangle (axis cs:33.4000015258789,0.00024);
\draw[fill=red,draw opacity=0] (axis cs:36,0) rectangle (axis cs:37,0.00024);
\draw[fill=red,draw opacity=0] (axis cs:39.5999984741211,0) rectangle (axis cs:40.5999984741211,0.00028);
\draw[fill=red,draw opacity=0] (axis cs:43.2000007629395,0) rectangle (axis cs:44.2000007629395,0.00012);
\draw[fill=red,draw opacity=0] (axis cs:46.7999992370605,0) rectangle (axis cs:47.7999992370605,0.00032);
\draw[fill=red,draw opacity=0] (axis cs:50.4000015258789,0) rectangle (axis cs:51.4000015258789,0.00032);
\draw[fill=red,draw opacity=0] (axis cs:54,0) rectangle (axis cs:55,0.00024);
\draw[fill=red,draw opacity=0] (axis cs:57.5999984741211,0) rectangle (axis cs:58.5999984741211,0.0002);
\draw[fill=red,draw opacity=0] (axis cs:61.2000007629395,0) rectangle (axis cs:62.2000007629395,0.00012);
\draw[fill=red,draw opacity=0] (axis cs:64.8000030517578,0) rectangle (axis cs:65.8000030517578,0.00016);
\draw[fill=red,draw opacity=0] (axis cs:68.4000015258789,0) rectangle (axis cs:69.4000015258789,0.00012);
\draw[fill=red,draw opacity=0] (axis cs:72,0) rectangle (axis cs:73,0.00012);
\draw[fill=red,draw opacity=0] (axis cs:75.5999984741211,0) rectangle (axis cs:76.5999984741211,8e-05);
\draw[fill=red,draw opacity=0] (axis cs:79.1999969482422,0) rectangle (axis cs:80.1999969482422,0);
\draw[fill=red,draw opacity=0] (axis cs:82.8000030517578,0) rectangle (axis cs:83.8000030517578,0);
\draw[fill=red,draw opacity=0] (axis cs:86.4000015258789,0) rectangle (axis cs:87.4000015258789,4e-05);
\draw[fill=black,draw opacity=0] (axis cs:1,0) rectangle (axis cs:2,0.21519759310457);
\addlegendimage{ybar,ybar legend,fill=black,draw opacity=0};
\addlegendentry{Non-members}

\draw[fill=black,draw opacity=0] (axis cs:4.59999990463257,0) rectangle (axis cs:5.59999990463257,0.0309399902423158);
\draw[fill=black,draw opacity=0] (axis cs:8.19999980926514,0) rectangle (axis cs:9.19999980926514,0.0215075622052366);
\draw[fill=black,draw opacity=0] (axis cs:11.8000001907349,0) rectangle (axis cs:12.8000001907349,0.0178077736217271);
\draw[fill=black,draw opacity=0] (axis cs:15.3999996185303,0) rectangle (axis cs:16.3999996185303,0.016384778012685);
\draw[fill=black,draw opacity=0] (axis cs:19,0) rectangle (axis cs:20,0.0175638315173199);
\draw[fill=black,draw opacity=0] (axis cs:22.6000003814697,0) rectangle (axis cs:23.6000003814697,0.0194340543177752);
\draw[fill=black,draw opacity=0] (axis cs:26.2000007629395,0) rectangle (axis cs:27.2000007629395,0.025776549032363);
\draw[fill=black,draw opacity=0] (axis cs:29.7999992370605,0) rectangle (axis cs:30.7999992370605,0.030980647259717);
\draw[fill=black,draw opacity=0] (axis cs:33.4000015258789,0) rectangle (axis cs:34.4000015258789,0.0390307367051553);
\draw[fill=black,draw opacity=0] (axis cs:37,0) rectangle (axis cs:38,0.0500487884208814);
\draw[fill=black,draw opacity=0] (axis cs:40.5999984741211,0) rectangle (axis cs:41.5999984741211,0.0531793787607741);
\draw[fill=black,draw opacity=0] (axis cs:44.2000007629395,0) rectangle (axis cs:45.2000007629395,0.0593185883883558);
\draw[fill=black,draw opacity=0] (axis cs:47.7999992370605,0) rectangle (axis cs:48.7999992370605,0.0607822410147992);
\draw[fill=black,draw opacity=0] (axis cs:51.4000015258789,0) rectangle (axis cs:52.4000015258789,0.0587087331273378);
\draw[fill=black,draw opacity=0] (axis cs:55,0) rectangle (axis cs:56,0.0551715726134331);
\draw[fill=black,draw opacity=0] (axis cs:58.5999984741211,0) rectangle (axis cs:59.5999984741211,0.0488697349162465);
\draw[fill=black,draw opacity=0] (axis cs:62.2000007629395,0) rectangle (axis cs:63.2000007629395,0.0440315498455033);
\draw[fill=black,draw opacity=0] (axis cs:65.8000030517578,0) rectangle (axis cs:66.8000030517578,0.0339486095300049);
\draw[fill=black,draw opacity=0] (axis cs:69.4000015258789,0) rectangle (axis cs:70.4000015258789,0.0285005691982436);
\draw[fill=black,draw opacity=0] (axis cs:73,0) rectangle (axis cs:74,0.0228085867620751);
\draw[fill=black,draw opacity=0] (axis cs:76.5999984741211,0) rectangle (axis cs:77.5999984741211,0.0199219385265897);
\draw[fill=black,draw opacity=0] (axis cs:80.1999969482422,0) rectangle (axis cs:81.1999969482422,0.0140266710034152);
\draw[fill=black,draw opacity=0] (axis cs:83.8000030517578,0) rectangle (axis cs:84.8000030517578,0.00955439908928281);
\draw[fill=black,draw opacity=0] (axis cs:87.4000015258789,0) rectangle (axis cs:88.4000015258789,0.00650512278419255);
\end{axis}

\node at ({$(current bounding box.south west)!0.6!(current bounding box.south east)$}|-{$(current bounding box.south west)!1.08!(current bounding box.north west)$})[
  anchor=north,
  text=black,
  rotate=0.0,
  font=\footnotesize
]{CIFAR-100, AlexNet without defense};
\end{tikzpicture}
}

&
\hspace{-1em}
\subfloat{% This file was created by matplotlib2tikz v0.7.4.
\begin{tikzpicture}
\pgfmathsetlengthmacro\MajorTickLength{
      \pgfkeysvalueof{/pgfplots/major tick length} * 0.5
    }
\begin{axis}[
height=5.75cm,
width=6.9cm,
legend cell align={left},
legend style={draw=white!80.0!black,font=\scriptsize},
tick align=outside,
tick pos=left,
grid=both,
ticks=both,
x grid style={white!69.01960784313725!black},
xlabel style={font=\small},
x label style={at={(axis description cs:0.5,-0.13)}},
xmajorgrids,
minor xtick={0,5,...,25},
xmin=-1.22999996185303, xmax=25.8299991989136,
xtick style={color=black},
xticklabel style={color=black,font=\scriptsize},
yticklabel style={color=black,font=\scriptsize},
y grid style={white!69.01960784313725!black},
ymajorgrids,
ymin=0, ymax=0.82,
ytick style={color=black},
major tick length=\MajorTickLength
]
\draw[fill=red,draw opacity=0] (axis cs:0,0) rectangle (axis cs:0.3,0.73568);
\addlegendimage{ybar,ybar legend,fill=red,draw opacity=0};
\addlegendentry{Members}

\draw[fill=red,draw opacity=0] (axis cs:1,0) rectangle (axis cs:1.3,0.11636);
\draw[fill=red,draw opacity=0] (axis cs:2,0) rectangle (axis cs:2.3,0.05584);
\draw[fill=red,draw opacity=0] (axis cs:3,0) rectangle (axis cs:3.3,0.03144);
\draw[fill=red,draw opacity=0] (axis cs:4,0) rectangle (axis cs:4.3,0.0194);
\draw[fill=red,draw opacity=0] (axis cs:5,0) rectangle (axis cs:5.3,0.013);
\draw[fill=red,draw opacity=0] (axis cs:6,0) rectangle (axis cs:6.3,0.00852);
\draw[fill=red,draw opacity=0] (axis cs:7,0) rectangle (axis cs:7.3,0.0054);
\draw[fill=red,draw opacity=0] (axis cs:8,0) rectangle (axis cs:8.3,0.00376);
\draw[fill=red,draw opacity=0] (axis cs:9,0) rectangle (axis cs:9.3,0.00216);
\draw[fill=red,draw opacity=0] (axis cs:10,0) rectangle (axis cs:10.3,0.00244);
\draw[fill=red,draw opacity=0] (axis cs:11,0) rectangle (axis cs:11.3,0.00212);
\draw[fill=red,draw opacity=0] (axis cs:12,0) rectangle (axis cs:12.3,0.00124);
\draw[fill=red,draw opacity=0] (axis cs:13,0) rectangle (axis cs:13.3,0.00064);
\draw[fill=red,draw opacity=0] (axis cs:14,0) rectangle (axis cs:14.3,0.00064);
\draw[fill=red,draw opacity=0] (axis cs:15,0) rectangle (axis cs:15.3,0.00044);
\draw[fill=red,draw opacity=0] (axis cs:16,0) rectangle (axis cs:16.3,0.00028);
\draw[fill=red,draw opacity=0] (axis cs:17,0) rectangle (axis cs:17.3,0.00028);
\draw[fill=red,draw opacity=0] (axis cs:18,0) rectangle (axis cs:18.3,0.0002);
\draw[fill=red,draw opacity=0] (axis cs:19,0) rectangle (axis cs:19.3,8e-05);
\draw[fill=red,draw opacity=0] (axis cs:20,0) rectangle (axis cs:20.3,4e-05);
\draw[fill=red,draw opacity=0] (axis cs:21,0) rectangle (axis cs:21.3,4e-05);
\draw[fill=red,draw opacity=0] (axis cs:22,0) rectangle (axis cs:22.3,0);
\draw[fill=red,draw opacity=0] (axis cs:23,0) rectangle (axis cs:23.3,0);
\draw[fill=red,draw opacity=0] (axis cs:24,0) rectangle (axis cs:24.3,0);
\draw[fill=black,draw opacity=0] (axis cs:0.300000011920929,0) rectangle (axis cs:0.600000011920929,0.356891307827757);
\addlegendimage{ybar,ybar legend,fill=black,draw opacity=0};
\addlegendentry{Non-members}

\draw[fill=black,draw opacity=0] (axis cs:1.29999995231628,0) rectangle (axis cs:1.59999995231628,0.0628701776852889);
\draw[fill=black,draw opacity=0] (axis cs:2.29999995231628,0) rectangle (axis cs:2.59999995231628,0.0436609572594846);
\draw[fill=black,draw opacity=0] (axis cs:3.29999995231628,0) rectangle (axis cs:3.59999995231628,0.0316551944933568);
\draw[fill=black,draw opacity=0] (axis cs:4.30000019073486,0) rectangle (axis cs:4.60000019073486,0.0286937730110453);
\draw[fill=black,draw opacity=0] (axis cs:5.30000019073486,0) rectangle (axis cs:5.60000019073486,0.0262926204578198);
\draw[fill=black,draw opacity=0] (axis cs:6.30000019073486,0) rectangle (axis cs:6.60000019073486,0.0261725628301585);
\draw[fill=black,draw opacity=0] (axis cs:7.30000019073486,0) rectangle (axis cs:7.60000019073486,0.0268929085961261);
\draw[fill=black,draw opacity=0] (axis cs:8.30000019073486,0) rectangle (axis cs:8.60000019073486,0.0302545221706419);
\draw[fill=black,draw opacity=0] (axis cs:9.30000019073486,0) rectangle (axis cs:9.60000019073486,0.0370978069473347);
\draw[fill=black,draw opacity=0] (axis cs:10.3000001907349,0) rectangle (axis cs:10.6000001907349,0.0436209380502641);
\draw[fill=black,draw opacity=0] (axis cs:11.3000001907349,0) rectangle (axis cs:11.6000001907349,0.0515847606851289);
\draw[fill=black,draw opacity=0] (axis cs:12.3000001907349,0) rectangle (axis cs:12.6000001907349,0.0525852409156395);
\draw[fill=black,draw opacity=0] (axis cs:13.3000001907349,0) rectangle (axis cs:13.6000001907349,0.0435408996318233);
\draw[fill=black,draw opacity=0] (axis cs:14.3000001907349,0) rectangle (axis cs:14.6000001907349,0.0366976148551305);
\draw[fill=black,draw opacity=0] (axis cs:15.3000001907349,0) rectangle (axis cs:15.6000001907349,0.0297342724507764);
\draw[fill=black,draw opacity=0] (axis cs:16.2999992370605,0) rectangle (axis cs:16.5999992370605,0.0230110453017448);
\draw[fill=black,draw opacity=0] (axis cs:17.2999992370605,0) rectangle (axis cs:17.5999992370605,0.0159276452697295);
\draw[fill=black,draw opacity=0] (axis cs:18.2999992370605,0) rectangle (axis cs:18.5999992370605,0.0126460701136546);
\draw[fill=black,draw opacity=0] (axis cs:19.2999992370605,0) rectangle (axis cs:19.5999992370605,0.00824395709940772);
\draw[fill=black,draw opacity=0] (axis cs:20.2999992370605,0) rectangle (axis cs:20.5999992370605,0.00508243957099408);
\draw[fill=black,draw opacity=0] (axis cs:21.2999992370605,0) rectangle (axis cs:21.5999992370605,0.00352169041139747);
\draw[fill=black,draw opacity=0] (axis cs:22.2999992370605,0) rectangle (axis cs:22.5999992370605,0.00160076836881703);
\draw[fill=black,draw opacity=0] (axis cs:23.2999992370605,0) rectangle (axis cs:23.5999992370605,0.00104049943973107);
\draw[fill=black,draw opacity=0] (axis cs:24.2999992370605,0) rectangle (axis cs:24.5999992370605,0.000680326556747239);
\end{axis}

\node at ({$(current bounding box.south west)!0.6!(current bounding box.south east)$}|-{$(current bounding box.south west)!1.065!(current bounding box.north west)$})[
  anchor=north,
  text=black,
  rotate=0.0,
  font=\footnotesize
]{CIFAR-100, DenseNet without defense};
\end{tikzpicture}
}

&
\hspace{-1.5em}
\subfloat{% This file was created by matplotlib2tikz v0.7.4.
\begin{tikzpicture}
\pgfmathsetlengthmacro\MajorTickLength{
      \pgfkeysvalueof{/pgfplots/major tick length} * 0.5
    }
\begin{axis}[
height=5.75cm,
width=6.9cm,
legend cell align={left},
legend style={draw=white!80.0!black,font=\scriptsize},
tick align=outside,
tick pos=left,
grid=both,
ticks=both,
x grid style={white!69.01960784313725!black},
xlabel style={font=\small},
x label style={at={(axis description cs:0.5,-0.13)}},
xmajorgrids,
% minor xtick={2.5,5,7.5,10,12.5},
xmin=2.50999996185303, xmax=13.2900008010864,
xtick style={color=black},
xticklabel style={color=black,font=\scriptsize},
yticklabel style={color=black,font=\scriptsize},
y grid style={white!69.01960784313725!black},
ymajorgrids,
ymin=0, ymax=0.82383,
ytick style={color=black},
major tick length=\MajorTickLength
]
\draw[fill=red,draw opacity=0] (axis cs:3,0) rectangle (axis cs:3.1,0);
\addlegendimage{ybar,ybar legend,fill=red,draw opacity=0};
\addlegendentry{Members}

\draw[fill=red,draw opacity=0] (axis cs:3.40000009536743,0) rectangle (axis cs:3.50000009536743,0);
\draw[fill=red,draw opacity=0] (axis cs:3.79999995231628,0) rectangle (axis cs:3.89999995231628,0);
\draw[fill=red,draw opacity=0] (axis cs:4.19999980926514,0) rectangle (axis cs:4.29999980926514,0);
\draw[fill=red,draw opacity=0] (axis cs:4.59999990463257,0) rectangle (axis cs:4.69999990463257,0.0893);
\draw[fill=red,draw opacity=0] (axis cs:5,0) rectangle (axis cs:5.1,0.7846);
\draw[fill=red,draw opacity=0] (axis cs:5.40000009536743,0) rectangle (axis cs:5.50000009536743,0.1224);
\draw[fill=red,draw opacity=0] (axis cs:5.80000019073486,0) rectangle (axis cs:5.90000019073486,0.0007);
\draw[fill=red,draw opacity=0] (axis cs:6.19999980926514,0) rectangle (axis cs:6.29999980926514,0.0004);
\draw[fill=red,draw opacity=0] (axis cs:6.59999990463257,0) rectangle (axis cs:6.69999990463257,0.0003);
\draw[fill=red,draw opacity=0] (axis cs:7,0) rectangle (axis cs:7.1,0.0003);
\draw[fill=red,draw opacity=0] (axis cs:7.40000009536743,0) rectangle (axis cs:7.50000009536743,0.0003);
\draw[fill=red,draw opacity=0] (axis cs:7.80000019073486,0) rectangle (axis cs:7.90000019073486,0.0001);
\draw[fill=red,draw opacity=0] (axis cs:8.19999980926514,0) rectangle (axis cs:8.29999980926514,0.0002);
\draw[fill=red,draw opacity=0] (axis cs:8.60000038146973,0) rectangle (axis cs:8.70000038146973,0.0004);
\draw[fill=red,draw opacity=0] (axis cs:9,0) rectangle (axis cs:9.1,0.0005);
\draw[fill=red,draw opacity=0] (axis cs:9.39999961853027,0) rectangle (axis cs:9.49999961853027,0.0001);
\draw[fill=red,draw opacity=0] (axis cs:9.80000019073486,0) rectangle (axis cs:9.90000019073486,0.0001);
\draw[fill=red,draw opacity=0] (axis cs:10.1999998092651,0) rectangle (axis cs:10.2999998092651,0.0002);
\draw[fill=red,draw opacity=0] (axis cs:10.6000003814697,0) rectangle (axis cs:10.7000003814697,0.0001);
\draw[fill=red,draw opacity=0] (axis cs:11,0) rectangle (axis cs:11.1,0);
\draw[fill=red,draw opacity=0] (axis cs:11.3999996185303,0) rectangle (axis cs:11.4999996185303,0);
\draw[fill=red,draw opacity=0] (axis cs:11.8000001907349,0) rectangle (axis cs:11.9000001907349,0);
\draw[fill=red,draw opacity=0] (axis cs:12.1999998092651,0) rectangle (axis cs:12.2999998092651,0);
\draw[fill=red,draw opacity=0] (axis cs:12.6000003814697,0) rectangle (axis cs:12.7000003814697,0);
\draw[fill=black,draw opacity=0] (axis cs:3.09999990463257,0) rectangle (axis cs:3.19999990463257,0);
\addlegendimage{ybar,ybar legend,fill=black,draw opacity=0};
\addlegendentry{Non-members}

\draw[fill=black,draw opacity=0] (axis cs:3.5,0) rectangle (axis cs:3.6,0);
\draw[fill=black,draw opacity=0] (axis cs:3.89999985694885,0) rectangle (axis cs:3.99999985694885,0);
\draw[fill=black,draw opacity=0] (axis cs:4.29999971389771,0) rectangle (axis cs:4.3999997138977,0.0011);
\draw[fill=black,draw opacity=0] (axis cs:4.69999980926514,0) rectangle (axis cs:4.79999980926514,0.0259);
\draw[fill=black,draw opacity=0] (axis cs:5.09999990463257,0) rectangle (axis cs:5.19999990463257,0.195);
\draw[fill=black,draw opacity=0] (axis cs:5.5,0) rectangle (axis cs:5.6,0.1326);
\draw[fill=black,draw opacity=0] (axis cs:5.90000009536743,0) rectangle (axis cs:6.00000009536743,0.0685);
\draw[fill=black,draw opacity=0] (axis cs:6.29999971389771,0) rectangle (axis cs:6.3999997138977,0.0512);
\draw[fill=black,draw opacity=0] (axis cs:6.69999980926514,0) rectangle (axis cs:6.79999980926514,0.0489);
\draw[fill=black,draw opacity=0] (axis cs:7.09999990463257,0) rectangle (axis cs:7.19999990463257,0.0402);
\draw[fill=black,draw opacity=0] (axis cs:7.5,0) rectangle (axis cs:7.6,0.0405);
\draw[fill=black,draw opacity=0] (axis cs:7.90000009536743,0) rectangle (axis cs:8.00000009536743,0.0423);
\draw[fill=black,draw opacity=0] (axis cs:8.30000019073486,0) rectangle (axis cs:8.40000019073486,0.0465);
\draw[fill=black,draw opacity=0] (axis cs:8.70000076293945,0) rectangle (axis cs:8.80000076293945,0.0582);
\draw[fill=black,draw opacity=0] (axis cs:9.10000038146973,0) rectangle (axis cs:9.20000038146973,0.0741);
\draw[fill=black,draw opacity=0] (axis cs:9.5,0) rectangle (axis cs:9.6,0.0771);
\draw[fill=black,draw opacity=0] (axis cs:9.90000057220459,0) rectangle (axis cs:10.0000005722046,0.0374);
\draw[fill=black,draw opacity=0] (axis cs:10.3000001907349,0) rectangle (axis cs:10.4000001907349,0.0247);
\draw[fill=black,draw opacity=0] (axis cs:10.7000007629395,0) rectangle (axis cs:10.8000007629395,0.0165);
\draw[fill=black,draw opacity=0] (axis cs:11.1000003814697,0) rectangle (axis cs:11.2000003814697,0.0094);
\draw[fill=black,draw opacity=0] (axis cs:11.5,0) rectangle (axis cs:11.6,0.0083);
\draw[fill=black,draw opacity=0] (axis cs:11.9000005722046,0) rectangle (axis cs:12.0000005722046,0.0015);
\draw[fill=black,draw opacity=0] (axis cs:12.3000001907349,0) rectangle (axis cs:12.4000001907349,0.0001);
\draw[fill=black,draw opacity=0] (axis cs:12.7000007629395,0) rectangle (axis cs:12.8000007629395,0);
\end{axis}

\node at ({$(current bounding box.south west)!0.6!(current bounding box.south east)$}|-{$(current bounding box.south west)!1.07!(current bounding box.north west)$})[
  anchor=north,
  text=black,
  rotate=0.0,
  font=\footnotesize
]{Purchase-100, FC without defense};
\end{tikzpicture}
}
    
\\[-.7ex]

\hspace{-3em}
\subfloat{% This file was created by matplotlib2tikz v0.7.4.
\begin{tikzpicture}

\pgfmathsetlengthmacro\MajorTickLength{
      \pgfkeysvalueof{/pgfplots/major tick length} * 0.5
    }
\begin{axis}[
height=5.75cm,
width=6.9cm,
legend cell align={left},
legend style={draw=white!80.0!black,font=\scriptsize},
tick align=outside,
tick pos=left,
grid=both,
ticks=both,
x grid style={white!69.01960784313725!black},
xlabel={Gradient norm},
xlabel style={font=\small},
ylabel={Cumulative fraction},
ylabel style={font=\small},
x label style={at={(axis description cs:0.5,-0.13)}},
xmajorgrids,
xmin=-4.42000007629395, xmax=92.8200016021729,
xtick style={color=black},
xticklabel style={color=black,font=\scriptsize},
yticklabel style={color=black,font=\scriptsize},
y grid style={white!69.01960784313725!black},
ymajorgrids,
minor ytick={0.05,0.1,...,0.2},
scaled y ticks={real:0.1},
ytick scale label code/.code={$\times 0.1$},
ymin=0, ymax=0.221535022354694,
ytick style={color=black},
major tick length=\MajorTickLength
]

\draw[fill=red,draw opacity=0] (axis cs:0,0) rectangle (axis cs:1,0.2109857355759);
\addlegendimage{ybar,ybar legend,fill=red,draw opacity=0};
\addlegendentry{Members}

\draw[fill=red,draw opacity=0] (axis cs:3.59999990463257,0) rectangle (axis cs:4.59999990463257,0.0391313604428359);
\draw[fill=red,draw opacity=0] (axis cs:7.19999980926514,0) rectangle (axis cs:8.19999980926514,0.0264849904194167);
\draw[fill=red,draw opacity=0] (axis cs:10.8000001907349,0) rectangle (axis cs:11.8000001907349,0.0255482222695337);
\draw[fill=red,draw opacity=0] (axis cs:14.3999996185303,0) rectangle (axis cs:15.3999996185303,0.020438577815627);
\draw[fill=red,draw opacity=0] (axis cs:18,0) rectangle (axis cs:19,0.0218437300404514);
\draw[fill=red,draw opacity=0] (axis cs:21.6000003814697,0) rectangle (axis cs:22.6000003814697,0.0247391952309985);
\draw[fill=red,draw opacity=0] (axis cs:25.2000007629395,0) rectangle (axis cs:26.2000007629395,0.0261017670853736);
\draw[fill=red,draw opacity=0] (axis cs:28.7999992370605,0) rectangle (axis cs:29.7999992370605,0.0305301256120928);
\draw[fill=red,draw opacity=0] (axis cs:32.4000015258789,0) rectangle (axis cs:33.4000015258789,0.0317649563551203);
\draw[fill=red,draw opacity=0] (axis cs:36,0) rectangle (axis cs:37,0.0370023419203747);
\draw[fill=red,draw opacity=0] (axis cs:39.5999984741211,0) rectangle (axis cs:40.5999984741211,0.0417287630402385);
\draw[fill=red,draw opacity=0] (axis cs:43.2000007629395,0) rectangle (axis cs:44.2000007629395,0.0434319778582074);
\draw[fill=red,draw opacity=0] (axis cs:46.7999992370605,0) rectangle (axis cs:47.7999992370605,0.0448371300830317);
\draw[fill=red,draw opacity=0] (axis cs:50.4000015258789,0) rectangle (axis cs:51.4000015258789,0.0472216308281882);
\draw[fill=red,draw opacity=0] (axis cs:54,0) rectangle (axis cs:55,0.0480306578667234);
\draw[fill=red,draw opacity=0] (axis cs:57.5999984741211,0) rectangle (axis cs:58.5999984741211,0.0461997019374069);
\draw[fill=red,draw opacity=0] (axis cs:61.2000007629395,0) rectangle (axis cs:62.2000007629395,0.0396849052586758);
\draw[fill=red,draw opacity=0] (axis cs:64.8000030517578,0) rectangle (axis cs:65.8000030517578,0.0403661911858633);
\draw[fill=red,draw opacity=0] (axis cs:68.4000015258789,0) rectangle (axis cs:69.4000015258789,0.0345326804343198);
\draw[fill=red,draw opacity=0] (axis cs:72,0) rectangle (axis cs:73,0.0297210985735576);
\draw[fill=red,draw opacity=0] (axis cs:75.5999984741211,0) rectangle (axis cs:76.5999984741211,0.0276772407919949);
\draw[fill=red,draw opacity=0] (axis cs:79.1999969482422,0) rectangle (axis cs:80.1999969482422,0.0247817756014477);
\draw[fill=red,draw opacity=0] (axis cs:82.8000030517578,0) rectangle (axis cs:83.8000030517578,0.0196295507770918);
\draw[fill=red,draw opacity=0] (axis cs:86.4000015258789,0) rectangle (axis cs:87.4000015258789,0.0175856929955291);
\draw[fill=black,draw opacity=0] (axis cs:1,0) rectangle (axis cs:2,0.206315881555351);
\addlegendimage{ybar,ybar legend,fill=black,draw opacity=0};
\addlegendentry{Non-members}

\draw[fill=black,draw opacity=0] (axis cs:4.59999990463257,0) rectangle (axis cs:5.59999990463257,0.0323667060315794);
\draw[fill=black,draw opacity=0] (axis cs:8.19999980926514,0) rectangle (axis cs:9.19999980926514,0.021606963215676);
\draw[fill=black,draw opacity=0] (axis cs:11.8000001907349,0) rectangle (axis cs:12.8000001907349,0.0180641210689761);
\draw[fill=black,draw opacity=0] (axis cs:15.3999996185303,0) rectangle (axis cs:16.3999996185303,0.0156147487206403);
\draw[fill=black,draw opacity=0] (axis cs:19,0) rectangle (axis cs:20,0.0168831736867428);
\draw[fill=black,draw opacity=0] (axis cs:22.6000003814697,0) rectangle (axis cs:23.6000003814697,0.017670471941565);
\draw[fill=black,draw opacity=0] (axis cs:26.2000007629395,0) rectangle (axis cs:27.2000007629395,0.020994620128592);
\draw[fill=black,draw opacity=0] (axis cs:29.7999992370605,0) rectangle (axis cs:30.7999992370605,0.02423129073175);
\draw[fill=black,draw opacity=0] (axis cs:33.4000015258789,0) rectangle (axis cs:34.4000015258789,0.0297423785155054);
\draw[fill=black,draw opacity=0] (axis cs:37,0) rectangle (axis cs:38,0.0349910335476534);
\draw[fill=black,draw opacity=0] (axis cs:40.5999984741211,0) rectangle (axis cs:41.5999984741211,0.0391899575733718);
\draw[fill=black,draw opacity=0] (axis cs:44.2000007629395,0) rectangle (axis cs:45.2000007629395,0.0436950531426322);
\draw[fill=black,draw opacity=0] (axis cs:47.7999992370605,0) rectangle (axis cs:48.7999992370605,0.0473691116651358);
\draw[fill=black,draw opacity=0] (axis cs:51.4000015258789,0) rectangle (axis cs:52.4000015258789,0.0504308271005555);
\draw[fill=black,draw opacity=0] (axis cs:55,0) rectangle (axis cs:56,0.0496872676376678);
\draw[fill=black,draw opacity=0] (axis cs:58.5999984741211,0) rectangle (axis cs:59.5999984741211,0.0477627607925469);
\draw[fill=black,draw opacity=0] (axis cs:62.2000007629395,0) rectangle (axis cs:63.2000007629395,0.0478064995844815);
\draw[fill=black,draw opacity=0] (axis cs:65.8000030517578,0) rectangle (axis cs:66.8000030517578,0.0467130297861173);
\draw[fill=black,draw opacity=0] (axis cs:69.4000015258789,0) rectangle (axis cs:70.4000015258789,0.0419455014652495);
\draw[fill=black,draw opacity=0] (axis cs:73,0) rectangle (axis cs:74,0.0367405852250361);
\draw[fill=black,draw opacity=0] (axis cs:76.5999984741211,0) rectangle (axis cs:77.5999984741211,0.0337663473734855);
\draw[fill=black,draw opacity=0] (axis cs:80.1999969482422,0) rectangle (axis cs:81.1999969482422,0.0290862966364869);
\draw[fill=black,draw opacity=0] (axis cs:83.8000030517578,0) rectangle (axis cs:84.8000030517578,0.0246249398591611);
\draw[fill=black,draw opacity=0] (axis cs:87.4000015258789,0) rectangle (axis cs:88.4000015258789,0.0227004330140402);
\end{axis}

\node at ({$(current bounding box.south west)!0.63!(current bounding box.south east)$}|-{$(current bounding box.south west)!1.02!(current bounding box.north west)$})[
  anchor=north,
  text=black,
  rotate=0.0,
  font=\footnotesize
]{CIFAR-100, AlexNet with defense};
\end{tikzpicture}
}
&
\hspace{-1em}
\subfloat{% This file was created by matplotlib2tikz v0.7.4.
\begin{tikzpicture}
\pgfmathsetlengthmacro\MajorTickLength{
      \pgfkeysvalueof{/pgfplots/major tick length} * 0.5
    }
\begin{axis}[
height=5.75cm,
width=6.9cm,legend cell align={left},
legend style={draw=white!80.0!black,font=\scriptsize},
tick align=outside,
tick pos=left,
grid=both,
ticks=both,
x grid style={white!69.01960784313725!black},
xlabel={Gradient norm},
xlabel style={font=\small},
x label style={at={(axis description cs:0.5,-0.13)}},
xmajorgrids,
minor xtick={0,5,...,25},
xmin=-1.22999996185303, xmax=25.8299991989136,
xtick style={color=black},
xticklabel style={color=black,font=\scriptsize},
yticklabel style={color=black,font=\scriptsize},
y grid style={white!69.01960784313725!black},
ymajorgrids,
ymin=0, ymax=0.409206,
ytick style={color=black},
major tick length=\MajorTickLength
]
\draw[fill=red,draw opacity=0] (axis cs:0,0) rectangle (axis cs:0.3,0.292);
\addlegendimage{ybar,ybar legend,fill=red,draw opacity=0};
\addlegendentry{Members}

\draw[fill=red,draw opacity=0] (axis cs:1,0) rectangle (axis cs:1.3,0.08732);
\draw[fill=red,draw opacity=0] (axis cs:2,0) rectangle (axis cs:2.3,0.06192);
\draw[fill=red,draw opacity=0] (axis cs:3,0) rectangle (axis cs:3.3,0.05356);
\draw[fill=red,draw opacity=0] (axis cs:4,0) rectangle (axis cs:4.3,0.05532);
\draw[fill=red,draw opacity=0] (axis cs:5,0) rectangle (axis cs:5.3,0.05464);
\draw[fill=red,draw opacity=0] (axis cs:6,0) rectangle (axis cs:6.3,0.06348);
\draw[fill=red,draw opacity=0] (axis cs:7,0) rectangle (axis cs:7.3,0.08236);
\draw[fill=red,draw opacity=0] (axis cs:8,0) rectangle (axis cs:8.3,0.08408);
\draw[fill=red,draw opacity=0] (axis cs:9,0) rectangle (axis cs:9.3,0.06352);
\draw[fill=red,draw opacity=0] (axis cs:10,0) rectangle (axis cs:10.3,0.04128);
\draw[fill=red,draw opacity=0] (axis cs:11,0) rectangle (axis cs:11.3,0.02592);
\draw[fill=red,draw opacity=0] (axis cs:12,0) rectangle (axis cs:12.3,0.01456);
\draw[fill=red,draw opacity=0] (axis cs:13,0) rectangle (axis cs:13.3,0.00936);
\draw[fill=red,draw opacity=0] (axis cs:14,0) rectangle (axis cs:14.3,0.00548);
\draw[fill=red,draw opacity=0] (axis cs:15,0) rectangle (axis cs:15.3,0.00272);
\draw[fill=red,draw opacity=0] (axis cs:16,0) rectangle (axis cs:16.3,0.00164);
\draw[fill=red,draw opacity=0] (axis cs:17,0) rectangle (axis cs:17.3,0.00048);
\draw[fill=red,draw opacity=0] (axis cs:18,0) rectangle (axis cs:18.3,0.00016);
\draw[fill=red,draw opacity=0] (axis cs:19,0) rectangle (axis cs:19.3,0.00016);
\draw[fill=red,draw opacity=0] (axis cs:20,0) rectangle (axis cs:20.3,0);
\draw[fill=red,draw opacity=0] (axis cs:21,0) rectangle (axis cs:21.3,4e-05);
\draw[fill=red,draw opacity=0] (axis cs:22,0) rectangle (axis cs:22.3,0);
\draw[fill=red,draw opacity=0] (axis cs:23,0) rectangle (axis cs:23.3,0);
\draw[fill=red,draw opacity=0] (axis cs:24,0) rectangle (axis cs:24.3,0);
\draw[fill=black,draw opacity=0] (axis cs:0.300000011920929,0) rectangle (axis cs:0.600000011920929,0.38972);
\addlegendimage{ybar,ybar legend,fill=black,draw opacity=0};
\addlegendentry{Non-members}

\draw[fill=black,draw opacity=0] (axis cs:1.29999995231628,0) rectangle (axis cs:1.59999995231628,0.06704);
\draw[fill=black,draw opacity=0] (axis cs:2.29999995231628,0) rectangle (axis cs:2.59999995231628,0.04356);
\draw[fill=black,draw opacity=0] (axis cs:3.29999995231628,0) rectangle (axis cs:3.59999995231628,0.0348);
\draw[fill=black,draw opacity=0] (axis cs:4.30000019073486,0) rectangle (axis cs:4.60000019073486,0.03076);
\draw[fill=black,draw opacity=0] (axis cs:5.30000019073486,0) rectangle (axis cs:5.60000019073486,0.02972);
\draw[fill=black,draw opacity=0] (axis cs:6.30000019073486,0) rectangle (axis cs:6.60000019073486,0.03632);
\draw[fill=black,draw opacity=0] (axis cs:7.30000019073486,0) rectangle (axis cs:7.60000019073486,0.0466);
\draw[fill=black,draw opacity=0] (axis cs:8.30000019073486,0) rectangle (axis cs:8.60000019073486,0.0622);
\draw[fill=black,draw opacity=0] (axis cs:9.30000019073486,0) rectangle (axis cs:9.60000019073486,0.06584);
\draw[fill=black,draw opacity=0] (axis cs:10.3000001907349,0) rectangle (axis cs:10.6000001907349,0.05792);
\draw[fill=black,draw opacity=0] (axis cs:11.3000001907349,0) rectangle (axis cs:11.6000001907349,0.0452);
\draw[fill=black,draw opacity=0] (axis cs:12.3000001907349,0) rectangle (axis cs:12.6000001907349,0.03588);
\draw[fill=black,draw opacity=0] (axis cs:13.3000001907349,0) rectangle (axis cs:13.6000001907349,0.02412);
\draw[fill=black,draw opacity=0] (axis cs:14.3000001907349,0) rectangle (axis cs:14.6000001907349,0.01472);
\draw[fill=black,draw opacity=0] (axis cs:15.3000001907349,0) rectangle (axis cs:15.6000001907349,0.00908);
\draw[fill=black,draw opacity=0] (axis cs:16.2999992370605,0) rectangle (axis cs:16.5999992370605,0.00416);
\draw[fill=black,draw opacity=0] (axis cs:17.2999992370605,0) rectangle (axis cs:17.5999992370605,0.00164);
\draw[fill=black,draw opacity=0] (axis cs:18.2999992370605,0) rectangle (axis cs:18.5999992370605,0.00036);
\draw[fill=black,draw opacity=0] (axis cs:19.2999992370605,0) rectangle (axis cs:19.5999992370605,0.00024);
\draw[fill=black,draw opacity=0] (axis cs:20.2999992370605,0) rectangle (axis cs:20.5999992370605,0.00012);
\draw[fill=black,draw opacity=0] (axis cs:21.2999992370605,0) rectangle (axis cs:21.5999992370605,0);
\draw[fill=black,draw opacity=0] (axis cs:22.2999992370605,0) rectangle (axis cs:22.5999992370605,0);
\draw[fill=black,draw opacity=0] (axis cs:23.2999992370605,0) rectangle (axis cs:23.5999992370605,0);
\draw[fill=black,draw opacity=0] (axis cs:24.2999992370605,0) rectangle (axis cs:24.5999992370605,0);
\end{axis}

\node at ({$(current bounding box.south west)!0.55!(current bounding box.south east)$}|-{$(current bounding box.south west)!1.06!(current bounding box.north west)$})[
  anchor=north,
  text=black,
  rotate=0.0,
  font=\footnotesize
]{CIFAR-100, DenseNet with defense};
\end{tikzpicture}
}
&
\hspace{-1.5em}
\subfloat{% This file was created by matplotlib2tikz v0.7.4.
\begin{tikzpicture}
\pgfmathsetlengthmacro\MajorTickLength{
      \pgfkeysvalueof{/pgfplots/major tick length} * 0.5
    }
\begin{axis}[
height=5.75cm,
width=6.9cm,
legend cell align={left},
legend style={draw=white!80.0!black,font=\scriptsize},
tick align=outside,
tick pos=left,
grid=both,
ticks=both,
x grid style={white!69.01960784313725!black},
xlabel={Gradient norm},
xlabel style={font=\small},
x label style={at={(axis description cs:0.5,-0.13)}},
xmajorgrids,
% minor xtick={2.5,5,7.5,10,12.5},
xmin=2.50999996185303, xmax=13.2900008010864,
xtick style={color=black},
xticklabel style={color=black,font=\scriptsize},
yticklabel style={color=black,font=\scriptsize},
y grid style={white!69.01960784313725!black},
ymajorgrids,
minor ytick={0.05,0.1,...,0.2},
scaled y ticks={real:0.1},
ytick scale label code/.code={$\times 0.1$},
ymin=0, ymax=0.1365,
ytick style={color=black},
major tick length=\MajorTickLength
]
\draw[fill=red,draw opacity=0] (axis cs:3,0) rectangle (axis cs:3.1,0);
\addlegendimage{ybar,ybar legend,fill=red,draw opacity=0};
\addlegendentry{Members}

\draw[fill=red,draw opacity=0] (axis cs:3.40000009536743,0) rectangle (axis cs:3.50000009536743,0.0002);
\draw[fill=red,draw opacity=0] (axis cs:3.79999995231628,0) rectangle (axis cs:3.89999995231628,0.0426);
\draw[fill=red,draw opacity=0] (axis cs:4.19999980926514,0) rectangle (axis cs:4.29999980926514,0.1008);
\draw[fill=red,draw opacity=0] (axis cs:4.59999990463257,0) rectangle (axis cs:4.69999990463257,0.1053);
\draw[fill=red,draw opacity=0] (axis cs:5,0) rectangle (axis cs:5.1,0.1039);
\draw[fill=red,draw opacity=0] (axis cs:5.40000009536743,0) rectangle (axis cs:5.50000009536743,0.1035);
\draw[fill=red,draw opacity=0] (axis cs:5.80000019073486,0) rectangle (axis cs:5.90000019073486,0.1013);
\draw[fill=red,draw opacity=0] (axis cs:6.19999980926514,0) rectangle (axis cs:6.29999980926514,0.1139);
\draw[fill=red,draw opacity=0] (axis cs:6.59999990463257,0) rectangle (axis cs:6.69999990463257,0.1273);
\draw[fill=red,draw opacity=0] (axis cs:7,0) rectangle (axis cs:7.1,0.118);
\draw[fill=red,draw opacity=0] (axis cs:7.40000009536743,0) rectangle (axis cs:7.50000009536743,0.0595);
\draw[fill=red,draw opacity=0] (axis cs:7.80000019073486,0) rectangle (axis cs:7.90000019073486,0.0168);
\draw[fill=red,draw opacity=0] (axis cs:8.19999980926514,0) rectangle (axis cs:8.29999980926514,0.0047);
\draw[fill=red,draw opacity=0] (axis cs:8.60000038146973,0) rectangle (axis cs:8.70000038146973,0.0017);
\draw[fill=red,draw opacity=0] (axis cs:9,0) rectangle (axis cs:9.1,0.0003);
\draw[fill=red,draw opacity=0] (axis cs:9.39999961853027,0) rectangle (axis cs:9.49999961853027,0.0002);
\draw[fill=red,draw opacity=0] (axis cs:9.80000019073486,0) rectangle (axis cs:9.90000019073486,0);
\draw[fill=red,draw opacity=0] (axis cs:10.1999998092651,0) rectangle (axis cs:10.2999998092651,0);
\draw[fill=red,draw opacity=0] (axis cs:10.6000003814697,0) rectangle (axis cs:10.7000003814697,0);
\draw[fill=red,draw opacity=0] (axis cs:11,0) rectangle (axis cs:11.1,0);
\draw[fill=red,draw opacity=0] (axis cs:11.3999996185303,0) rectangle (axis cs:11.4999996185303,0);
\draw[fill=red,draw opacity=0] (axis cs:11.8000001907349,0) rectangle (axis cs:11.9000001907349,0);
\draw[fill=red,draw opacity=0] (axis cs:12.1999998092651,0) rectangle (axis cs:12.2999998092651,0);
\draw[fill=red,draw opacity=0] (axis cs:12.6000003814697,0) rectangle (axis cs:12.7000003814697,0);
\draw[fill=black,draw opacity=0] (axis cs:3.09999990463257,0) rectangle (axis cs:3.19999990463257,0);
\addlegendimage{ybar,ybar legend,fill=black,draw opacity=0};
\addlegendentry{Non-members}

\draw[fill=black,draw opacity=0] (axis cs:3.5,0) rectangle (axis cs:3.6,0.0003);
\draw[fill=black,draw opacity=0] (axis cs:3.89999985694885,0) rectangle (axis cs:3.99999985694885,0.0551);
\draw[fill=black,draw opacity=0] (axis cs:4.29999971389771,0) rectangle (axis cs:4.3999997138977,0.1215);
\draw[fill=black,draw opacity=0] (axis cs:4.69999980926514,0) rectangle (axis cs:4.79999980926514,0.106);
\draw[fill=black,draw opacity=0] (axis cs:5.09999990463257,0) rectangle (axis cs:5.19999990463257,0.085);
\draw[fill=black,draw opacity=0] (axis cs:5.5,0) rectangle (axis cs:5.6,0.0712);
\draw[fill=black,draw opacity=0] (axis cs:5.90000009536743,0) rectangle (axis cs:6.00000009536743,0.0761);
\draw[fill=black,draw opacity=0] (axis cs:6.29999971389771,0) rectangle (axis cs:6.3999997138977,0.0854);
\draw[fill=black,draw opacity=0] (axis cs:6.69999980926514,0) rectangle (axis cs:6.79999980926514,0.1055);
\draw[fill=black,draw opacity=0] (axis cs:7.09999990463257,0) rectangle (axis cs:7.19999990463257,0.13);
\draw[fill=black,draw opacity=0] (axis cs:7.5,0) rectangle (axis cs:7.6,0.0931);
\draw[fill=black,draw opacity=0] (axis cs:7.90000009536743,0) rectangle (axis cs:8.00000009536743,0.0379);
\draw[fill=black,draw opacity=0] (axis cs:8.30000019073486,0) rectangle (axis cs:8.40000019073486,0.0171);
\draw[fill=black,draw opacity=0] (axis cs:8.70000076293945,0) rectangle (axis cs:8.80000076293945,0.0098);
\draw[fill=black,draw opacity=0] (axis cs:9.10000038146973,0) rectangle (axis cs:9.20000038146973,0.0049);
\draw[fill=black,draw opacity=0] (axis cs:9.5,0) rectangle (axis cs:9.6,0.0009);
\draw[fill=black,draw opacity=0] (axis cs:9.90000057220459,0) rectangle (axis cs:10.0000005722046,0.0002);
\draw[fill=black,draw opacity=0] (axis cs:10.3000001907349,0) rectangle (axis cs:10.4000001907349,0);
\draw[fill=black,draw opacity=0] (axis cs:10.7000007629395,0) rectangle (axis cs:10.8000007629395,0);
\draw[fill=black,draw opacity=0] (axis cs:11.1000003814697,0) rectangle (axis cs:11.2000003814697,0);
\draw[fill=black,draw opacity=0] (axis cs:11.5,0) rectangle (axis cs:11.6,0);
\draw[fill=black,draw opacity=0] (axis cs:11.9000005722046,0) rectangle (axis cs:12.0000005722046,0);
\draw[fill=black,draw opacity=0] (axis cs:12.3000001907349,0) rectangle (axis cs:12.4000001907349,0);
\draw[fill=black,draw opacity=0] (axis cs:12.7000007629395,0) rectangle (axis cs:12.8000007629395,0);
\end{axis}

\node at ({$(current bounding box.south west)!0.6!(current bounding box.south east)$}|-{$(current bounding box.south west)!1.02!(current bounding box.north west)$})[
  anchor=north,
  text=black,
  rotate=0.0,
  font=\footnotesize
]{Purchase-100, FC with defense};
\end{tikzpicture}
}
     
\end{tabular}}
\vspace{-1em}
\caption{
Distributions of gradient norms of members and non-members of private training data.
(\emph{Upper row}): Unlike the distribution of non-members, that of the members of the unprotected model, $\theta_\textsf{up}$,  is skewed towards 0 as $\theta_\textsf{up}$ memorizes the members.
(\emph{Lower row}): The  distributions of gradient norms for members and non-members for the protected model, $\theta_\textsf{p}$, of DMP are almost indistinguishable.
}

\label{fig:gradients}
\vspace*{-2em}
\end{figure*}

\begin{figure*}
\centering
\resizebox{17cm}{2.5cm}
{\begin{tabular}{ccc}

\hspace{-2.5em}
\subfloat{% This file was created by matplotlib2tikz v0.7.5.
\begin{tikzpicture}
\pgfmathsetlengthmacro\MajorTickLength{
      \pgfkeysvalueof{/pgfplots/major tick length} * 0.5
    }
\begin{axis}[
height=5.75cm,
width=6.9cm,
legend cell align={left},
legend style={at={(0.97,0.03)}, anchor=south east, draw=white!80.0!black,font=\scriptsize},
tick align=outside,
tick pos=left,
x grid style={lightgray!92.02614379084967!black},
xlabel={Generalization error},
xlabel style={font=\small},
ylabel style={font=\small},
xmajorgrids,
grid=both,
xmin=-0.0725806451612904, xmax=0.789170506912442,
xtick style={color=black},
xticklabel style={color=black,font=\scriptsize},
yticklabel style={color=black,font=\scriptsize},
y grid style={lightgray!92.02614379084967!black},
ylabel={Cumulative fraction},
ymajorgrids,
minor ytick={0.25,0.5,0.75,1},
ymin=-0.0495, ymax=1.0395,
ytick style={color=black},
major tick length=\MajorTickLength
]
\addplot [line width=0.44000000000000006pt, black]
table {%
0.0786516853932584 0
0.108108108108108 0.01
0.119266055045872 0.02
0.12719298245614 0.03
0.128440366972477 0.04
0.137037037037037 0.05
0.138121330724071 0.06
0.142857142857143 0.07
0.14344262295082 0.08
0.147058823529412 0.09
0.149319788510664 0.1
0.153225806451613 0.11
0.153846153846154 0.12
0.15625 0.13
0.156626506024096 0.14
0.158536585365854 0.15
0.170940170940171 0.16
0.173913043478261 0.17
0.177631578947368 0.18
0.184465061192262 0.19
0.186440677966102 0.2
0.19047619047619 0.21
0.191176470588235 0.22
0.193236714975845 0.23
0.198581560283688 0.24
0.204819277108434 0.25
0.205249841872233 0.26
0.206349206349206 0.27
0.211764705882353 0.28
0.213114754098361 0.29
0.214285714285714 0.3
0.215827338129496 0.31
0.216981132075472 0.32
0.217700729927007 0.33
0.21978021978022 0.34
0.220930232558139 0.35
0.231578947368421 0.36
0.232142857142857 0.37
0.234313725490196 0.38
0.235142818464871 0.39
0.235294117647059 0.4
0.236363636363636 0.41
0.238095238095238 0.42
0.238461538461539 0.43
0.242718446601942 0.44
0.245151640189808 0.45
0.245614035087719 0.46
0.25 0.47
0.251748251748252 0.48
0.254545454545455 0.49
0.258064516129032 0.5
0.261146496815287 0.51
0.261363636363636 0.52
0.264367816091954 0.53
0.264705882352941 0.54
0.265060240963855 0.55
0.26530612244898 0.56
0.266666666666667 0.57
0.267175572519084 0.58
0.268817204301075 0.59
0.27009900990099 0.6
0.272727272727273 0.61
0.273972602739726 0.62
0.280373831775701 0.63
0.280487804878049 0.64
0.28169014084507 0.65
0.283333333333333 0.66
0.285714285714286 0.67
0.287234042553192 0.68
0.288888888888889 0.69
0.3 0.7
0.305555555555556 0.71
0.306451612903226 0.72
0.306451612903226 0.73
0.313253012048193 0.74
0.321428571428571 0.75
0.32258064516129 0.76
0.32258064516129 0.77
0.323076923076923 0.78
0.325842696629214 0.79
0.329896907216495 0.8
0.333333333333333 0.81
0.333333333333333 0.82
0.340659340659341 0.83
0.347368421052632 0.84
0.35 0.85
0.355140186915888 0.86
0.357142857142857 0.87
0.361111111111111 0.88
0.362745098039216 0.89
0.365384615384615 0.9
0.366666666666667 0.91
0.384615384615385 0.92
0.388888888888889 0.93
0.392857142857143 0.94
0.424242424242424 0.95
0.431818181818182 0.96
0.434343434343434 0.97
0.434782608695652 0.98
0.75 0.99
};
\addlegendentry{No Defense}
\addplot [line width=0.44000000000000006pt, black, mark=*, mark size=1, mark repeat=10, mark options={solid}]
table {%
-0.033410138248848 0
-0.00403225806451613 0.01
0 0.02
0.000835421888053411 0.03
0.00384872824631854 0.04
0.00785163124407739 0.05
0.0127211290001987 0.06
0.0232198142414861 0.07
0.0279794313369631 0.08
0.0291875 0.09
0.0353033268101761 0.1
0.0361471861471861 0.11
0.0405486885740126 0.12
0.0427272727272727 0.13
0.0449438202247191 0.14
0.0461760461760462 0.15
0.0485059110081388 0.16
0.0520087739552066 0.17
0.0583554376657824 0.18
0.059178433889602 0.19
0.0602404428299602 0.2
0.0631513316112942 0.21
0.0648148148148148 0.22
0.0700854700854701 0.23
0.0705882352941176 0.24
0.0716876466823129 0.25
0.0720086936627773 0.26
0.0729372285827082 0.27
0.0729832886321437 0.28
0.0756572899430042 0.29
0.0757095410628019 0.3
0.0769980506822612 0.31
0.0821904851981463 0.32
0.0823762376237623 0.33
0.0862023305084746 0.34
0.0868812737971617 0.35
0.0869565217391305 0.36
0.0884814176716712 0.37
0.0886568940336452 0.38
0.0889914304548451 0.39
0.0897435897435898 0.4
0.0898395721925134 0.41
0.0909090909090909 0.42
0.0916825142628366 0.43
0.0917431192660551 0.44
0.0919811320754716 0.45
0.0938256658595642 0.46
0.0957906098219105 0.47
0.0960876153740184 0.48
0.0962485790071997 0.49
0.0980392156862745 0.5
0.0989836500220946 0.51
0.100449162923642 0.52
0.105799185238438 0.53
0.107088989441931 0.54
0.11051509524792 0.55
0.110969387755102 0.56
0.111111111111111 0.57
0.114261301848687 0.58
0.115023474178404 0.59
0.115510948905109 0.6
0.117730496453901 0.61
0.1187195251219 0.62
0.120809614168248 0.63
0.12258064516129 0.64
0.123341139734582 0.65
0.126696832579186 0.66
0.131194295900178 0.67
0.132485289027345 0.68
0.133454106280193 0.69
0.135811556864188 0.7
0.136554621848739 0.71
0.137158469945355 0.72
0.141544903016321 0.73
0.15851814516129 0.74
0.160191787558397 0.75
0.160477453580902 0.76
0.166666666666667 0.77
0.171758241758242 0.78
0.179616424883206 0.79
0.186363636363636 0.8
0.186677631578947 0.81
0.197222222222222 0.82
0.1996996996997 0.83
0.201589864613092 0.84
0.201595744680851 0.85
0.203846153846154 0.86
0.205651491365777 0.87
0.214802065404475 0.88
0.231120280070018 0.89
0.232142857142857 0.9
0.232868757259001 0.91
0.233134920634921 0.92
0.240514075887393 0.93
0.241240758598521 0.94
0.253787878787879 0.95
0.253846153846154 0.96
0.257575757575758 0.97
0.28030303030303 0.98
0.295757575757576 0.99
};
\addlegendentry{DMP}
\addplot [line width=0.44000000000000006pt, black, dashed]
table {%
0 0
0.106999923035481 0.01
0.110091743119266 0.02
0.118572292800968 0.03
0.125518590998043 0.04
0.126559714795009 0.05
0.128710199676201 0.06
0.128787878787879 0.07
0.153102189781022 0.08
0.154704504752721 0.09
0.155555555555556 0.1
0.157303370786517 0.11
0.163636363636364 0.12
0.167435728411338 0.13
0.169642857142857 0.14
0.172298106877546 0.15
0.176646390773883 0.16
0.177008641294356 0.17
0.177272727272727 0.18
0.180961310698776 0.19
0.181818181818182 0.2
0.184225577671014 0.21
0.185408299866131 0.22
0.189154125583809 0.23
0.196557971014493 0.24
0.198731173922014 0.25
0.199637616586467 0.26
0.203271028037383 0.27
0.203463203463203 0.28
0.204184322033898 0.29
0.204825834542816 0.3
0.20595361716857 0.31
0.207268192749621 0.32
0.219117647058824 0.33
0.222523219814242 0.34
0.223893065998329 0.35
0.227787716159809 0.36
0.229233511586453 0.37
0.229708853238265 0.38
0.230197444831591 0.39
0.230990099009901 0.4
0.234345351043643 0.41
0.235941644562334 0.42
0.238115581792761 0.43
0.242857142857143 0.44
0.243955211770433 0.45
0.248894783377542 0.46
0.24957264957265 0.47
0.251260080645161 0.48
0.256828885400314 0.49
0.259302628155087 0.5
0.262795864308297 0.51
0.263324393095385 0.52
0.264030612244898 0.53
0.264875 0.54
0.268627450980392 0.55
0.269191270860077 0.56
0.275493727487269 0.57
0.27570564516129 0.58
0.278601694915254 0.59
0.278681318681319 0.6
0.278702518139138 0.61
0.278889342588812 0.62
0.279589371980676 0.63
0.280405405405405 0.64
0.284758689672418 0.65
0.288619426550461 0.66
0.29438982070561 0.67
0.294503546099291 0.68
0.294545454545455 0.69
0.296166549894905 0.7
0.299309597942331 0.71
0.301809210526316 0.72
0.302777777777778 0.73
0.303605717943247 0.74
0.304029304029304 0.75
0.318579234972678 0.76
0.321475916924437 0.77
0.331088664421998 0.78
0.334308510638298 0.79
0.334487135110188 0.8
0.334545454545455 0.81
0.336612125224629 0.82
0.338709677419355 0.83
0.339075630252101 0.84
0.342948717948718 0.85
0.347430538480452 0.86
0.348484848484849 0.87
0.358506731946144 0.88
0.361538461538462 0.89
0.371526924022621 0.9
0.372805507745267 0.91
0.374444444444444 0.92
0.375 0.93
0.383484162895928 0.94
0.398585663773706 0.95
0.416269841269841 0.96
0.417108331566674 0.97
0.434782608695652 0.98
0.462121212121212 0.99
};
\addlegendentry{AdvReg}
\end{axis}

\node at ({$(current bounding box.south west)!0.6!(current bounding box.south east)$}|-{$(current bounding box.south west)!1.06!(current bounding box.north west)$})[
  scale=0.8,
  anchor=north,
  text=black,
  rotate=0.0
]{Purchase-100, Fully Connected};
\end{tikzpicture}
}

&
\hspace{-1.5em}
\subfloat{% This file was created by matplotlib2tikz v0.7.5.
\begin{tikzpicture}
\pgfmathsetlengthmacro\MajorTickLength{
      \pgfkeysvalueof{/pgfplots/major tick length} * 0.5
    }
\begin{axis}[
height=5.75cm,
width=6.9cm,
legend cell align={left},
legend style={at={(0.97,0.03)}, anchor=south east, draw=white!80.0!black,font=\scriptsize},
tick align=outside,
tick pos=left,
x grid style={lightgray!92.02614379084967!black},
xlabel={Generalization error},
xlabel style={font=\small},
xmajorgrids,
grid=both,
xmin=-0.0770407712868131, xmax=0.941011966552505,
xtick style={color=black},
xticklabel style={color=black,font=\scriptsize},
yticklabel style={color=black,font=\scriptsize},
y grid style={lightgray!92.02614379084967!black},
ymajorgrids,
minor ytick={0.25,0.5,0.75,1},
ymin=-0.0495, ymax=1.0395,
ytick style={color=black},
major tick length=\MajorTickLength
]
\addplot [line width=0.44000000000000006pt, black]
table {%
0.229829225892218 0
0.251012145748988 0.01
0.318548387096774 0.02
0.32319391634981 0.03
0.340163934426229 0.04
0.357621951219512 0.05
0.37117903930131 0.06
0.377682403433476 0.07
0.39344262295082 0.08
0.398636743021423 0.09
0.405529953917051 0.1
0.411352357320099 0.11
0.411698079014136 0.12
0.419354838709677 0.13
0.441605839416058 0.14
0.449799196787149 0.15
0.480782781372109 0.16
0.484615384615385 0.17
0.4921875 0.18
0.5 0.19
0.505902887972072 0.2
0.509433962264151 0.21
0.510579310814029 0.22
0.514056224899598 0.23
0.517416714465097 0.24
0.523255813953488 0.25
0.526748971193416 0.26
0.546938775510204 0.27
0.547244094488189 0.28
0.549336753848079 0.29
0.556862745098039 0.3
0.578125 0.31
0.597560975609756 0.32
0.615686274509804 0.33
0.623316362698975 0.34
0.628205128205128 0.35
0.630434782608696 0.36
0.632231404958678 0.37
0.633204633204633 0.38
0.636 0.39
0.638314758527524 0.4
0.639850622406639 0.41
0.646570151574038 0.42
0.650980392156863 0.43
0.655606214116852 0.44
0.65587044534413 0.45
0.66260162601626 0.46
0.673469387755102 0.47
0.67680608365019 0.48
0.682767581614423 0.49
0.698795995670996 0.5
0.7 0.51
0.709433962264151 0.52
0.7125 0.53
0.713260545584138 0.54
0.719827586206897 0.55
0.724 0.56
0.727642276422764 0.57
0.727659574468085 0.58
0.727969348659004 0.59
0.729885369696311 0.6
0.732946017913338 0.61
0.738047427806464 0.62
0.738771798267307 0.63
0.739622641509434 0.64
0.741082382172529 0.65
0.741702446670756 0.66
0.744192099954448 0.67
0.745386137329311 0.68
0.74818320610687 0.69
0.751679254378917 0.7
0.752880787764509 0.71
0.758064516129032 0.72
0.759967741935484 0.73
0.762069481880803 0.74
0.763110531868677 0.75
0.766539234108728 0.76
0.773662551440329 0.77
0.774326419848808 0.78
0.77601652992278 0.79
0.781307173049019 0.8
0.782852834063307 0.81
0.794979933337868 0.82
0.795149606299213 0.83
0.796295425696958 0.84
0.797635369453361 0.85
0.797665369649806 0.86
0.803846153846154 0.87
0.810368679532406 0.88
0.814516129032258 0.89
0.816896325459318 0.9
0.819496736947791 0.91
0.819894522811031 0.92
0.838820660221235 0.93
0.85593220338983 0.94
0.856140070354973 0.95
0.877046619172603 0.96
0.884178863805507 0.97
0.885979124474217 0.98
0.894736842105263 0.99
};
\addlegendentry{No Defense}
\addplot [line width=0.44000000000000006pt, black, mark=*, mark size=1, mark repeat=10, mark options={solid}]
table {%
-0.0307656468395714 0
-0.0107296137339056 0.01
-0.00938604950176797 0.02
-0.00520532099479476 0.03
-0.00495744856647112 0.04
-0.00155038759689924 0.05
0.0048371647509578 0.06
0.00827583709096177 0.07
0.00857818304626817 0.08
0.00905707196029781 0.09
0.00977857176437402 0.1
0.0144905660377359 0.11
0.0184067900603333 0.12
0.0216905604170545 0.13
0.0268675455116133 0.14
0.0281303934061325 0.15
0.0294372294372294 0.16
0.0303250849102378 0.17
0.0317816256680885 0.18
0.032963284544778 0.19
0.0330532212885154 0.2
0.0340994781769372 0.21
0.0365068002863278 0.22
0.0398788725832751 0.23
0.0402022917866974 0.24
0.041869918699187 0.25
0.0426993665746305 0.26
0.0456737457108412 0.27
0.045751857228247 0.28
0.0464500523012552 0.29
0.0469018932874355 0.3
0.0473700787401575 0.31
0.047874159352559 0.32
0.049947472500309 0.33
0.0501018994049075 0.34
0.0507775309425579 0.35
0.052099173553719 0.36
0.0526519301411356 0.37
0.054153945042584 0.38
0.0544838709677419 0.39
0.0545038288466093 0.4
0.0576828063241107 0.41
0.058691304819632 0.42
0.0593467575709529 0.43
0.06 0.44
0.0608972178763349 0.45
0.0609465492277992 0.46
0.0630032025380863 0.47
0.0631263154588006 0.48
0.0631496062992126 0.49
0.0639595418711885 0.5
0.0642394205108655 0.51
0.0690290456431535 0.52
0.0700123112810212 0.53
0.0702258064516129 0.54
0.0707648913223826 0.55
0.0708870601956846 0.56
0.0714490628586 0.57
0.0715179577024153 0.58
0.0721700141268964 0.59
0.0738340322785614 0.6
0.0749403865461847 0.61
0.0770180282605165 0.62
0.0782295823775258 0.63
0.0784774845396742 0.64
0.0790621226351805 0.65
0.0802808302808303 0.66
0.0808590072913509 0.67
0.0808985391234741 0.68
0.0832738034493316 0.69
0.0848896706108091 0.7
0.0858379336195734 0.71
0.0867577695935905 0.72
0.0877811650325451 0.73
0.0882594158097281 0.74
0.0898325209348832 0.75
0.0900051820183962 0.76
0.0902676903921976 0.77
0.0920185684336628 0.78
0.0930668016194332 0.79
0.0931453497429506 0.8
0.093936678614098 0.81
0.0966883980409328 0.82
0.09775582940679 0.83
0.102661661255411 0.84
0.10327868852459 0.85
0.108980876301138 0.86
0.112801013941698 0.87
0.117679389312977 0.88
0.120389344262295 0.89
0.12061790668348 0.9
0.121040189125295 0.91
0.121474884774627 0.92
0.123523622047244 0.93
0.144066688460281 0.94
0.162292387957305 0.95
0.164410979794129 0.96
0.171245862656255 0.97
0.174089068825911 0.98
0.199262164938825 0.99
};
\addlegendentry{DMP}
\addplot [line width=0.44000000000000006pt, black, dashed]
table {%
0.244490358126722 0
0.30671007702692 0.01
0.313593073593074 0.02
0.315644600875412 0.03
0.340011614401858 0.04
0.353687020353687 0.05
0.364640198511166 0.06
0.366972786818826 0.07
0.368445444915254 0.08
0.375581284995729 0.09
0.380154304979253 0.1
0.382307927762473 0.11
0.386100386100386 0.12
0.386692131398014 0.13
0.396306252489048 0.14
0.397487634450813 0.15
0.416632935303245 0.16
0.417442414093155 0.17
0.421072796934866 0.18
0.430689736162217 0.19
0.431801402731635 0.2
0.436974789915966 0.21
0.448377383740381 0.22
0.454158564187574 0.23
0.456088560885609 0.24
0.458580564345992 0.25
0.459251640709178 0.26
0.465396825396825 0.27
0.477014829142489 0.28
0.47840625 0.29
0.478693958146013 0.3
0.486150367439231 0.31
0.496484235766825 0.32
0.496780981094152 0.33
0.498487426734732 0.34
0.505191497698109 0.35
0.508255045660977 0.36
0.508754052802223 0.37
0.510499878963931 0.38
0.5110640814167 0.39
0.514611455381019 0.4
0.517833333333333 0.41
0.52026199227217 0.42
0.521701149425287 0.43
0.522030651340996 0.44
0.522557611165206 0.45
0.525145856744816 0.46
0.525764775764776 0.47
0.526024983984625 0.48
0.531378650847677 0.49
0.532297590974245 0.5
0.533710204343091 0.51
0.535121328224776 0.52
0.537452168367347 0.53
0.538229571984436 0.54
0.538352941176471 0.55
0.539006546014632 0.56
0.541517978588917 0.57
0.543831168831169 0.58
0.544570215776441 0.59
0.547844040339871 0.6
0.553382761816497 0.61
0.55719696969697 0.62
0.563927496122381 0.63
0.565863949778738 0.64
0.571143743274891 0.65
0.571907071907072 0.66
0.572210485884386 0.67
0.582123695730884 0.68
0.584752284752285 0.69
0.584883166428231 0.7
0.586250358977632 0.71
0.588181818181818 0.72
0.594102752593319 0.73
0.595582329317269 0.74
0.598425196850394 0.75
0.600401606425703 0.76
0.600433416768861 0.77
0.611750454270139 0.78
0.61358024691358 0.79
0.617136978248089 0.8
0.617269833420898 0.81
0.617858529388928 0.82
0.62560076371058 0.83
0.627257245678298 0.84
0.62775 0.85
0.628043535794627 0.86
0.629166666666667 0.87
0.63043263043263 0.88
0.639837398373984 0.89
0.6409825468649 0.9
0.643683331191364 0.91
0.644243755593524 0.92
0.648864076416677 0.93
0.650309331778577 0.94
0.652694809504845 0.95
0.665806118497483 0.96
0.670140136543299 0.97
0.682820208580528 0.98
0.689522900134997 0.99
};
\addlegendentry{AdvReg}
\end{axis}

\node at ({$(current bounding box.south west)!0.6!(current bounding box.south east)$}|-{$(current bounding box.south west)!1.06!(current bounding box.north west)$})[
  scale=0.8,
  anchor=north,
  text=black,
  rotate=0.0
]{CIFAR-100, AlexNet};
\end{tikzpicture}
}

&
\hspace{-1.5em}
\subfloat{% This file was created by matplotlib2tikz v0.7.5.
\begin{tikzpicture}
\pgfmathsetlengthmacro\MajorTickLength{
      \pgfkeysvalueof{/pgfplots/major tick length} * 0.5
    }
\begin{axis}[
height=5.75cm,
width=6.9cm,
legend cell align={left},
legend style={at={(0.97,0.03)}, anchor=south east, draw=white!80.0!black,font=\scriptsize},
tick align=outside,
tick pos=left,
x grid style={lightgray!92.02614379084967!black},
xlabel={Generalization error},
xlabel style={font=\small},
xmajorgrids,
grid=both,
xmin=-0.0835217409564695, xmax=0.706579876618951,
xtick style={color=black},
xticklabel style={color=black,font=\scriptsize},
yticklabel style={color=black,font=\scriptsize},
y grid style={lightgray!92.02614379084967!black},
ymajorgrids,
minor ytick={0.25,0.5,0.75,1},
ymin=-0.0495, ymax=1.0395,
ytick style={color=black},
major tick length=\MajorTickLength
]
\addplot [line width=0.44000000000000006pt, black]
table {%
0.0753857273041512 0
0.0815500289184499 0.01
0.121747967479675 0.02
0.137096774193548 0.03
0.139069264069264 0.04
0.15450643776824 0.05
0.156182488245012 0.06
0.163561377971858 0.07
0.166486829577912 0.08
0.168858560794045 0.09
0.170791151106112 0.1
0.171257225880244 0.11
0.176670870113493 0.12
0.184251230372627 0.13
0.192824972821867 0.14
0.198039215686275 0.15
0.209677419354839 0.16
0.211981566820276 0.17
0.21304347826087 0.18
0.213980463980464 0.19
0.23855261568201 0.2
0.239366410435593 0.21
0.242915943803789 0.22
0.245136186770428 0.23
0.260275073106462 0.24
0.26151962363859 0.25
0.264150943396226 0.26
0.272993207397491 0.27
0.280201454205857 0.28
0.286953647855903 0.29
0.290076335877863 0.3
0.29269089958159 0.31
0.29682868083004 0.32
0.297749136079201 0.33
0.305220883534137 0.34
0.311190826980975 0.35
0.314183246520366 0.36
0.318771653543307 0.37
0.320951417004049 0.38
0.321138211382114 0.39
0.323633052890695 0.4
0.325699745547074 0.41
0.332 0.42
0.335459718478586 0.43
0.336986910662092 0.44
0.337110346146491 0.45
0.341616544450853 0.46
0.345846688759711 0.47
0.348489231758277 0.48
0.354117131773245 0.49
0.356475429678044 0.5
0.358478483606557 0.51
0.358520669464078 0.52
0.362141732283465 0.53
0.362204724409449 0.54
0.364933880528956 0.55
0.368 0.56
0.371647509578544 0.57
0.373096611157908 0.58
0.376 0.59
0.379010469436001 0.6
0.38 0.61
0.380934649041147 0.62
0.388826446280992 0.63
0.390499967991806 0.64
0.414212397692205 0.65
0.419848807908509 0.66
0.424139595210087 0.67
0.426340996168582 0.68
0.426595952365458 0.69
0.428454541079078 0.7
0.428511568046452 0.71
0.434022309711286 0.72
0.437779135912745 0.73
0.439545130564375 0.74
0.445055301482274 0.75
0.448562611661523 0.76
0.451884413548695 0.77
0.45580720459277 0.78
0.456062141168524 0.79
0.457618917044373 0.8
0.460091991341991 0.81
0.47678218994237 0.82
0.492021106491845 0.83
0.499967741935484 0.84
0.541803278688525 0.85
0.544311052123552 0.86
0.544622922351774 0.87
0.544909040603168 0.88
0.54841935483871 0.89
0.570588415944967 0.9
0.579740440728159 0.91
0.581073446327684 0.92
0.589908849738113 0.93
0.615384615384615 0.94
0.631369932647096 0.95
0.654947916666667 0.96
0.656239289608973 0.97
0.66112438603478 0.98
0.670666166729159 0.99
};
\addlegendentry{No Defense}
\addplot [line width=0.44000000000000006pt, black, mark=*, mark size=1, mark repeat=10, mark options={solid}]
table {%
-0.0476080310666777 0
-0.0415812017374517 0.01
-0.0408560311284046 0.02
-0.0401327800829875 0.03
-0.0331265508684864 0.04
-0.0278688524590164 0.05
-0.0158980505961769 0.06
-0.0145769861729552 0.07
-0.012 0.08
-0.0118583797155225 0.09
-0.0101521196198632 0.1
-0.00963995354239255 0.11
-0.00840235675860301 0.12
-0.00725890079502245 0.13
-0.00686572199730084 0.14
-0.00663568924757485 0.15
-0.0055433070866141 0.16
-0.00310528869480842 0.17
-0.00266671139138608 0.18
-0.00181429404522782 0.19
-0.00169280944245997 0.2
-0.00145974066309507 0.21
0.00140458015267175 0.22
0.00326428123038291 0.23
0.0076319367198121 0.24
0.00832529733204768 0.25
0.0101214574898785 0.26
0.0107677030968782 0.27
0.0151009304769854 0.28
0.0153021495445572 0.29
0.0157866872021948 0.3
0.016987134282983 0.31
0.0193200392285061 0.32
0.0195360195360196 0.33
0.0228014070994563 0.34
0.0245619834710744 0.35
0.0246498599439776 0.36
0.026118033728481 0.37
0.0263986158029167 0.38
0.0269464956600596 0.39
0.0280539772727273 0.4
0.0288776796973519 0.41
0.0294386655711784 0.42
0.0299294914591848 0.43
0.0306438967856291 0.44
0.0306906638901284 0.45
0.0308044525144076 0.46
0.0348661589382692 0.47
0.036206896551724 0.48
0.0366946244757911 0.49
0.0376644469709274 0.5
0.0382943253138075 0.51
0.0389435695538057 0.52
0.0392276422764227 0.53
0.0403225806451613 0.54
0.0417637795275591 0.55
0.046188679245283 0.56
0.0478663496433488 0.57
0.047877748125556 0.58
0.0488633394303633 0.59
0.0494618805181414 0.6
0.0514745670995671 0.61
0.0528304303278688 0.62
0.0536426241391809 0.63
0.0546190995482162 0.64
0.0547284652088146 0.65
0.0564944625824211 0.66
0.0566607745460389 0.67
0.0568310544537869 0.68
0.0568385378842895 0.69
0.0572192684645442 0.7
0.0581395348837209 0.71
0.0607426213900349 0.72
0.0608715184186882 0.73
0.0635779161138891 0.74
0.0678218164211109 0.75
0.0702729871516979 0.76
0.0722409698787652 0.77
0.0731772396771396 0.78
0.0765155702966648 0.79
0.0765692087507667 0.8
0.0809692671394799 0.81
0.0845708415055572 0.82
0.0876953609526172 0.83
0.0878180516691611 0.84
0.0906222108886905 0.85
0.0908444789041805 0.86
0.0921304264677758 0.87
0.0942136979200415 0.88
0.0950136182694322 0.89
0.103101762988532 0.9
0.10463573906537 0.91
0.105782036706188 0.92
0.112088980923695 0.93
0.112193548387097 0.94
0.113424518743668 0.95
0.117408906882591 0.96
0.120088790233074 0.97
0.143665014614309 0.98
0.175942519580512 0.99
};
\addlegendentry{DMP}
\addplot [line width=0.44000000000000006pt, black, dashed]
table {%
0.0486352357320099 0
0.0515118292896071 0.01
0.0653812364174412 0.02
0.0682251082251082 0.03
0.0716122918640856 0.04
0.0897930897930898 0.05
0.0913003177966102 0.06
0.0925583117363938 0.07
0.0926248548199768 0.08
0.0931986931986932 0.09
0.100661162253646 0.1
0.104531028428658 0.11
0.104927022279138 0.12
0.10607858649789 0.13
0.110632891288811 0.14
0.11608071562305 0.15
0.122245179063361 0.16
0.135839357429719 0.17
0.137295898048586 0.18
0.145051579626048 0.19
0.150921066172687 0.2
0.151901070505722 0.21
0.154440154440154 0.22
0.156443093842703 0.23
0.16067986120036 0.24
0.161210727969349 0.25
0.161944275787627 0.26
0.162183711658937 0.27
0.162862246923382 0.28
0.164032567049808 0.29
0.16423494201272 0.3
0.165943468296409 0.31
0.16706256273001 0.32
0.18016821602479 0.33
0.182777463727153 0.34
0.182877875559673 0.35
0.183879658733722 0.36
0.186260500902881 0.37
0.188031764038571 0.38
0.188078431372549 0.39
0.193199920350458 0.4
0.195944460151415 0.41
0.196315449256626 0.42
0.19653730950265 0.43
0.196842510589141 0.44
0.199604877716466 0.45
0.200379234356583 0.46
0.200999118424919 0.47
0.204632391078838 0.48
0.205001133676675 0.49
0.2090625 0.5
0.212802933673469 0.51
0.216535433070866 0.52
0.216786226685796 0.53
0.217809810941524 0.54
0.219129298374581 0.55
0.219319902561064 0.56
0.222357328145266 0.57
0.222655998971788 0.58
0.224218177181294 0.59
0.225718390804598 0.6
0.2265 0.61
0.226991093326449 0.62
0.233118027011157 0.63
0.233712121212121 0.64
0.233858379452249 0.65
0.23697448796263 0.66
0.238271604938272 0.67
0.238561105376003 0.68
0.239215290396393 0.69
0.247089947089947 0.7
0.251841013602598 0.71
0.255552529653837 0.72
0.256129569169707 0.73
0.259836686312601 0.74
0.260455004270518 0.75
0.261649670840325 0.76
0.262270081852417 0.77
0.265285126396238 0.78
0.267329628546359 0.79
0.268029633611638 0.8
0.271103896103896 0.81
0.272471999744727 0.82
0.273141273141273 0.83
0.273293172690763 0.84
0.273484180624939 0.85
0.274316779312371 0.86
0.274620620745644 0.87
0.282391649648287 0.88
0.285983290081651 0.89
0.304522368541103 0.9
0.317925537085642 0.91
0.319450449152135 0.92
0.323078155028828 0.93
0.329090909090909 0.94
0.335974320833007 0.95
0.338253968253968 0.96
0.342948717948718 0.97
0.35625 0.98
0.397524181666773 0.99
};
\addlegendentry{AdvReg}
\end{axis}

\node at ({$(current bounding box.south west)!0.6!(current bounding box.south east)$}|-{$(current bounding box.south west)!1.06!(current bounding box.north west)$})[
  scale=0.8,
  anchor=north,
  text=black,
  rotate=0.0
]{CIFAR-100, DenseNet};
\end{tikzpicture}
}
     
\end{tabular}}
\vspace{-1em}

\caption{The empirical CDF of the generalization error of models trained with DMP, adversarial regularization (AdvReg), and without defense. The y-axis is the fraction of classes that have generalization error less than the values on x-axis. The generalization error reduction due to DMP is much larger ($10\times$ for CIFAR100 and $2\times$ for Purchase) than due to AdvReg. 
The low generalization error improves membership privacy due to DMP.
}
\label{fig:generalization_error}
\vspace*{-2em}
\end{figure*}
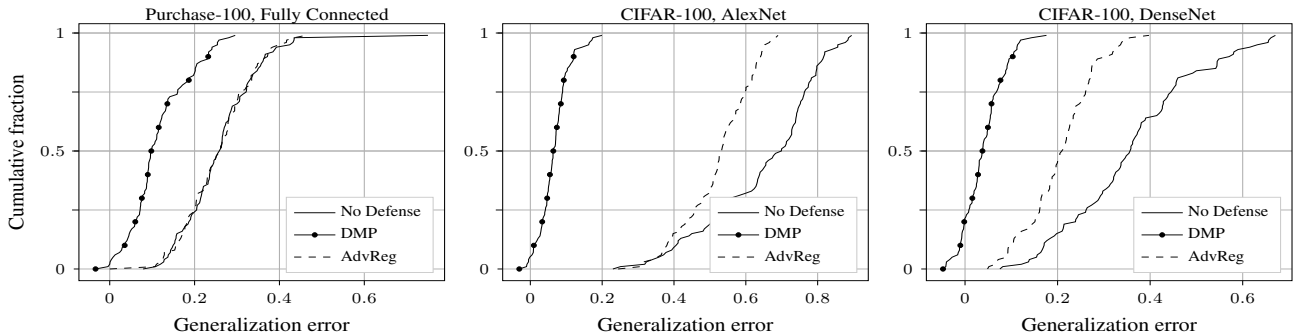

\begin{table*}
\fontsize{8.5}{9}\selectfont{}

\begin{center}
\setlength{\extrarowheight}{0.01cm}
\hspace{-2em}

\begin{tabular}{ |c|c|c|c|c|c|c|c|c| } 

\hline

\multicolumn{3}{|c|}{{Experimental setup}} & \multicolumn{6}{c|}{{Near-equal $A_\mathsf{test}$ as DMP}} \\ \hline

Dataset & Model & Regularization & $E_\text{gen}$ & $A_\text{test}$ & $A_\textsf{wb}$ & $A_\textsf{bb}$ & $A_\textsf{bl}$ & $A_\textsf{nn}$  \\ \hline \hline

\multirow{4}{*}{Purchase} & \multirow{4}{*}{FC} & WD &  21.7 &  78.1 &  69.7 &  70.1 &  60.9 &  55.6 \\ 
& & WD + DR &  22.1 &  77.4 &  77.1 &  76.8 &  61.5 &  60.0  \\ 
& & WD + LS &  21.1 &  78.4 &  76.5 &  76.8 &  60.6 &  56.4 \\ 
& & WD + CP &  22.9 &  76.9 &  70.1 &  70.5 &  61.5 &  58.5 \\ \hline\hline

\multirow{4}{*}{Texas} & \multirow{4}{*}{FC} & WD &  49.0 &  50.4 &  84.1 &  82.1 &  74.5 &  56.2 \\ 
& & WD + DR &  41.1 &  52.1 &  82.1 &  81.2 &  70.6 &  60.2 \\ 
& & WD + LS &  50.9 &  49.1 &  86.0 &  85.7 &  75.5 &  56.9 \\ 
& & WD + CP &  45.5 &  54.2 &  90.4 &  90.2 &  72.8 &  65.6 \\ \hline\hline

\multirow{4}{*} {CIFAR100} &  \multirow{4}{*} {DenseNet12} & WD &  31.0 &  67.8 &  72.9 &  72.9 &  65.5 & N/A \\ 
& & WD + DR &  31.0 &  68.2 &  73.7 &  73.6 &  65.5 & N/A  \\ 
& & WD + LS &  31.6 &  68.0 &  70.3 &  70.1  &  65.8 & N/A \\ 
& & WD + CP &  31.1 &  67.5 &  74.3 &  74.7 &  65.6 & N/A  \\ \hline\hline

\multirow{4}{*} {CIFAR10} &  \multirow{4}{*} {AlexNet} & WD &  31.0 &  68.9 &  73.2 &  73.3 &  65.5 & N/A  \\ 
& & WD + DR &  30.6 &  69.4 &  73.8 &  73.4 &  65.3 & N/A \\ 
& & WD + LS &  29.9 &  69.9 &  74.8 &  75.0 &  65.5 & N/A \\ 
& & WD + CP &  29.9 &  70.0 &  70.6 &  71.1 &  65.5 & N/A \\ \hline

\end{tabular}
\end{center}
\vspace*{-1em}
\caption{Generalization error ($E_\textsf{gen}$), test accuracy ($A_\textsf{test}$), and various MIA risks (evaluated using MIAs from Section~\ref{setup:attacks}) of models trained using state-of-the-art regularization techniques. Here we provide MIA risks for regularized models whose accuracy is close to that of DMP-trained models. We note that, for the same test accuracy, DMP-trained models provide significantly higher resistance to MIAs.}
\label{table:regularization_comparison_eq_acc}
\end{table*}

In this section, we show the indistinguishability of the statistics of different features of the target models trained with and without defenses, on the members and non-members of their training data.
Such indistinguishability is necenssary to hinder membership inference attacks  (MIAs)~\cite{shokri2017membership}.

\paragraphb{Effect of softmax temperature. }
Figure \ref{fig:training} shows the effect of softmax temperature, $T$, of \emph{unprotected model}, $\theta_\mathsf{up}$, on the training and test accuracies of the protected mode, $\theta_\textsf{p}$.
As expected, we observe in Figure \ref{fig:training} that with the increase in the softmax temperature of $\theta_\textsf{up}$, the generalization error of $\theta_\textsf{p}$ decreases.
From left to right, the generalization errors of $\theta_\textsf{p}$ when the softmax temperatures of $\theta_\textsf{up}$ are set at 2, 4, and 6 are 4.7\% (66.3, 61.6), 3.6\% (66.7, 63.1), and 0.8\% (55.7, 54.9), respectively; parentheses show the corresponding training and test accuracies, respectively.
We keep the temperature of softmax layer in $\theta_\textsf{p}$ constant at  4.0.
This reduction in generalization error improves membership privacy.

\paragraphb{Indistinguishability of gradient norms. }
To assess the efficacy of DMP against the stronger whitebox MIAs (Nasr et al.~\shortcite{nasr2019comprehensive}), we study the gradients of loss of the predictions of unprotected and protected models on  members and non-members of the private traininig data, $D_\textsf{tr}$.
Figure \ref{fig:gradients} shows the fraction of members and non-members given on y-axes that fall in a particular range of gradient norm values given on x-axes.
Gradients are computed with respect to the parameters of the given model.
We note that the distribution of the norms of unprotected model (upper figures) is heavily skewed to the left for the members, i.e., towards lower gradient norm values, unlike that for the non-members.
This is because, $\theta_\textsf{up}$ memorizes $D_\textsf{tr}$ and its loss and the gradient of the loss on the members is very small compared to the non-members.
However, for the protected model both  members and non-members are evenly distributed across a large range of gradient norm values.
This implies that \emph{DMP significantly reduces the unintended memorization of $D_\textsf{tr}$ in the model parameters}.
Hence, DMP significant reduces (by 27.6\%) the MIA risk to the large capacity Dense19.

\paragraphb{Indistinguishability of train and test accuracies. }
In Figure \ref{fig:generalization_error}, we show the cumulative fraction of classes on y-axis for which the generalization error of the target models is lesser than the corresponding value on the x-axis; the closer the line to the line $x=0$, the lower the generalization error.
Figure \ref{fig:generalization_error} implies that, the models trained using DMP have significantly lower generalization error than those trained using adversarially regularization or without defense.
We observe that, with the no defense case as the baseline, \emph{the generalization error reduction using DMP is more than twice that using adversarial regularization}.
DMP reduces the error by half for Purchase and by $10\times$ for CIFAR100.

\section{Missing experimental details}\label{missing_exp}

\subsubsection{Best tradeoffs due to adversarial regularization}\label{missing_exp:adv}
Table~\ref{table:advreg_best_tradeoffs} gives the results for best tradeoffs due to adversarial regularization that we obtain by tuning its $\lambda$ parameter~\cite{nasr2018machine}.

\begin{table}
\fontsize{8.5}{9}\selectfont{}
\begin{center}
\setlength{\extrarowheight}{0.01cm}

\begin{tabular}{ |c|c|c|c|c|c|c| } 
\hline

{Dataset} & \multicolumn{6}{c|}{Adversarial regularization} \\ \cline{2-7}

\& model & $E_\textsf{gen}$ & $A_\textsf{test}$ & $A_\textsf{wb}$ & $A_\textsf{bb}$ & $A_\textsf{bl}$ & $A_\textsf{nn}$ \\ \hline

P-FC & 22.4 & 68.1 & 62.3 & 61.9 & 61.4 & 51.4  \\ \cline{1-7}

T-FC & 15.5 & 45.3 & 66.8 & 66.3 & 57.8 & 51.2  \\ \cline{1-7}

C100-A & 50.9 & 31.6 & 79.3 & 78.3 & 75.5 & N/A \\ \cline{2-7}

C100-D12  & 19.4 & 58.4 & 61.9 & 61.7 & 59.7 & N/A \\ \cline{2-7}

C100-D19 & 30.8 & 53.7 & 69.5 & 68.7  & 65.4 & N/A \\ \cline{1-7}

C10-A & 29.8 & 62.6 & 65.2 & 65.0 & 64.9 & N/A \\ \cline{1-7}

\end{tabular}
\end{center}
\vspace*{-1em}
\caption{Best tradeoffs between test accuracy ($A_\mathsf{test}$) and membership inference risks (evaluated using MIAs from Section~\ref{setup:attacks}) due to adversarial regularization. DMP significantly improves the tradeoffs over the adversarial regularization (results for DMP are in Table~\ref{table:performance_comparison}).
}
\label{table:advreg_best_tradeoffs}
\vspace*{-1em}
\end{table}

\subsubsection{Best tradeoffs due to other regularizations}\label{missing_exp:other}
We see from the `Equivalent $A_\textsf{test}$' column in Table~\ref{table:regularization_comparison_eq_acc} that all regularization techniques improve the classification performance over the corresponding accuracies of baseline models from the Table 2 of main paper.
However, they reduce overfitting negligibly: the maximum reduction in $E_\text{gen}$ due to the regularizations is 1.8\% for Purchase, 10.2\% for Texas, 3.8\% for CIFAR100, and  2.6\% for CIFAR10.
This is because these techniques aim to produce models that generalize better to test data,
but they do not necessarily reduce the memorization of the private training data by the models.
Consequently, these techniques fail to reduce the membership inference risk: the maximum reduction in $A_\textsf{wb}$ due to the regularizations is 7\% for Purchase, 1.9\% for Texas, 1.9\% for CIFAR100, and 6.8\% for CIFAR10.
Note that, the confidence penalty and the label smoothing techniques reduce the inference risk, but not the generalization error.
This is because the corresponding models have smoother output distributions, which are more indistinguishable than the output distributions of models without any privacy.

\end{document}